\documentclass[journal]{IEEEtran} 
\usepackage{amsmath,amsfonts}
\usepackage{algorithm}
\usepackage{algorithmic}
\usepackage{array}

\usepackage{hyperref}
\hypersetup{colorlinks=true, linkcolor=blue, anchorcolor=blue, citecolor=blue}
\usepackage{balance}

\usepackage{textcomp}
\usepackage{stfloats}
\usepackage{url}
\usepackage{verbatim}
\usepackage[sort]{cite}
\usepackage{epsfig}
\usepackage{amsmath}
\usepackage{amssymb}
\usepackage{subfigure}
\usepackage{multirow}
\usepackage{booktabs}
\usepackage[table,xcdraw]{xcolor}

\usepackage{bibentry}

\begin{document}
\title{Adaptive Guidance Learning for Camouflaged Object Detection}

\author{Zhennan Chen, 
        Xuying Zhang,
        Tian-Zhu Xiang*,
        Ying Tai*
        
        \IEEEcompsocitemizethanks{
            \IEEEcompsocthanksitem Z. Chen and Y. Tai are with the PCALab, School of Intelligence Science and Technology, Nanjing University, Suzhou, China. (e-mail: znexec1025@gmail.com, yingtai@nju.edu.cn).
            \IEEEcompsocthanksitem X. Zhang is with VCIP, 
              College of Computer Science, Nankai University, Tianjin, China. (e-mail: zhangxuying1004@gmail.com).
            \IEEEcompsocthanksitem T.-Z. Xiang is with the G42, Abu Dhabi, UAE. (e-mail: tianzhu.xiang19@gmail.com).
            \IEEEcompsocthanksitem (* Corresponding author: T.-Z. Xiang and Y. Tai)
        }
}





\maketitle

\begin{abstract}
Camouflaged object detection (COD) aims to segment objects visually embedded in their surroundings, which is a very challenging task due to the high similarity between the objects and the background. To address it, most methods often incorporate additional information (\textit{e.g.}, boundary, texture, and frequency clues) to guide feature learning for better detecting camouflaged objects from the background. Although progress has been made, these methods are basically individually tailored to specific auxiliary cues, thus lacking adaptability and not consistently achieving high segmentation performance. 
To this end, this paper proposes an adaptive guidance learning network, dubbed \textit{AGLNet}, which is a unified end-to-end learnable model for exploring and adapting different additional cues in CNN models to guide accurate camouflaged feature learning. 
Specifically, we first design a straightforward additional information generation (AIG) module to learn additional camouflaged object cues, which can be adapted for the exploration of effective camouflaged features. Then we present a hierarchical feature combination (HFC) module to deeply integrate additional cues and image features to guide camouflaged feature learning in a multi-level fusion manner.
Followed by a recalibration decoder (RD), different features are further aggregated and refined for accurate object prediction. 
Extensive experiments on three widely used COD benchmark datasets demonstrate that the proposed method achieves significant performance improvements under different additional cues, and outperforms the recent 20 state-of-the-art methods by a large margin. Our code will be made publicly available at: \textcolor{blue}{{https://github.com/ZNan-Chen/AGLNet}}. 
\end{abstract}


\begin{figure}[t]
    \centering
    \subfigure[FDCOD with different additional cues.]{
		\begin{minipage}[t]{0.99\columnwidth}
			\centering
			\includegraphics[width=\linewidth,height=0.5\columnwidth]{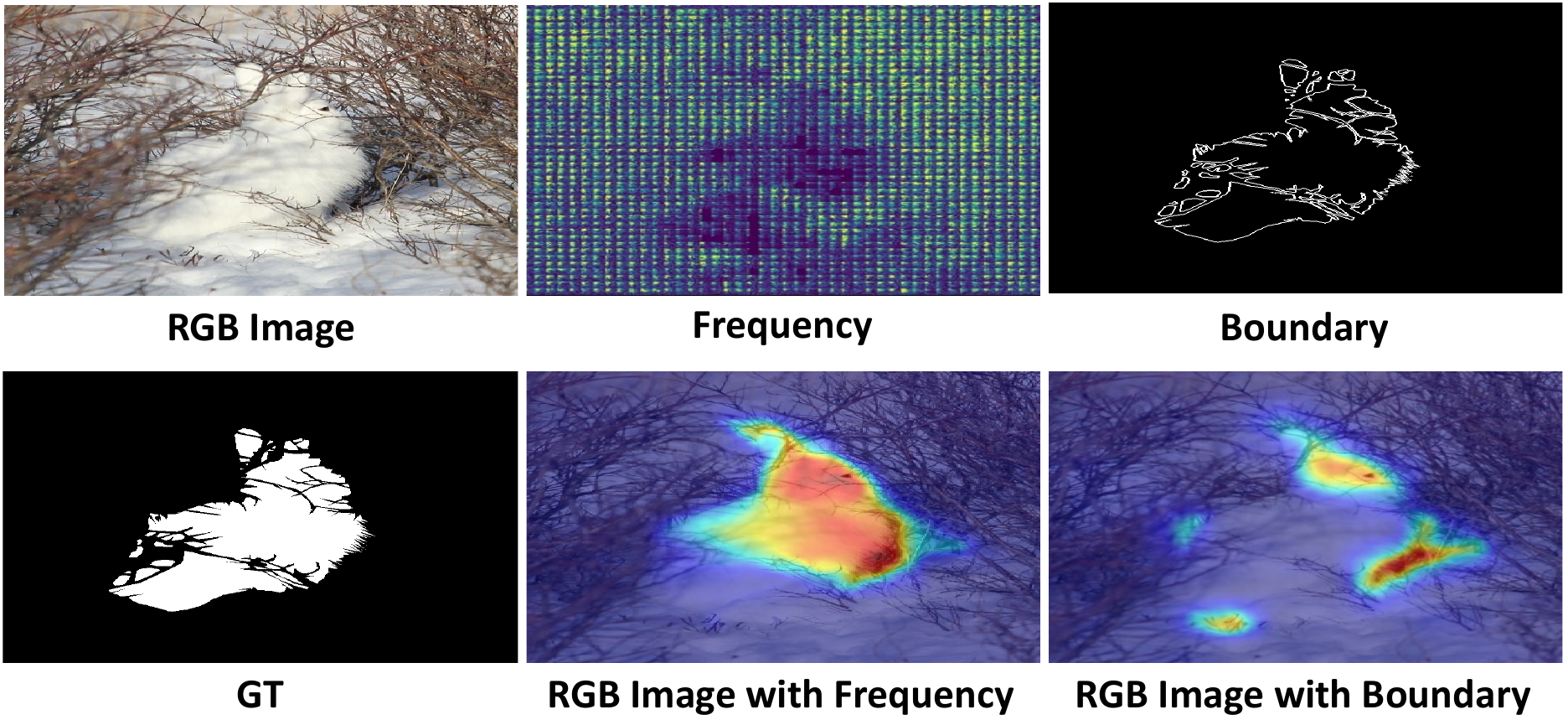}\\
            \label{fig:DGNet_FDCOD_BGNet:FDCOD}
		\end{minipage}%
	}
    \subfigure[DGNet with different additional cues.]{
		\begin{minipage}[t]{0.99\columnwidth}
			\centering
			\includegraphics[width=\linewidth,height=0.5\columnwidth]{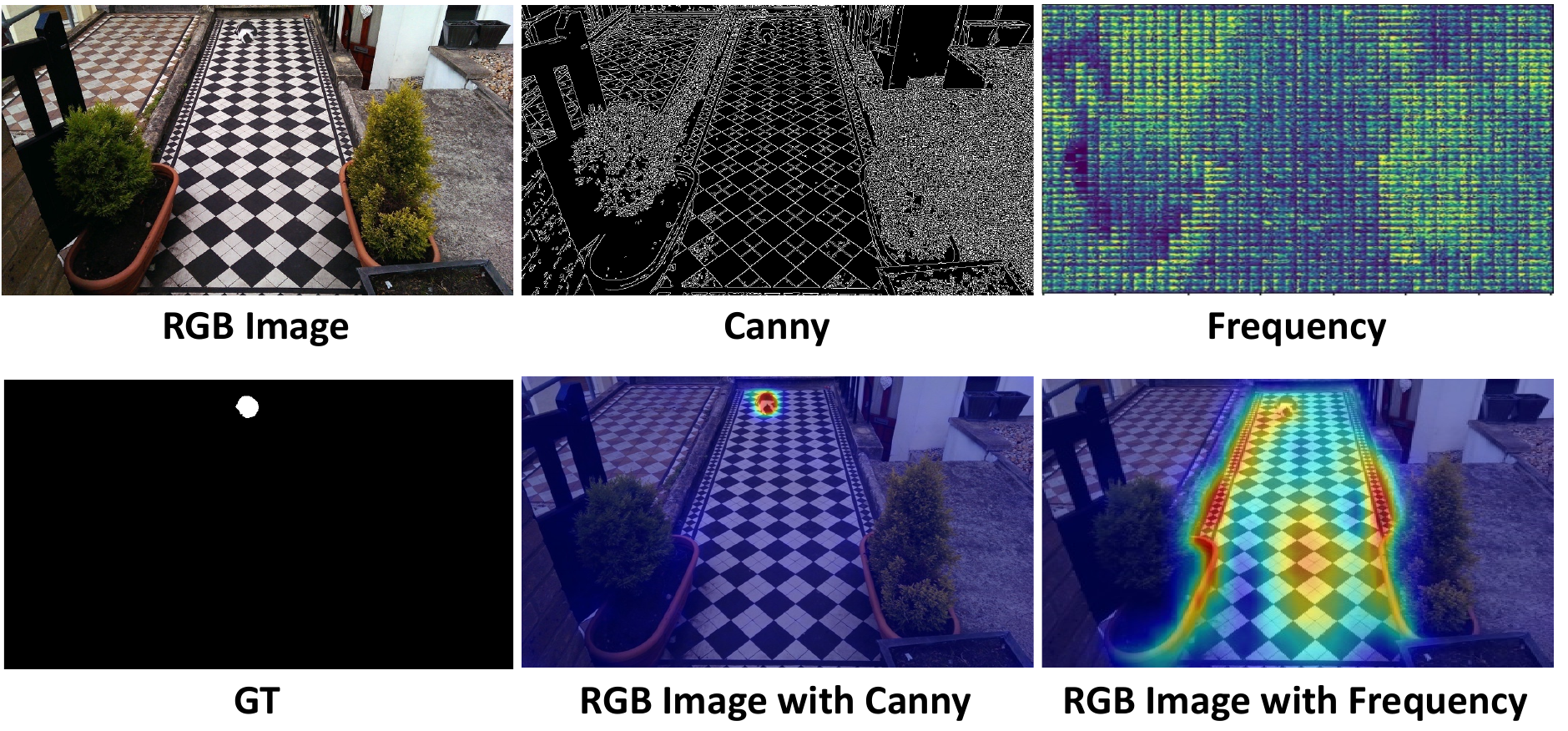}\\
            \label{fig:DGNet_FDCOD_BGNet:DGNet}
		\end{minipage}%
	} 
  \caption{\textbf{Visual comparisons between FDCOD \cite{zhong2022detecting} and DGNet \cite{ji2022gradient} with different additional cues}. We show the feature maps from the previous layer of the network outputs to better visualize the model performance. From (a), we can see that FDCOD well involves frequency domain clues for camouflaged object detection, but is not applicable to boundary cues. From (b), DGNet fails to identify the camouflaged objects with frequency domain clues due to the weak feature changes around objects in the frequency domain. 
  }
  \label{fig:DGNet_FDCOD}
\end{figure}

\section{Introduction}

Camouflaged object detection (COD) is the task of spotting and segmenting objects that are perfectly hidden in complex environments. Recent years have witnessed increasing research enthusiasm from the computer vision community on COD, which facilitates the wide application in various fields, such as medicine (\textit{e.g.}, polyp segmentation~\cite{fan2020pranet} and lung infection segmentation~\cite{fan2020inf}), industry (\textit{e.g.}, surface defect detection~\cite{tabernik2020segmentation} and autonomous driving~\cite{prakash2021multi}), art~\cite{feng2013facilitating} (\textit{e.g.}, recreational arts and style transformation), ecology~\cite{kumar2021early} (\textit{e.g.}, species search) and society~\cite{liu2019concealed} (\textit{e.g.}, search and rescue).

In recent years, numerous deep learning-based methods have been proposed for camouflaged object detection and have made great progress. Some methods adopt the coarse-to-fine learning strategy to explore contextual cues and aggregate multi-scale features for COD, such as SINet~\cite{fan2020camouflaged}, PFNet~\cite{mei2021camouflaged} and FSPNet~\cite{huang2023feature}. Some methods introduce uncertainty-aware learning to model the confidence of model predictions, such as UGTR~\cite{yang2021uncertainty} and ZoomNet~\cite{pang2022zoom}. 
As we know, species often adopt various camouflage strategies to conceal themselves deliberately in the surroundings for self-protection, making the high intrinsic similarities in appearance (\textit{e.g.}, color, texture, and shape) between candidate objects and backgrounds. This camouflage ability of species easily deceives the visual system~\cite{stevens2009animal}, which makes it very difficult to identify camouflaged targets from only a single image feature.  
To address the above limitations, some methods resort to other additional information, such as boundary~\cite{sun2022boundary, zhai2021mutual, zhu2022can}, texture~\cite{zhu2021inferring}, edge~\cite{ji2022gradient}, saliency~\cite{zhang2023referring}, and frequency~\cite{zhong2022detecting}. 
However, we observe that almost all of these methods are designed for a specific type of additional information, and thus lack sufficient adaptability for different types of additional cues, and do not consistently achieve good detection performance. 
For instance, as shown in Fig.~\ref{fig:DGNet_FDCOD_BGNet:FDCOD}, 
FDCOD~\cite{zhong2022detecting} is specially designed to incorporate frequency domain clues for effective camouflaged object detection, but is not applicable to other additional features (\textit{e.g.}, boundary). 
Similarly, as shown in Fig.~\ref{fig:DGNet_FDCOD_BGNet:DGNet}, the spectrum feature shows a small gradient difference around the camouflaged object, so the DGNet~\cite{ji2022gradient}, adapted to additional edge features (\textit{i.e.} Canny), fails to detect the camouflaged object under frequency domain clues.

To this end, we propose an adaptive guidance learning network, termed \textit{AGLNet}, which is able to unify the exploration and guidance of any kind of effective additional cues into an end-to-end learnable model to fully aggregate additional features and image features to guide camouflaged object detection. Specifically, the additional cue is first learned in convolutional space by the designed additional information generation (AIG) module. Then, the learned additional cue is fully integrated with image features in a multi-level fusion manner by the proposed hierarchical feature combination (HFC) module, to guide camouflaged feature learning. After that, a recalibration decoder (RD) is presented to further fuse and refine different features for accurate camouflaged object segmentation through multi-layer, multi-step calibration. It is noted that extensive experiments show that the proposed model can be adapted to explore and incorporate various additional information, such as boundary, edge, texture, and frequency cues. Our contributions can be summarized as follows: 

\begin{itemize}
  \item We propose a powerful adaptive guidance learning network that can involve various additional cues into image features to guide the detection of camouflaged objects. To the best of our knowledge, we are the first to explore a unified end-to-end framework to adapt to various additional information for COD tasks. 
  \item We propose a hierarchical feature combination (HFC) module to deeply integrate additional cues with image features in a multi-level manner to make full use of additional information. Furthermore, we design a recalibration decoder (RD) for iterative calibration and aggregation of different features for object prediction.  
  \item Extensive quantitative and qualitative experiments demonstrate the applicability and effectiveness of the proposed method to different additional information and its superior performance over the recent 20 state-of-the-art COD methods by a large margin.
\end{itemize}

\section{Related Work}
\noindent
\subsection{Camouflaged Object Detection}
Camouflaged object detection (COD) is a challenging task that aims to discover objects that are highly similar to the environment~\cite{chen2023diffusion}. Recent developments in deep learning techniques and the release of large-scale COD datasets (\textit{e.g.}, COD10K~\cite{fan2020camouflaged}) have paved the way for research into deep learning-based camouflaged object detection. After that, several COD methods were proposed, and the performance leaderboard has been continuously refreshed on several widely used COD benchmarks. 
Some methods adopt the coarse-to-fine learning strategy to explore and integrate multi-scale camouflaged features for object detection, such as SINet~\cite{fan2020camouflaged,fan2021concealed}, Camoformer~\cite{yin2022camoformer}, C$^{2}$FNet \cite{sun2021context}, PFNet~\cite{mei2021camouflaged}, SegMaR~\cite{jia2022segment},  PreyNet~\cite{zhang2022preynet} and FSPNet~\cite{huang2023feature}. In particular, C$^{2}$FNet proposes a context-aware fusion network that combines multilevel features and attention coefficients to generate rich contextual information. PFNet uses a disturbance mining strategy to locate potential objects globally and refines predictions by focusing on key regions. SegMaR, pioneering multi-stage detection, excels in small camouflage object scenarios. PreyNet also utilizes multi-stage detection to distinguish between sensory and cognitive mechanisms. Furthermore, FSPNet introduces a novel transformer-based pyramid network, achieving accurate segmentation of camouflaged objects by gradually shrinking neighboring transformer features in a hierarchical manner.
Some methods incorporate confidence-aware learning to improve feature learning for difficult samples, such as UGTR~\cite{yang2021uncertainty} and ZoomNet~\cite{pang2022zoom}. 
UGTR combines Bayesian learning and transformer to address the camouflage object problem by introducing probabilistic information and determinism. ZoomNet considers the expressive characteristics of different scales and enhances feature expression through scale aggregation. 
Inspired by the advances in multi-modal learning~\cite{zhang2021rstnet,zhang2023temo}, 
some methods introduce additional information, such as boundary~\cite{sun2022boundary,zhai2021mutual, zhu2022can, dong2023unified}, edge~\cite{ji2022gradient, he2023camouflaged}, texture~\cite{zhu2021inferring}, fixations~\cite{lv2021simultaneously}, motion~\cite{cheng2022implicit, pang2023zoomnext}, and saliency~\cite{zhang2023referring}, to facilitate the camouflaged feature exploration. 
%
%
Classification~\cite{le2019anabranch} and saliency detection~\cite{li2021uncertainty} are jointly learned with COD based on a multi-task learning framework, respectively. The concept behind~\cite{le2019anabranch} and~\cite{li2021uncertainty} is that introducing different tasks can enhance the accuracy and robustness of camouflage detection segmentation.
More recently, collaborative feature exploration from a group of relevant images has been proposed to enhance camouflaged object detection performance via learning from multiple images with objects of the same category~\cite{zhang2023collaborative,zhang2023referring}. 


\begin{figure*}[t]
\includegraphics[width=\textwidth]{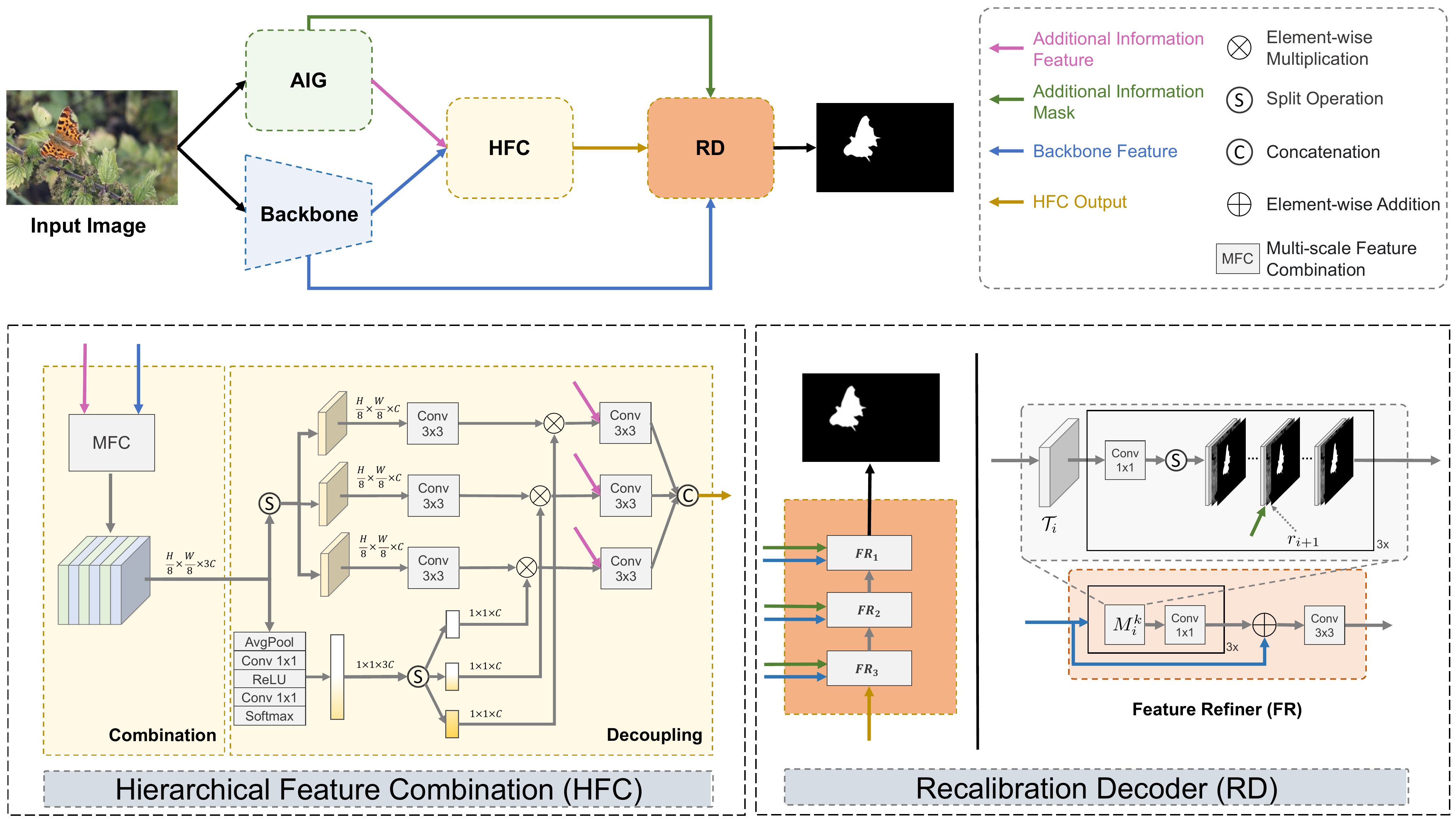}
\caption{\textbf{Overall architecture of our adaptive guidance learning network (AGLNet) for COD.} The input image is processed by a visual backbone and an additional information generation (AIG) module to extract multi-scale image features and learn additional cues, respectively. Both sets of features are deeply integrated to guide the learning of camouflaged features in the hierarchical feature combination (HFC) module, which consists of combination and decoupling. Finally, the fused features are iteratively aggregated and refined with backbone and additional features by the recalibration decoder (RD) for object prediction.  
%
}
\label{fig:overall}
\end{figure*}

\noindent
\subsection{Additional Cues for COD}

In salient object detection, many method have attempted to enhance performance by integrating additional information, such as edge information~\cite{Zhao_2019_ICCV,wang2019salient} and high-resolution input~\cite{zeng2019towards, hu2023high}. Zhao \emph{et al.}~\cite{Zhao_2019_ICCV} have utilized extensive edge and location information to more precisely locate the boundaries of salient objects. 
Zeng \emph{et al.}~\cite{zeng2019towards} have input higher resolution image features, combining global semantic information with local high-resolution details to iteratively produce high-resolution predictions. 

Building on these salient object detection methods, various studies have explored integrating additional cues into camouflaged object detection. 

By introducing auxiliary cues such as texture maps and edge maps into camouflaged models, these models can be sensitized to discern subtle distinctions between the foreground and background elements, notably variations in texture, the salience of edges, or gradient transitions. Moreover, a segment of the academic community posits that exclusive reliance on the RGB color model may not exhaustively harness the entirety of data inherent in images. Consequently, the frequency domain has been proposed as an ancillary cue. Frequency domain analysis can provide information about different frequencies in an image, which might not be prominent or easily detectable in the RGB domain. 
For example, Zhu \emph{et al.} \cite{zhu2022can} and Sun \emph{et al.} \cite{sun2022boundary} have introduced boundary cues to highlight the camouflaged boundary between the background and foreground of an image and enhance the understanding of the boundary by the model. Ji \emph{et al.} \cite{ji2022fast} and He \emph{et al.} \cite{he2023camouflaged} have incorporated edge information for exploring the semantic information of target edges.
Zhu \emph{et al.} \cite{zhu2021inferring} combined texture labels to make the network more focused on the structure and details of the target.
Zhong \emph{et al.}  \cite{zhong2022detecting} and Cong \emph{et al.} \cite{cong2023frequency} have used  the frequency domain cues to improve camouflaged object detection. He \emph{et al.} \cite{He_2023_CVPR} decomposes foreground and background features into different frequency bands while constructing edge information to assist in generating accurate predictions.


Despite the strides made in amalgamating image features with auxiliary cues for COD, there are still challenges to be addressed. Predominantly, the extant approaches are bespoke solutions tailored to specific types of additional information, thus limiting their applicability to other additional types of cues. In this paper, we propose a novel adaptive guidance learning model that alleviates this issue by a unified end-to-end framework to adapt any kind of additional information and guide camouflaged feature learning by hierarchical feature combination.


\section{Methodology}
\label{sec3}
The overall architecture of the proposed AGLNet is shown in Fig.~\ref{fig:overall}. The framework first uses additional information generation (AIG) to learn additional cues, which can be used as guidance for camouflaged feature learning. Then, the learned additional features are deeply integrated with multi-scale backbone features to explore the critical camouflaged object features by the designed hierarchical feature combination (HFC) module. To make full use of additional cues, we adopt a multi-level fusion manner to incorporate additional information at the combination stage and decoupling stage. After that, a recalibration decoder (RD) is adopted to aggregate and refine multiple features in a multi-level calibration manner for camouflaged object segmentation.

\subsection{Additional Information Generation (AIG)}
\label{subsec-AIG}
The additional cues contain valuable features that are not perceived in the backbone network. Some additional information, such as frequency domain cues, also shows large modal differences from RGB spatial domain features. If the two features are integrated directly, they may interfere with each other, resulting in the loss of key features or the introduction of noise. 
To avoid this issue, we design a simple but effective additional information generation (AIG) module to learn the additional cues in CNN space, so that they can be easily merged into image features. Specifically, AIG contains three layers, where each consisting of an averaging pooling operation and a convolution operation. AIG learns the additional feature $\mathcal{A} \in \mathbb{R}^{\frac{H}{8} \times \frac{W}{8} \times C}$ ($C$ = 64) from the input RGB image $\mathbf{I} \in \mathbb{R}^{H\times W \times 3}$, which is the explicit semantic cues complementary to backbone features. Then a 1$\times$1 convolution layer is adopted to produce the prediction of additional cues $r^{s}$, supervised by the ground truth of additional information. 
We follow the existing approaches in the COD task to obtain additional information labels including boundary, texture, edge (\textit{i.e.} Canny), and frequency, which are detailed below.

\begin{itemize}
    \item Boundary~\cite{sun2022boundary,zhai2021mutual,zhu2022can}. Get the object boundary from ground truth (GT) as the corresponding boundary map. 
    
    \item Texture~\cite{zhu2021inferring}. Get object texture via contour edge map (ConEdge), texture map (Texture), and GT:
    \begin{equation}
        \text{ConEdge} + \text {Canny} \times GT = \text{Texture}
    \end{equation}

    \item Canny~\cite{ji2022gradient}. Get the object's canny information via standard canny edge detector~\cite{canny1986computational} and GT:
    \begin{equation}
        \mathbf{Z}^{G}=\mathcal{F}_{E}(\mathcal{I}(x, y)) \otimes \mathbf{Z}^{C}
    \end{equation}
    where $\mathcal{F}_{E}$ represents the standard canny edge
    detector for input $\mathcal{I}(x, y)$ with discrete pixel coordinates ($x, y$). $\otimes$ means element-wise multiplication. $\mathbf{Z}^{C}$ means the object-level ground-truth.

    \item Frequency~\cite{zhong2022detecting}. Get frequency domain information by discrete cosine transform (DCT) of RGB images.

\end{itemize}

\subsection{Hierarchical Feature Combination (HFC)}
\label{subsec-FCD}

\vspace{10pt}
\noindent
\textbf{Preliminary.}
Motivated by the observation~\cite{wu2019cascaded} that low-level features consume more computational resources and contribute less to performance, we adopt the top-three high-level features of the visual backbone as our multi-scale backbone features, denoted as $\mathcal{X}^{r}_{i}$, $i \in \{1, 2, 3\}$ whose resolution is $\frac{H}{k} \times \frac{W}{k}$, $k \in \{8, 16, 32\}$.

\vspace{10pt}
\noindent\textbf{Combination.} 
We observe that cascade structures can efficiently aggregate multi-level features for accurate object detection~\cite{wu2019cascaded} and the cooperation of adjacent features can localize objects well. Therefore, we designed a novel multi-scale feature combination (MFC) module. 
First, we build a convolution block with different kernel sizes to enhance visual features. Specifically, we process feature $\mathcal{X}^{r}_{i}$ with a $1\times1$  convolution operation, followed by two parallel convolution operations with a $5\times5$ kernel and a $7\times7$ kernel, respectively. Then, the element-wise summation is performed over the features of the two branches. Finally, the summed features are fed to a $3\times3$ convolution layer to get the final result $\mathcal{X}^{c}_{i} \in  \mathbb{R}^{\frac{H}{k} \times \frac{W}{k} \times C}$.
We utilize the high-resolution details of the shallow layers for accurate localization and the semantic information of the deep layers to ensure semantic consistency between layers. It is defined as: 
\begin{equation}
\left\{
\begin{aligned}
g_{3} &= \mathcal{X}^{c}_{3},\\
g_{2} &= \mathcal{X}^{c}_{2} \otimes \rm UP_{\times 2}\left(g_{3}\right), \\
g_{1} &= \mathcal{X}^{c}_{1} \otimes \rm UP_{\times 2}\left(g_{2}\right) \otimes \rm UP_{\times 2}(\mathcal{X}^{c}_{2}) \otimes \rm UP_{\times 4}(g_{3})
\end{aligned}
\right.
\end{equation}
where $\rm UP_{\times t}(\cdot)$ denotes a $\rm t \times$ bilinear upsampling operation. $g_{i}  \in  \mathbb{R}^{\frac{H}{k} \times \frac{W}{k} \times C}$.

Next, those features are integrated with the additional information feature $\mathcal{A}$ to generate the initial combined feature $\mathcal{S}$. The integration process can be denoted as:
\begin{equation}
\left\{
\begin{aligned}
    \mathcal{S}_{3} &= \rm Conv_{3\times3}([\rm Dn_{\times 4}(\mathcal{A}), g_{3}]),\\
    \mathcal{S}_{2} &= \rm Conv_{3\times3}([\rm Dn_{\times 2}(\mathcal{A}), g_{2}, \mathcal{S}_{3}]), \\
    \mathcal{S}_{1} &= \rm Conv_{3\times3}([\mathcal{A},g_{1},\mathcal{S}_{2}]),\\
    \mathcal{S} &= \rm Conv_{3\times3}(\mathcal{S}_{1}),
\end{aligned}
\right.
\end{equation}
where $[\cdot]$ is concatenation, $\rm Dn_{\times t}(\cdot)$ is a $\rm t \times$ bilinear downsampling and ${\rm Conv}_{3\times3}$ is a 3$\times$3 convolution operation. The corresponding channel numbers of $\mathcal{S}_{3}, \mathcal{S}_{2}, \mathcal{S}_{1}, \mathcal{S}$ are $C$, 2$C$, 3$C$, 3$C$, respectively.

\vspace{10pt}
\noindent
\textbf{Decoupling.} 
To further explore camouflaged object semantics, we design a dual-branch architecture to guide decoupling. 
In the first branch, the $\mathcal{S}$ is decoupled into three groups of features, \textit{i.e.}, $\{s_{1}, s_{2}, s_{3}\}$, which are then processed by a convolution operation, respectively. In the other branch, the $\mathcal{S}$ is first processed by two convolution layers after an average pooling. The activation function of the last convolution layer is Softmax, which is used to learn the weights of feature channels, namely $w \in \mathbb{R}^{1 \times 1 \times 3C}$.
The $w$ is then split into $\{w_{1}, w_{2}, w_{3}\}$. Each decoupled feature is multiplied by its corresponding weight.

To capture more features of camouflaged objects, additional information features are incorporated to guide feature learning of camouflaged objects. The above operations are described as:
\begin{equation}
\left\{
\begin{aligned}
    d_{1} &= {\rm Conv}_{3\times3}([w_{1} \otimes {\rm Conv}_{3\times3}(s_{1}),\mathcal{A}]),\\
    d_{2} &= {\rm Conv}_{3\times3}([w_{2} \otimes {\rm Conv}_{3\times3}(s_{2}),\mathcal{A}]), \\
    d_{3} &= {\rm Conv}_{3\times3}([w_{3} \otimes {\rm Conv}_{3\times3}(s_{3}),\mathcal{A}]),
\end{aligned}
\right.
\end{equation} 
where $\otimes$ denotes element-wise multiplication. Then, we obtain the initial prediction map $r_{4} = {\rm Conv}_{1\times1}$ $([d_{1},d_{2},d_{3}])$.

\begin{table}[ht]
\caption{The specific parameters of $M^{k}_{i}$ module.}
\centering
\footnotesize
\renewcommand\arraystretch{1.5}{
\begin{tabular}{c|c|c|c}
\toprule
\textbf{No.} & \textbf{Operation}       & \textbf{Input Size} & \textbf{Output Size} \\ \midrule
\multirow{2}{*}{\#1} & Conv$_{1\times1}$         & $ 4 \times C$           & $ 4 \times \frac{C}{2^{q}}$            \\ \cline{2-4} 
                     & Split \& Concat &  $ 4 \times \frac{C}{2^{q}}$          &$ 4 \times \frac{C}{2^{q}}+2^{(n+1)}$             \\\hline
\multirow{2}{*}{\#2} & Conv$_{1\times1}$          & $ 4 \times \frac{C}{2^{q}}+2^{(n+1)}$           & $ 3 \times \frac{C}{2^{q}}$            \\ \cline{2-4} 
                     & Split \& Concat & $ 3 \times \frac{C}{2^{q}}$           & $ 3 \times \frac{C}{2^{q}}+2^{(n+1)}$            \\ \hline
\multirow{2}{*}{\#3} & Conv$_{1\times1}$         &$ 3 \times \frac{C}{2^{q}}+2^{(n+1)}$            & $ 2 \times \frac{C}{2^{q}}$            \\ \cline{2-4} 
                     & Split \& Concat &$ 2 \times \frac{C}{2^{q}}$            & $ 2 \times \frac{C}{2^{q}}+2^{(n+1)}$    \\ 
                     \bottomrule       
\end{tabular}
}

\label{FR}
\end{table}

\subsection{Recalibration Decoder (RD)} 
\label{subsec-RD}
Inspired by \cite{liu2022cmx}, after extracting the aggregated features, we design a recalibration decoder (RD) module which employs iterative calibration to refine the consistency of image features and additional features. 
RD further combines multi-scale backbone features and additional features to enhance feature representation. It consists of three levels of iterative optimization, and each level is a well-designed feature refiner (FR), whose architecture is shown in Fig.~\ref{fig:overall}. 
For each FR, the backbone visual feature $\mathcal{X}^{r}_i$ of the corresponding scale is first split, and then is combined with the prediction map from the previous level and learned additional features. 
In FR, we perform multiple feature splits and merges, where the number of splits is $n=\{4,3,2\}$, and each split feature is merged with $r_{i+1}$ and additional cue mask. The specific parameters of $M^{k}_{i}$ module is shown in Table \ref{FR}, where $q=\{2^{2}, 2^{1}, 2^{0}\}$.
The RD module can be formulated as:
\begin{equation}
\left\{
\begin{aligned}
    r_{3} &= {\rm FR}_{3}(\mathcal{X}^{r}_{3}, r_{4}, r^{s}),\\
    r_{2} &= {\rm FR}_{2}(\mathcal{X}^{r}_{2}, r_{3}, r^{s}), \\
    r_{1} &= {\rm FR}_{1}(\mathcal{X}^{r}_{1}, r_{2}, r^{s}),
\end{aligned}
\right.
\end{equation}

On one hand, FR can facilitate the fusion of image features and additional features at different scales. On the other hand, FR adopts multiple iterations to boost accurate segmentation within a certain scale. 


\subsection{Loss Function}
\label{subsec-loss}
Our loss function consists of additional information generation loss and camouflaged object detection loss. For the former, we reshape $r^{s}$ to the input image size and calculate the MSE loss. For the latter, we also reshape each prediction ($r_i$) to the input image size and adopt the weighted BCE loss ($\mathcal{L}_{BCE}$) and the weighted IoU loss ($\mathcal{L}_{IoU}$) \cite{wei2020f3net}. 
Therefore, our loss function is defined as:
\begin{equation}
    \begin{aligned}
        \mathcal{L}_{total} =  
        &\sum\nolimits_{i=1}^{3}(\mathcal{L}_{BCE}(r_{i}, GT) + \mathcal{L}_{IoU}(r_{i}, GT)) + \\ 
        &\mathcal{L}_{MSE}(r^{s}, \mathcal{D}^{s}),
    \end{aligned}
\end{equation}

\begin{table*}[t]
\caption{Quantitative comparisons of our proposed method and other 20 state-of-the-art methods on three widely used benchmark datasets. The higher the $S_{\alpha}$, $F_{\beta}^{\omega}$, $F_{m}$, and $E_{m}$, the better the performance. The smaller the $MAE$, the better. The best results are marked in $\mathbf{bold}$. The second-best results are marked in \underline{underline}.}
\resizebox{\textwidth}{!}{
\renewcommand{\arraystretch}{1.3}
\begin{tabular}{c|ccccccccccccccc}
\toprule[1pt]
\multirow{2}{*}{\textbf{Method}} & \multicolumn{5}{c|}{\textbf{COD10K}}                                                                       & \multicolumn{5}{c|}{\textbf{NC4K}}                                                                  & \multicolumn{5}{c}{\textbf{CAMO}}                                                                                                                      \\ \cline{2-16} 
                                 & $S_{\alpha}\uparrow$              & $F_{\beta}^{\omega}\uparrow$             & $F_{m}\uparrow$             & $E_{m}\uparrow$             & \multicolumn{1}{c|}{$MAE\downarrow$}                                 & $S_{\alpha}$              & $F_{\beta}^{\omega}\uparrow$             & $F_{m}\uparrow$             & $E_{m}\uparrow$             & \multicolumn{1}{c|}{$MAE\downarrow$}                                 & $S_{\alpha}$              & $F_{\beta}^{\omega}\uparrow$             & $F_{m}\uparrow$             & $E_{m}\uparrow$             & \multicolumn{1}{c}{$MAE\downarrow$}            \\ \midrule[1pt]
\multicolumn{1}{l|}{2020 SINet \cite{fan2020camouflaged}}    & 0.772          & 0.543          & 0.640          & 0.810          & \multicolumn{1}{c|}{0.051}          & 0.810          & 0.665          & 0.741          & 0.841          & \multicolumn{1}{c|}{0.066}          & 0.753          & 0.602          & 0.676          & 0.774          & \multicolumn{1}{c}{0.097}          \\
\multicolumn{1}{l|}{2021 PFNet \cite{mei2021camouflaged}}        & 0.797          & 0.656          & 0.698          & 0.875          & \multicolumn{1}{c|}{0.039}          & 0.826          & 0.743          & 0.783          & 0.884          & \multicolumn{1}{c|}{0.054}          & 0.774          & 0.683          & 0.737          & 0.832          & \multicolumn{1}{c}{0.087}          \\
\multicolumn{1}{l|}{2021 LSR \cite{lv2021simultaneously}}          & 0.805          & 0.660           & 0.703          & 0.876          & \multicolumn{1}{c|}{0.039}          & 0.832          & 0.743          & 0.785          & 0.888          & \multicolumn{1}{c|}{0.053}          & 0.793          & 0.703          & 0.753          & 0.850           & \multicolumn{1}{c}{0.083}          \\
\multicolumn{1}{l|}{2021 C$^{2}$FNet \cite{sun2021context}}    & 0.811          & 0.680          & 0.722          & 0.890          & \multicolumn{1}{c|}{0.036}          & 0.839          & 0.763          & 0.805          & 0.896          & \multicolumn{1}{c|}{0.050}          & 0.782          & 0.698          & 0.751          & 0.838          & \multicolumn{1}{c}{0.082}          \\
\multicolumn{1}{l|}{2021 MGL \cite{zhai2021mutual}}       & 0.815          & 0.667          & 0.709          & 0.852          & \multicolumn{1}{c|}{0.035}          & 0.832          & 0.739          & 0.782          & 0.868          & \multicolumn{1}{c|}{0.053}          & 0.772          & 0.670          & 0.725          & 0.811          & \multicolumn{1}{c}{0.089}         \\
\multicolumn{1}{l|}{2021 UGTR \cite{yang2021uncertainty}}      & 0.818          & 0.668          & 0.725          & 0.894          & \multicolumn{1}{c|}{0.035}          & 0.839          & 0.749          & 0.812          & 0.892          & \multicolumn{1}{c|}{0.048}          & 0.784          & 0.687          & 0.741          & 0.844          & \multicolumn{1}{c}{0.086}                   \\
\multicolumn{1}{l|}{2021 UJSC \cite{li2021uncertainty}}      & 0.818          & 0.702          & 0.737          & 0.892          & \multicolumn{1}{c|}{0.033}          & 0.840          & 0.772          & 0.817          & 0.899          & \multicolumn{1}{c|}{0.047}          & 0.793          & 0.721          & 0.766          & 0.854          & \multicolumn{1}{c}{0.078}          \\
\multicolumn{1}{l|}{2022 SINet-V2 \cite{fan2021concealed}}   & 0.815          & 0.674          & 0.711          & 0.885          & \multicolumn{1}{c|}{0.037}          & 0.848          & 0.768          & 0.801          & 0.902          & \multicolumn{1}{c|}{0.047}          & 0.819          & 0.743          & 0.781          & 0.882          & \multicolumn{1}{c}{0.070}          \\
\multicolumn{1}{l|}{2022 R-MGL\_v2 \cite{zhai2022mgl}}          & 0.816          & 0.689         & 0.733          & 0.879          & \multicolumn{1}{c|}{0.034}          & 0.838          & 0.758          & 0.801          & 0.899          & \multicolumn{1}{c|}{0.050}          & 0.769          & 0.672          & 0.731          & 0.847          & \multicolumn{1}{c}{0.086}          \\
\multicolumn{1}{l|}{2022 BSANet \cite{zhu2022can}}    & 0.818          & 0.699          & 0.738          & 0.890          & \multicolumn{1}{c|}{0.034}          & 0.841          & 0.771          & 0.817          & 0.897          & \multicolumn{1}{c|}{0.048}          & 0.794          & 0.717          & 0.763          & 0.851          & \multicolumn{1}{c}{0.079}          \\
\multicolumn{1}{l|}{2022 FAPNet \cite{zhou2022feature}}    & 0.822          & 0.694          & 0.731          & 0.888          & \multicolumn{1}{c|}{0.036}      & 0.851              & 0.775              & 0.810              & 0.899          & \multicolumn{1}{c|}{0.047}          & 0.815          & 0.734          & 0.776          & 0.865         & \multicolumn{1}{c}{0.076}          \\
\multicolumn{1}{l|}{2022 BGNet \cite{sun2022boundary}}    & 0.831          & 0.722          & 0.753          & 0.901          & \multicolumn{1}{c|}{0.033}    & 0.851          & 0.788          & 0.820   & 0.907        & \multicolumn{1}{c|}{0.044}          & 0.812          & 0.749          & 0.789          & 0.870          & \multicolumn{1}{c}{0.073}          \\
\multicolumn{1}{l|}{2022 SegMaR \cite{jia2022segment}}    & 0.833          & 0.724          & 0.757          & 0.899          & \multicolumn{1}{c|}{0.034}          & 0.841          & 0.781          & 0.820          & 0.896          & \multicolumn{1}{c|}{0.046}          & 0.815          & 0.753          & 0.795          & 0.874          & \multicolumn{1}{c}{0.071}          \\
\multicolumn{1}{l|}{2022 FDCOD \cite{zhong2022detecting}}     & 0.837 & 0.731  & 0.749          & {0.918}          & \multicolumn{1}{c|}{0.030}          & 0.834          & 0.750          & 0.784          & 0.894          & \multicolumn{1}{c|}{0.052}          & 0.844          & 0.778          & 0.809          & 0.898          & \multicolumn{1}{c}{0.062}              \\ 
\multicolumn{1}{l|}{2022 ZoomNet \cite{pang2022zoom}}      & 0.838          & 0.729      & 0.766          & 0.888          & \multicolumn{1}{c|}{{0.029}} & 0.853          & 0.784          & 0.818          & 0.896          & \multicolumn{1}{c|}{0.043}          & 0.820          & 0.752          & 0.794          & 0.878          & \multicolumn{1}{c}{0.066}          \\ 
\multicolumn{1}{l|}{2023 DGNet \cite{ji2022gradient}}    & 0.822          & 0.693          & 0.728          & 0.896          & \multicolumn{1}{c|}{0.033}          & {0.857}          & 0.784          & 0.814          & 0.911          & \multicolumn{1}{c|}{0.042}          & 0.839          & 0.769          & 0.806          & 0.901          & \multicolumn{1}{c}{0.057}   \\
\multicolumn{1}{l|}{2023 FEDER \cite{he2023camouflaged}}       & 0.822          & 0.716          & 0.751          & 0.900          & \multicolumn{1}{c|}{0.032}          & 0.847          & 0.789          & 0.824          & 0.907          & \multicolumn{1}{c|}{0.044}          & 0.802          & 0.738          & 0.781          & 0.867          & \multicolumn{1}{c}{0.071}          \\
\multicolumn{1}{l|}{2023 PopNet \cite{wu2023source}}       & 0.851          & 0.757          & 0.786          & 0.910          & \multicolumn{1}{c|}{0.028}          & 0.861          & 0.802          & 0.833          & 0.909          & \multicolumn{1}{c|}{0.042}          & 0.808          & 0.744          & 0.784          & 0.859          & \multicolumn{1}{c}{0.077}          \\
\multicolumn{1}{l|}{2023 HitNet \cite{hu2023high}}        & \underline{0.868}         & \textbf{0.798}           & \underline{0.806}          & \underline{0.932}          & \multicolumn{1}{c|}{\underline{0.024}}          & 0.870          & \underline{0.825}          & \underline{0.853}          & \underline{0.921}          & \multicolumn{1}{c|}{0.039}          & 0.844          & \underline{0.801}          & \underline{0.831}          & \underline{0.902}          & \multicolumn{1}{c}{0.057}
\\
\multicolumn{1}{l|}{2023 FSPNet \cite{huang2023feature}}        & 0.851          & 0.735          & 0.769          & 0.895          & \multicolumn{1}{c|}{0.026}          & \underline{0.879}          & 0.816          & 0.843          & 0.915          & \multicolumn{1}{c|}{\underline{0.035}}          & \underline{0.856}          & 0.799          & 0.830          & 0.899          & \multicolumn{1}{c}{\underline{0.050}}          \\ \hline
\rowcolor[HTML]{EFEFEF}
\multicolumn{1}{l|}{AGLNet-Boundary}        & 0.870 & 0.785 & 0.808 & 0.930 & \multicolumn{1}{c|}{0.024} & 0.883 & 0.830 & 0.854 & 0.929 & \multicolumn{1}{c|}{0.035} & 0.867 & 0.816 & 0.843 & 0.917 & \multicolumn{1}{c}{0.053}  \\
\rowcolor[HTML]{EFEFEF}
\multicolumn{1}{l|}{AGLNet-Texture}        & 0.871 & 0.786 & 0.809 & 0.928 & \multicolumn{1}{c|}{0.024} & 0.884 & 0.834 & 0.857 & 0.930 & \multicolumn{1}{c|}{0.034} & 0.868 & 0.823 & 0.850 & \textbf{0.920} & \multicolumn{1}{c}{\textbf{0.049}}  \\
\rowcolor[HTML]{EFEFEF}
\multicolumn{1}{l|}{AGLNet-Canny}        & 0.873 & 0.789 & 0.811 & 0.930 & \multicolumn{1}{c|}{0.023} & 0.884 & 0.833 & 0.855 & 0.929 & \multicolumn{1}{c|}{0.034} & 0.870 & 0.823 & 0.848 & 0.916 & \multicolumn{1}{c}{0.050}  \\
\rowcolor[HTML]{EFEFEF}
\multicolumn{1}{l|}{AGLNet-Frequency}        & \textbf{0.875} & \underline{0.791} & \textbf{0.813} & \textbf{0.933} & \multicolumn{1}{c|}{\textbf{0.023}} & \textbf{0.889} & \textbf{0.836} & \textbf{0.858} & \textbf{0.934} & \multicolumn{1}{c|}{\textbf{0.033}} & \textbf{0.874} & \textbf{0.825} & \textbf{0.851} & 0.918 & \multicolumn{1}{c}{0.050}  \\
\bottomrule[1pt]
\end{tabular}
}
\label{tab:quant}
\end{table*}

\begin{figure*}[t]
	\centering
	\subfigure[{\scriptsize Image}]{
		\begin{minipage}[t]{0.1\textwidth}
			\centering
			\includegraphics[width=1.15\textwidth,height=0.840\textwidth]{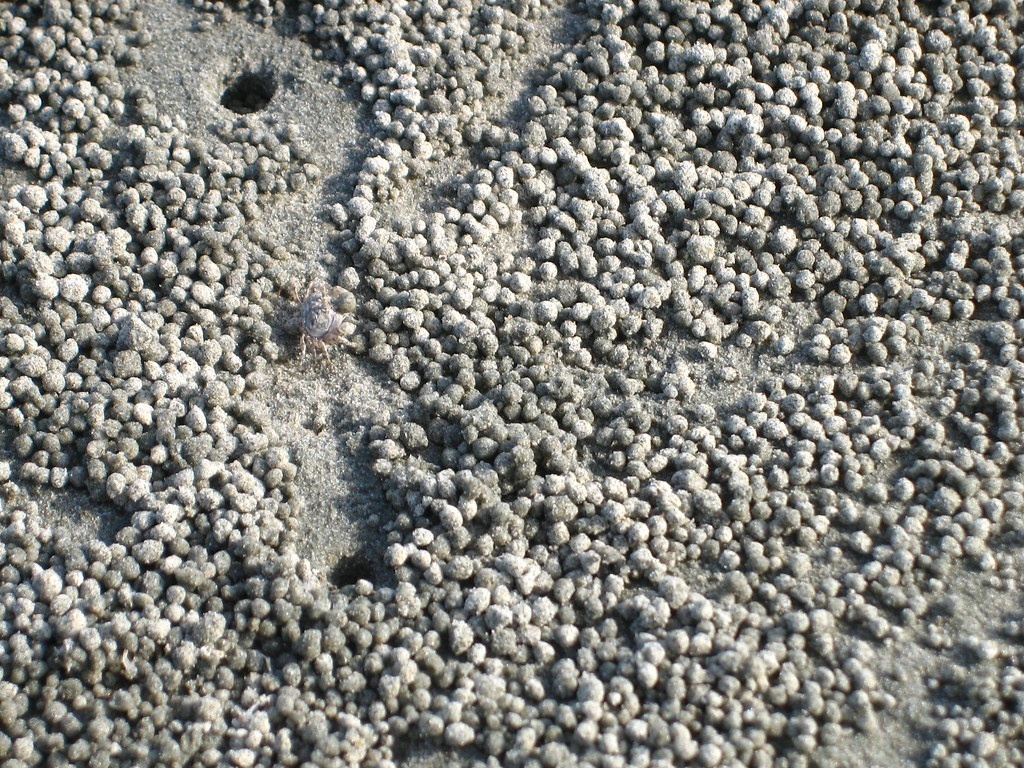}\\
			\vspace{0.01\linewidth}
            \includegraphics[width=1.15\textwidth,height=0.840\textwidth]{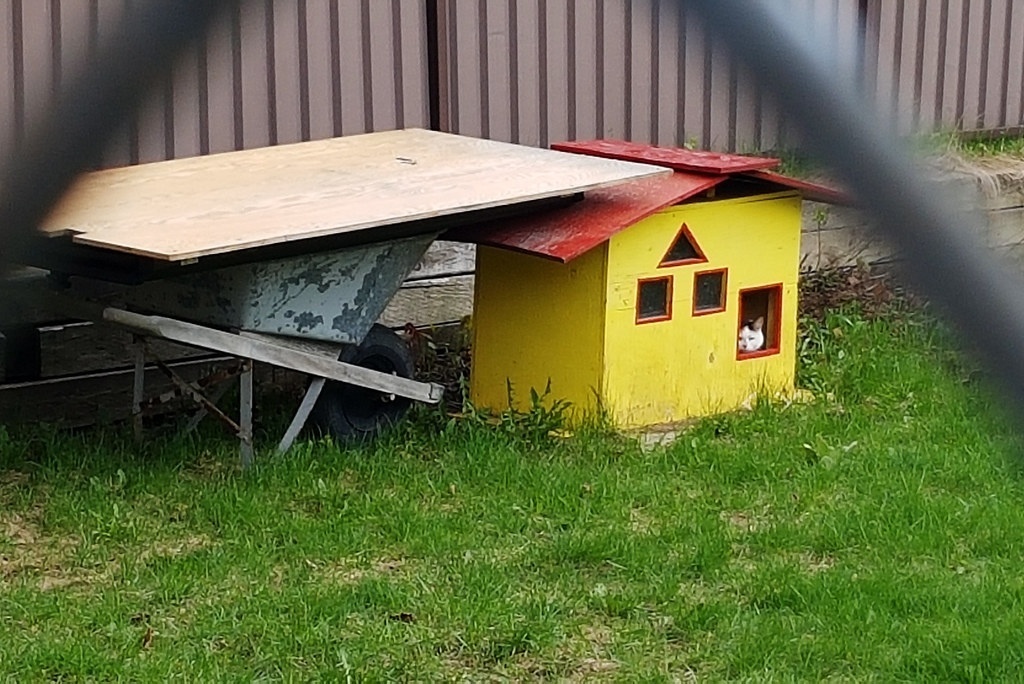}\\
            \vspace{0.01\linewidth}
			\includegraphics[width=1.15\linewidth,height=0.840\textwidth]{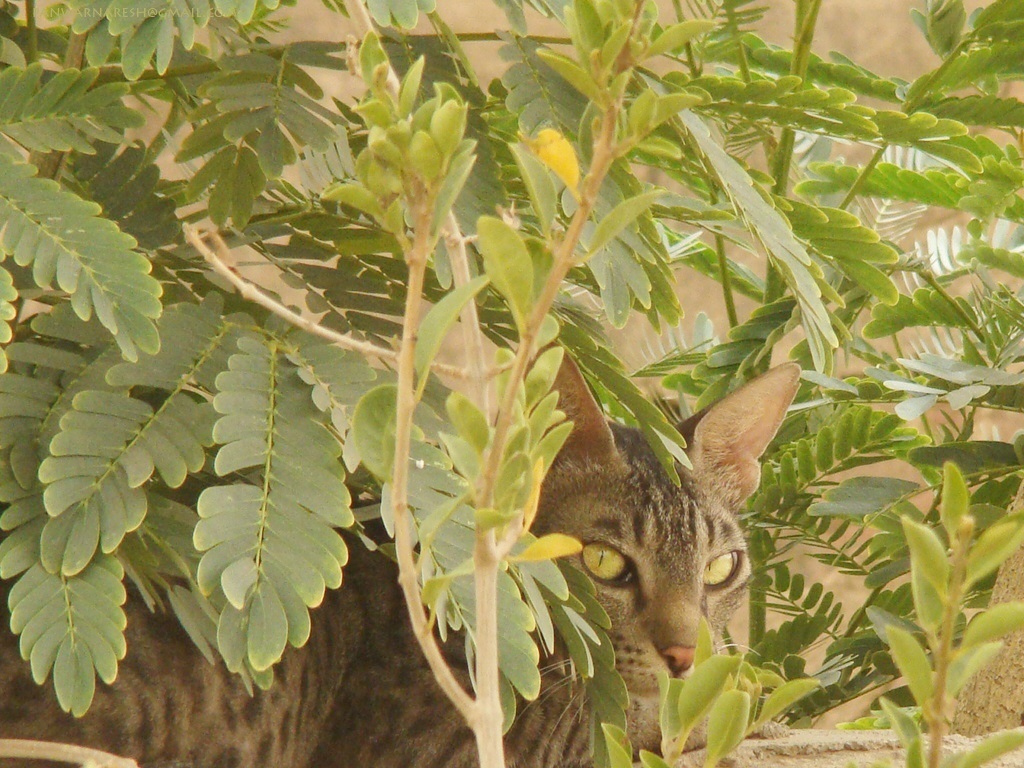}\\
			\vspace{0.01\linewidth}
			\includegraphics[width=1.15\linewidth,height=0.840\textwidth]{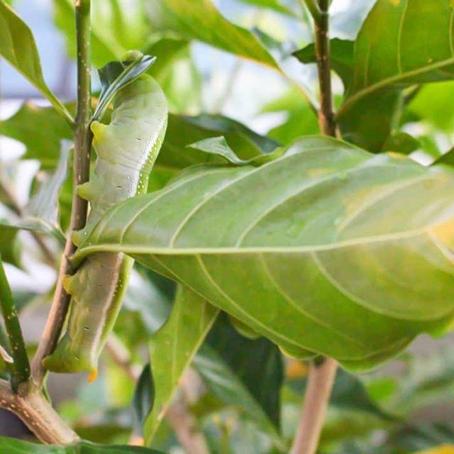}\\
			\vspace{0.01\linewidth}
			\includegraphics[width=1.15\linewidth,height=0.840\textwidth]{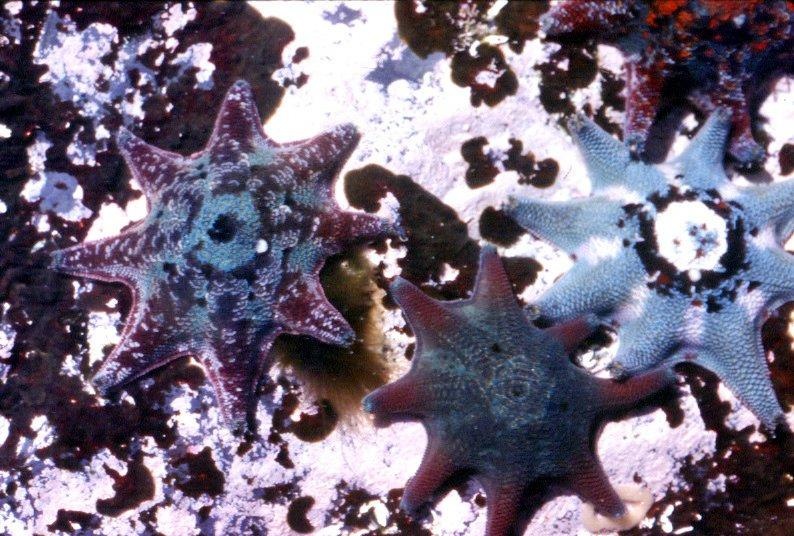}\\
			\vspace{0.01\linewidth}

            \includegraphics[width=1.15\linewidth,height=0.840\textwidth]{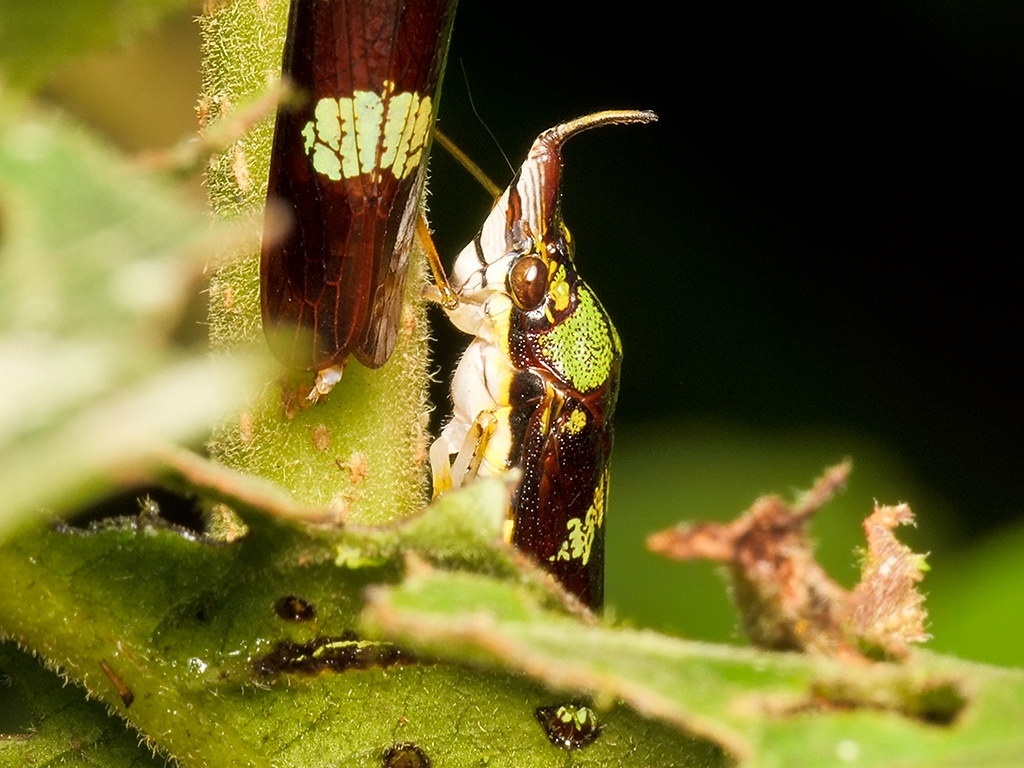}\\
            \vspace{0.08\linewidth}
		\end{minipage}%
	}\hspace{0.018\columnwidth}
	\subfigure[{\scriptsize GT}]{
		\begin{minipage}[t]{0.1\textwidth}
			\centering
			\includegraphics[width=1.15\linewidth,height=0.84\textwidth]{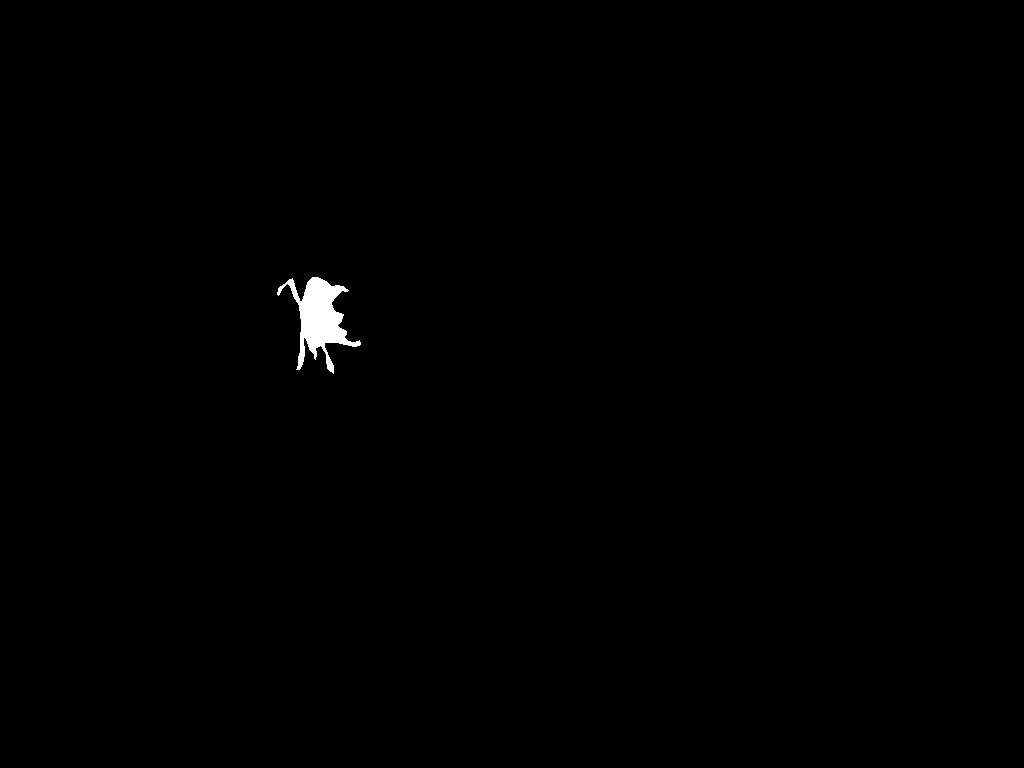}\\
			\vspace{0.01\linewidth}
            \includegraphics[width=1.15\linewidth,height=0.84\textwidth]{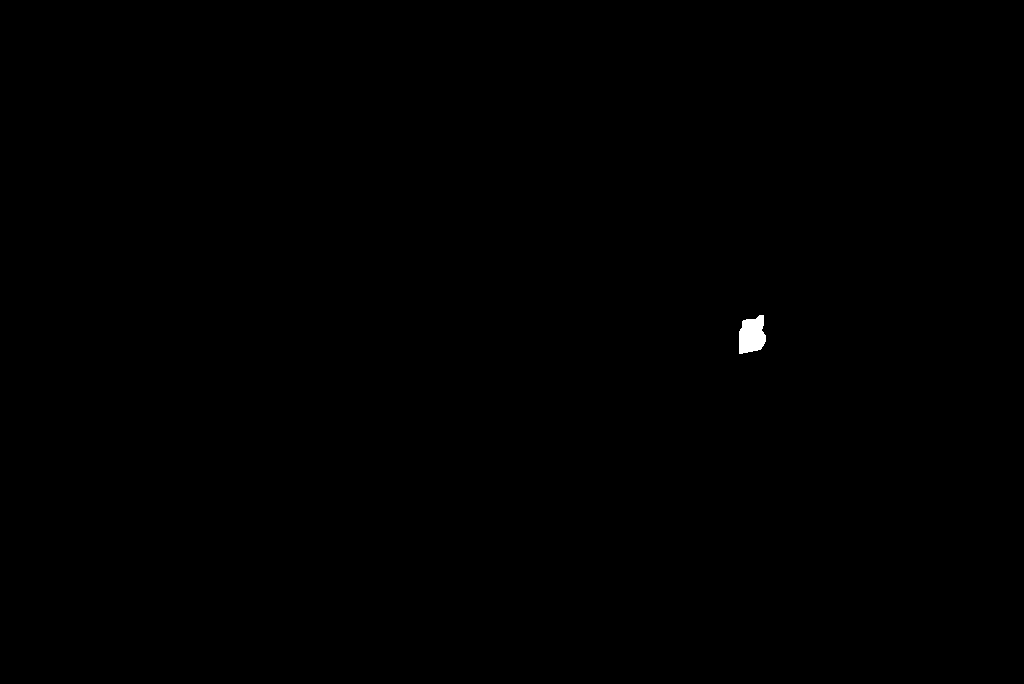}\\
            \vspace{0.01\linewidth}
			\includegraphics[width=1.15\linewidth,height=0.84\textwidth]{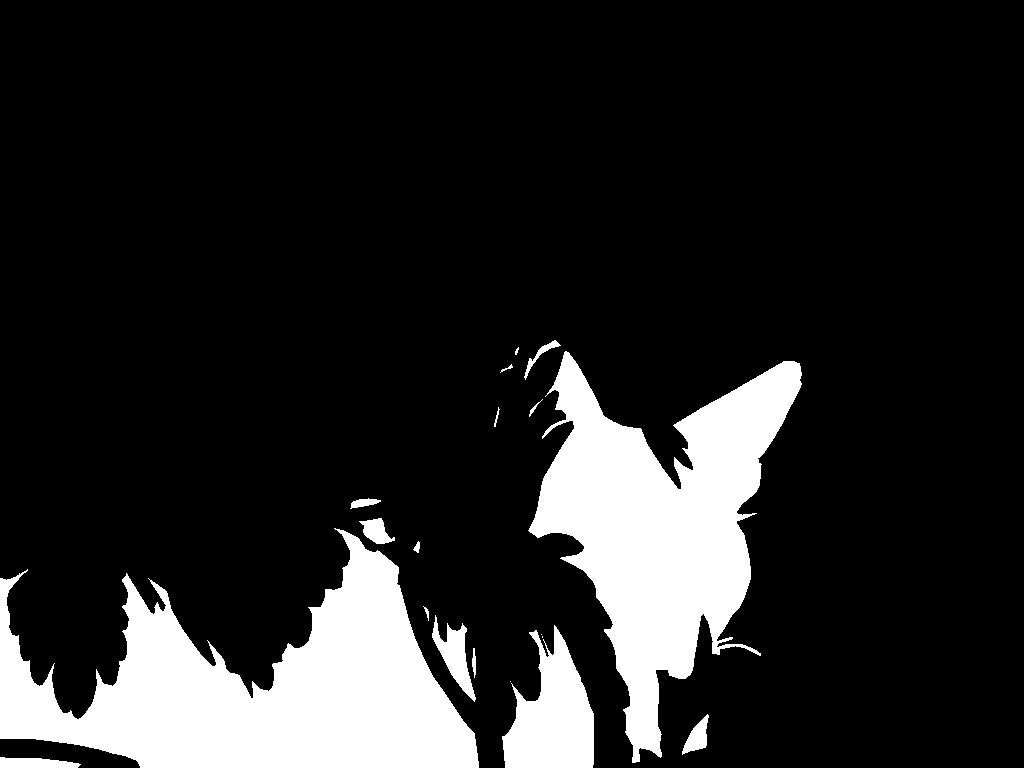}\\
			\vspace{0.01\linewidth}
			\includegraphics[width=1.15\linewidth,height=0.84\textwidth]{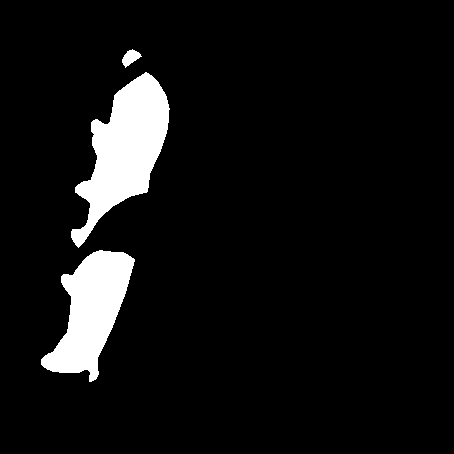}\\
			\vspace{0.01\linewidth}
			\includegraphics[width=1.15\linewidth,height=0.84\textwidth]{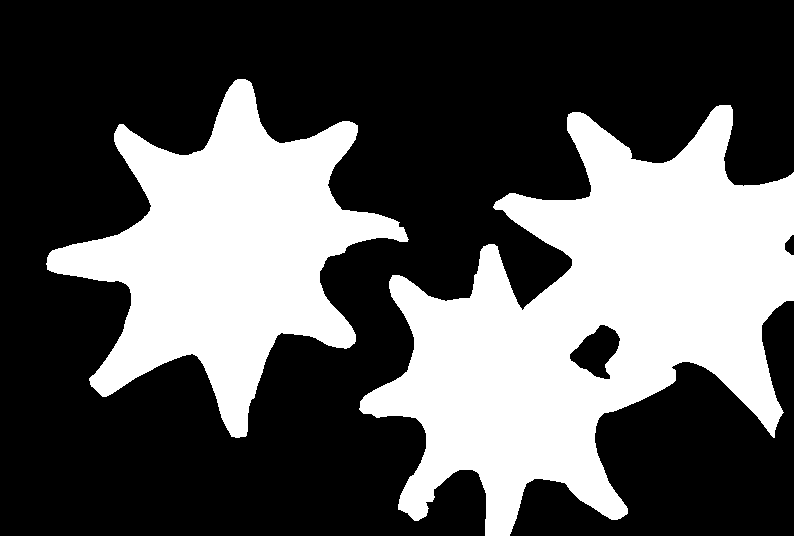}\\
			\vspace{0.01\linewidth}
           \includegraphics[width=1.15\linewidth,height=0.840\textwidth]{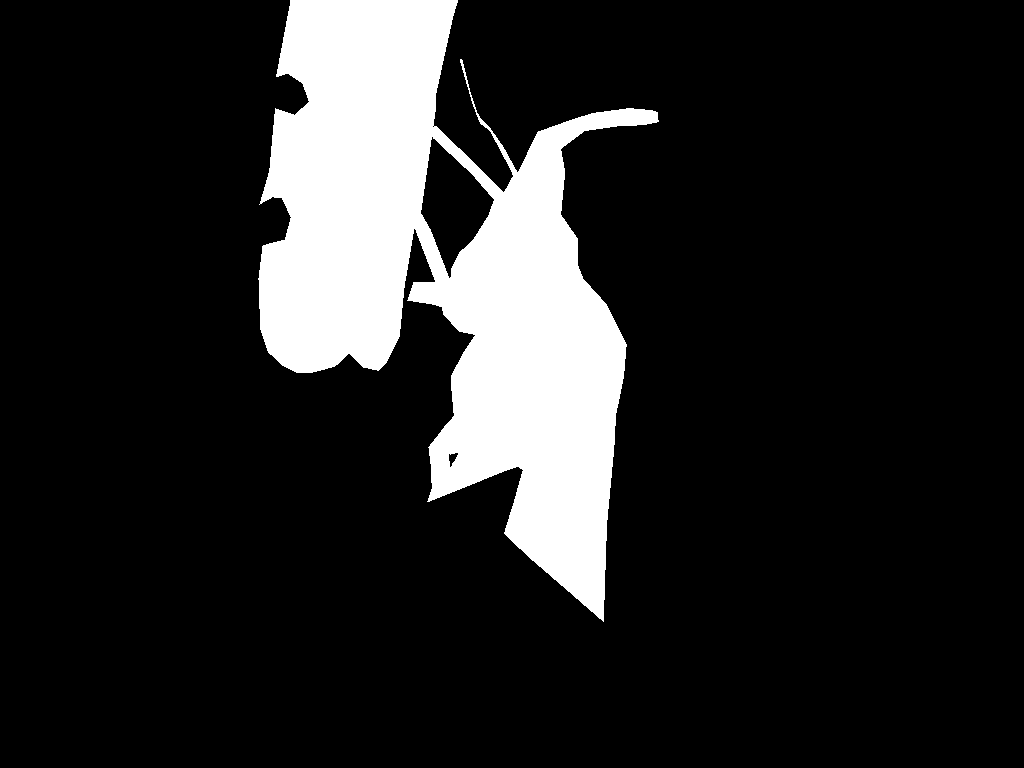}\\
			\vspace{0.08\linewidth}
		\end{minipage}%
	}\hspace{0.018\columnwidth}
	\subfigure[{\scriptsize Ours}]{
		\begin{minipage}[t]{0.1\textwidth}
			\centering
			\includegraphics[width=1.15\linewidth,height=0.84\textwidth]{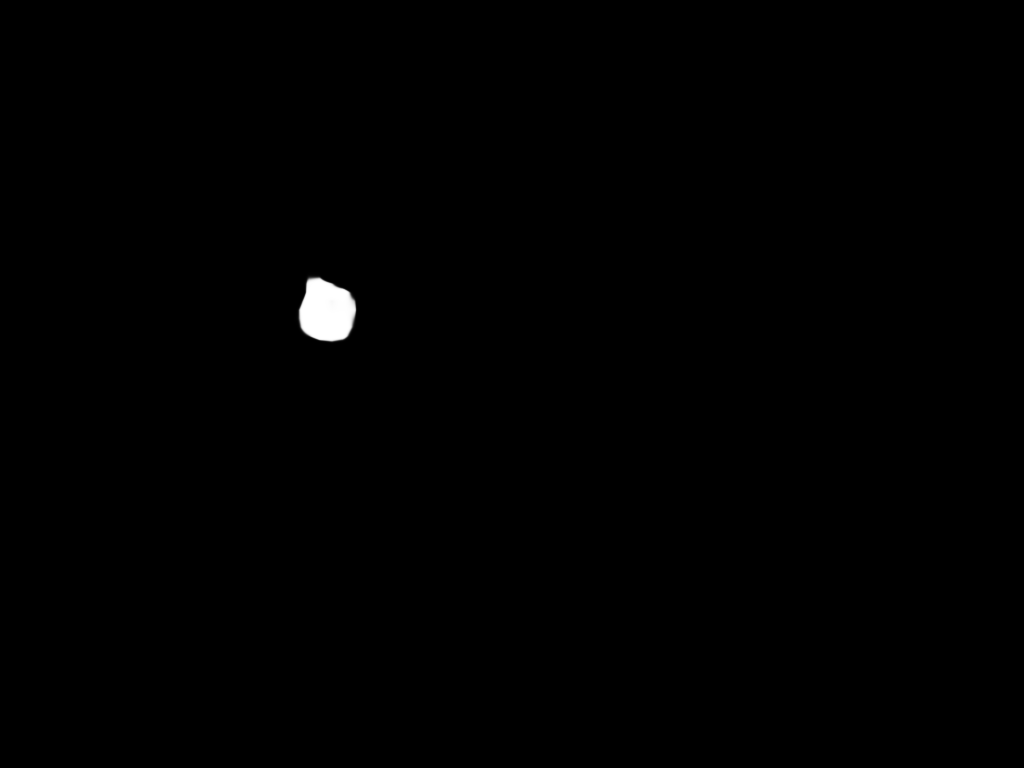}\\
			\vspace{0.01\linewidth}
            \includegraphics[width=1.15\linewidth,height=0.84\textwidth]{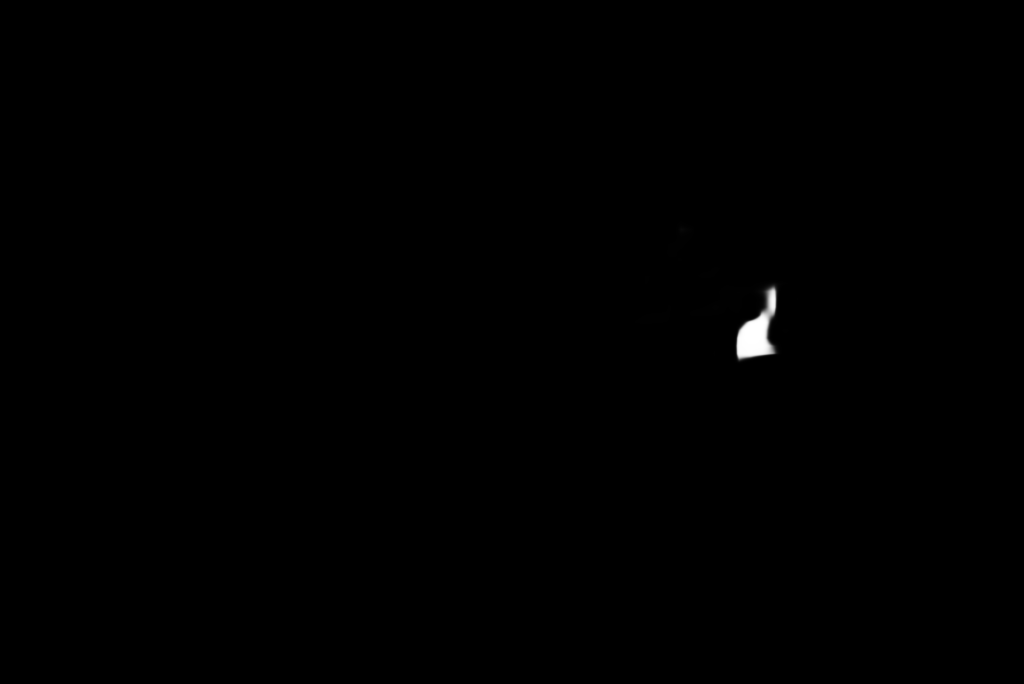}\\
            \vspace{0.01\linewidth}
			\includegraphics[width=1.15\linewidth,height=0.84\textwidth]{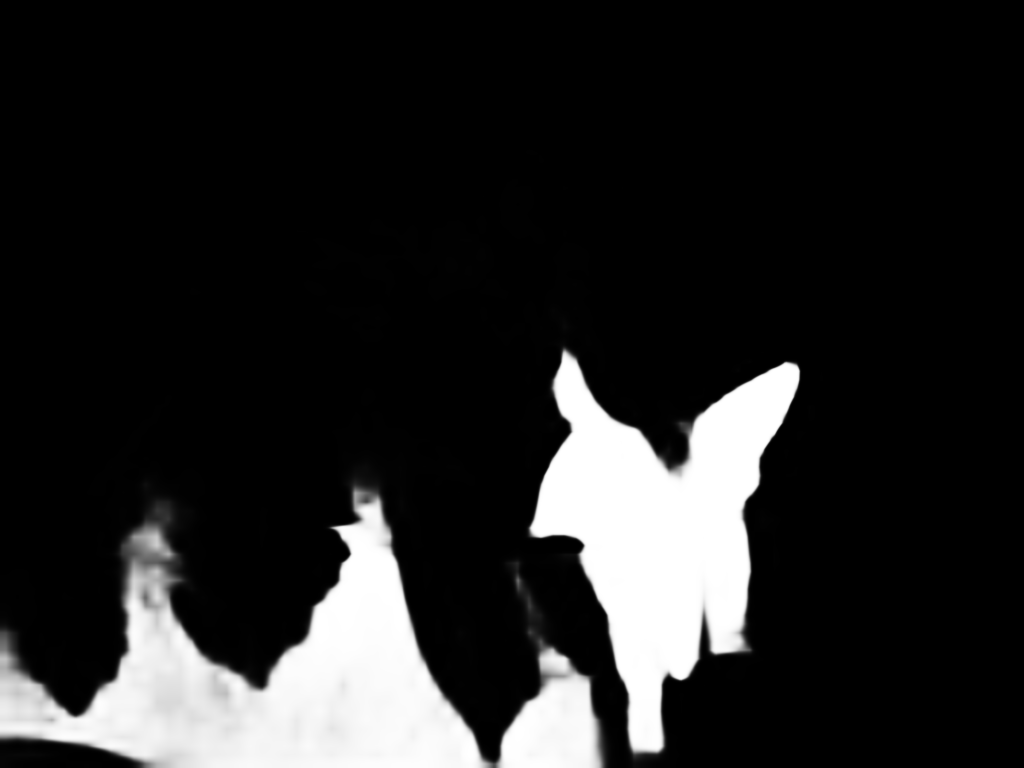}\\
			\vspace{0.01\linewidth}
			\includegraphics[width=1.15\linewidth,height=0.84\textwidth]{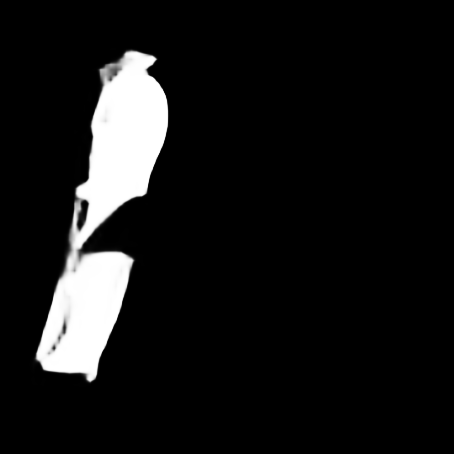}\\
			\vspace{0.01\linewidth}
			\includegraphics[width=1.15\linewidth,height=0.840\textwidth]{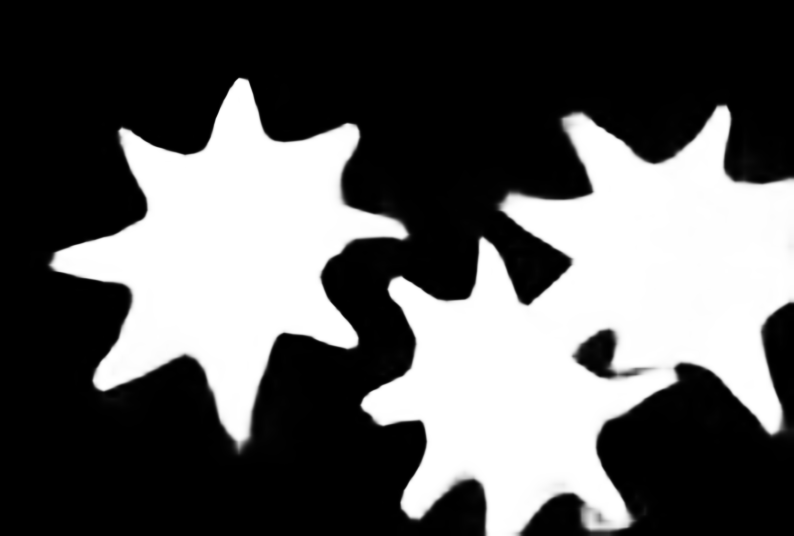}\\
			\vspace{0.01\linewidth}

            \includegraphics[width=1.15\linewidth,height=0.840\textwidth]{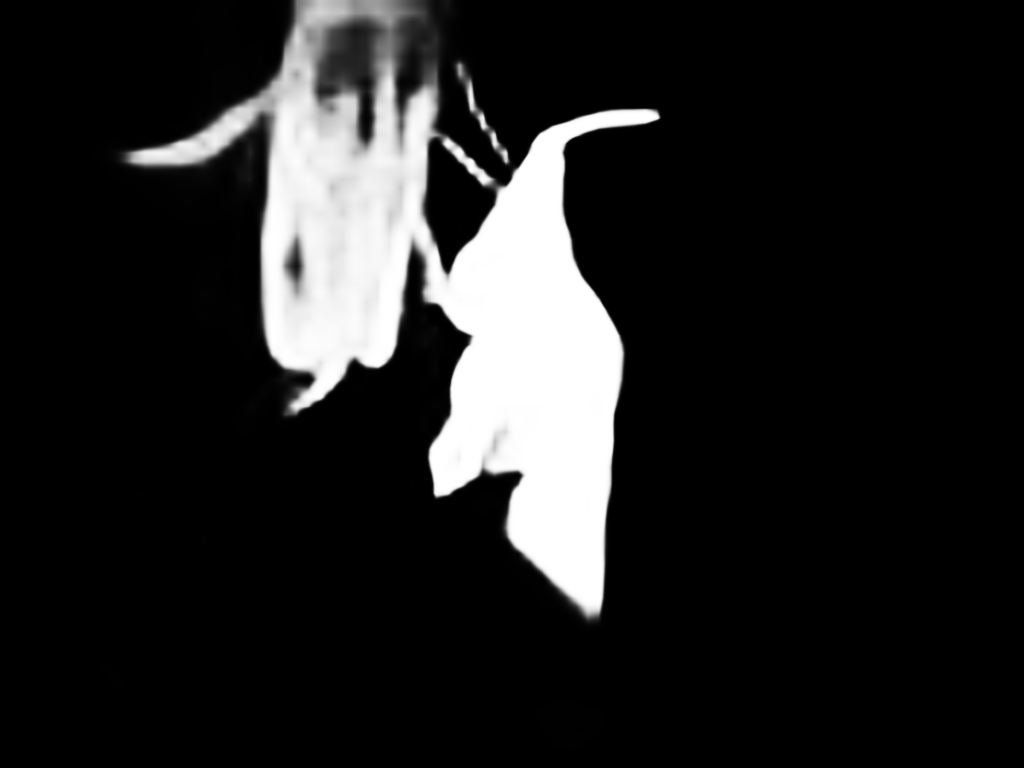}\\
        	\vspace{0.08\linewidth}
		\end{minipage}%
	}\hspace{0.018\columnwidth}
	\subfigure[{\scriptsize FSPNet}]{
		\begin{minipage}[t]{0.1\textwidth}
			\centering
			\includegraphics[width=1.15\linewidth,height=0.84\textwidth]{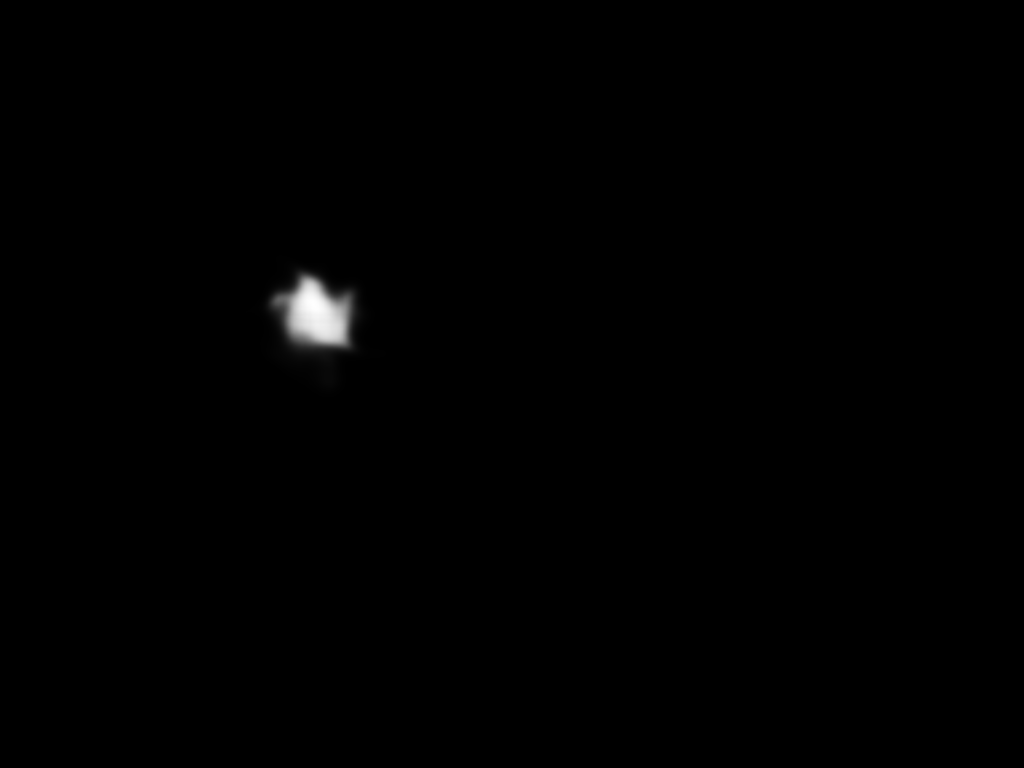}\\
			\vspace{0.01\linewidth}
            \includegraphics[width=1.15\linewidth,height=0.84\textwidth]{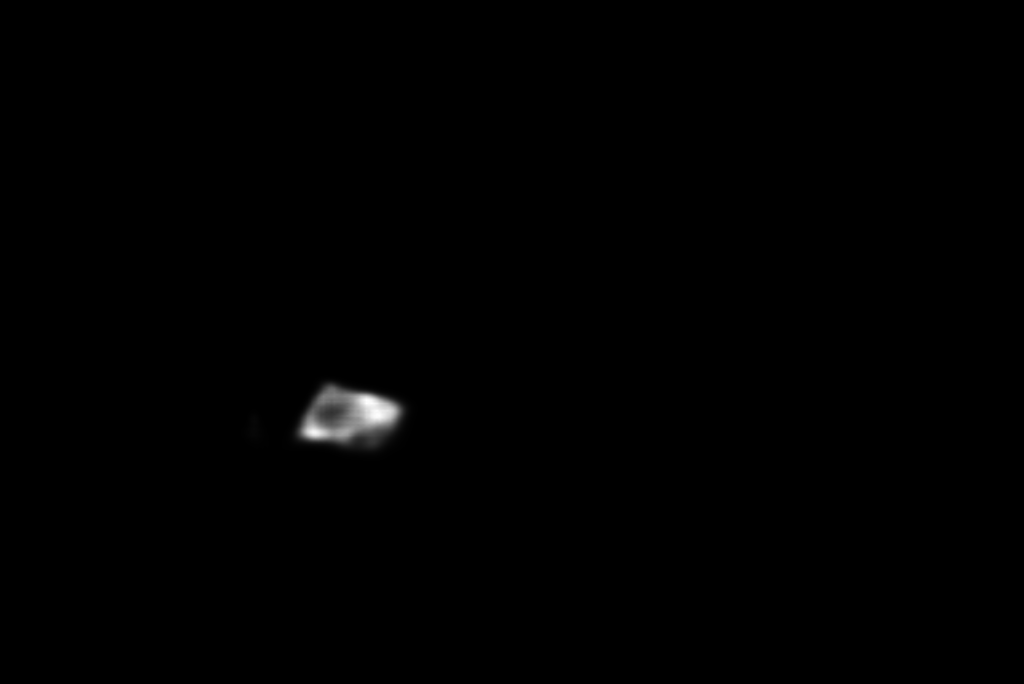}\\
            \vspace{0.01\linewidth}
			\includegraphics[width=1.15\linewidth,height=0.84\textwidth]{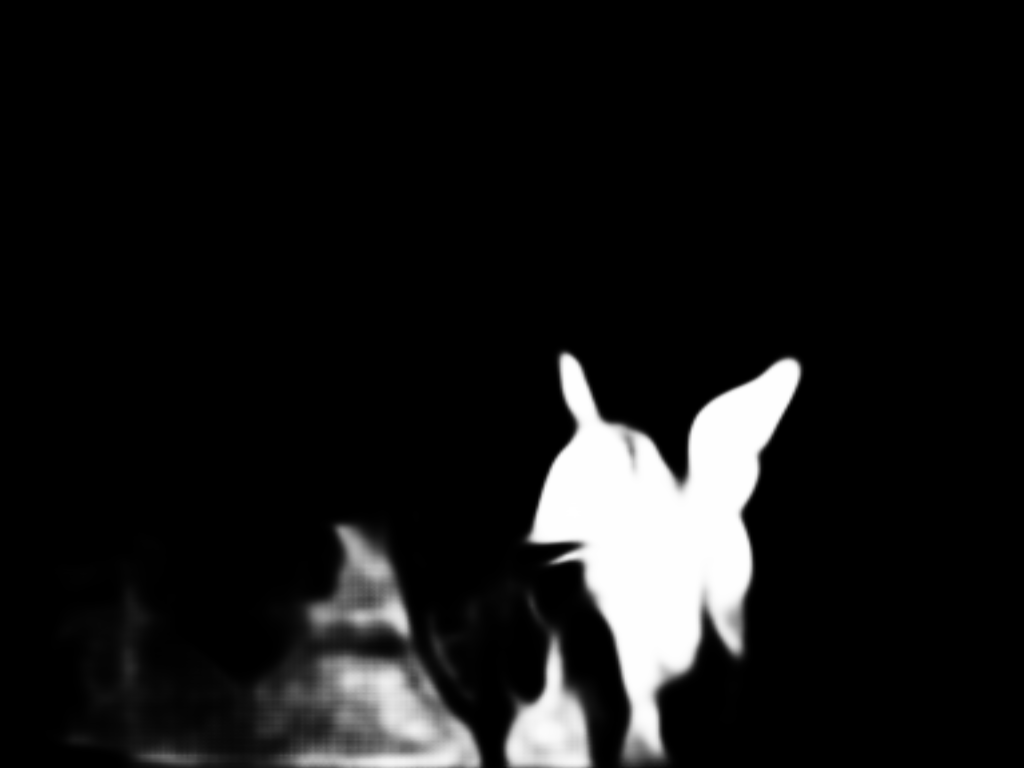}\\
			\vspace{0.01\linewidth}
			\includegraphics[width=1.15\linewidth,height=0.84\textwidth]{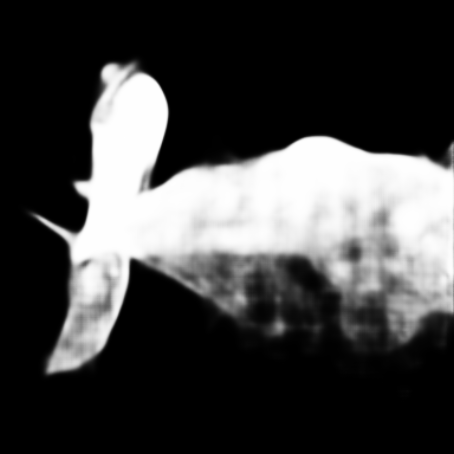}\\
			\vspace{0.01\linewidth}
			\includegraphics[width=1.15\linewidth,height=0.84\textwidth]{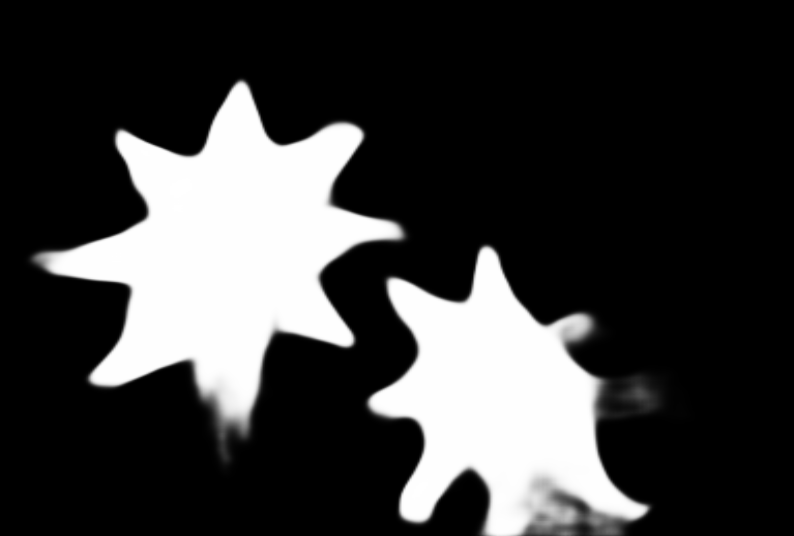}\\
			\vspace{0.01\linewidth}

            \includegraphics[width=1.15\linewidth,height=0.840\textwidth]{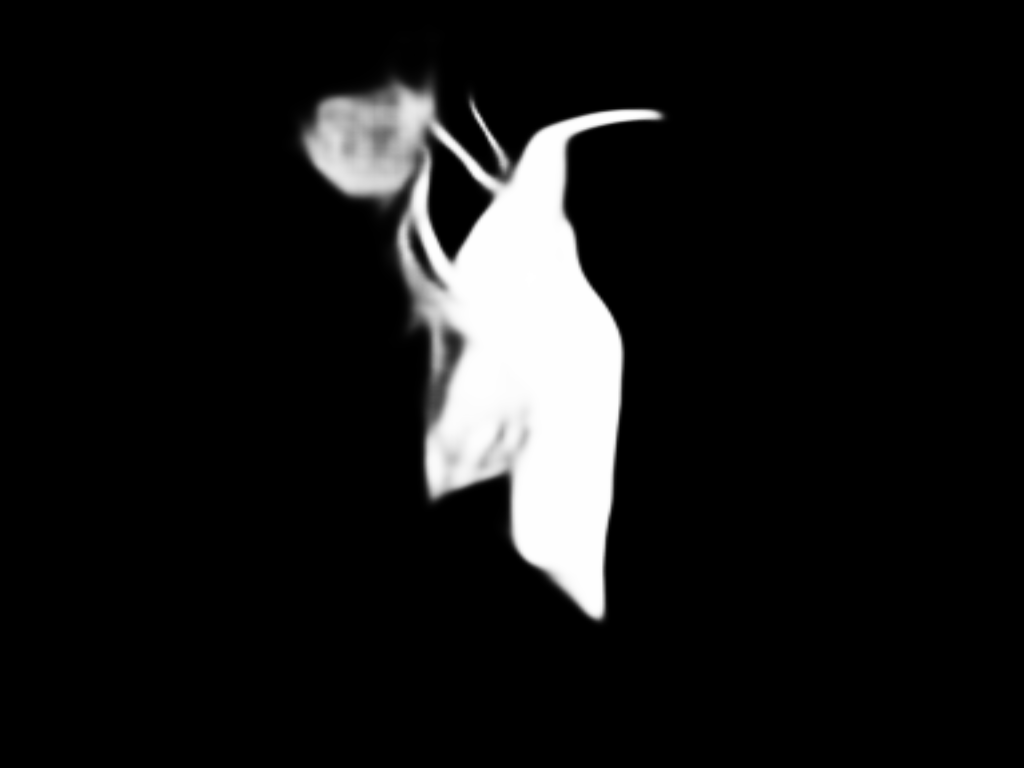}\\
			\vspace{0.08\linewidth}
		\end{minipage}%
	}\hspace{0.018\columnwidth}
	\subfigure[{\scriptsize FEDER}]{
		\begin{minipage}[t]{0.1\textwidth}
			\centering
			\includegraphics[width=1.15\linewidth,height=0.84\textwidth]{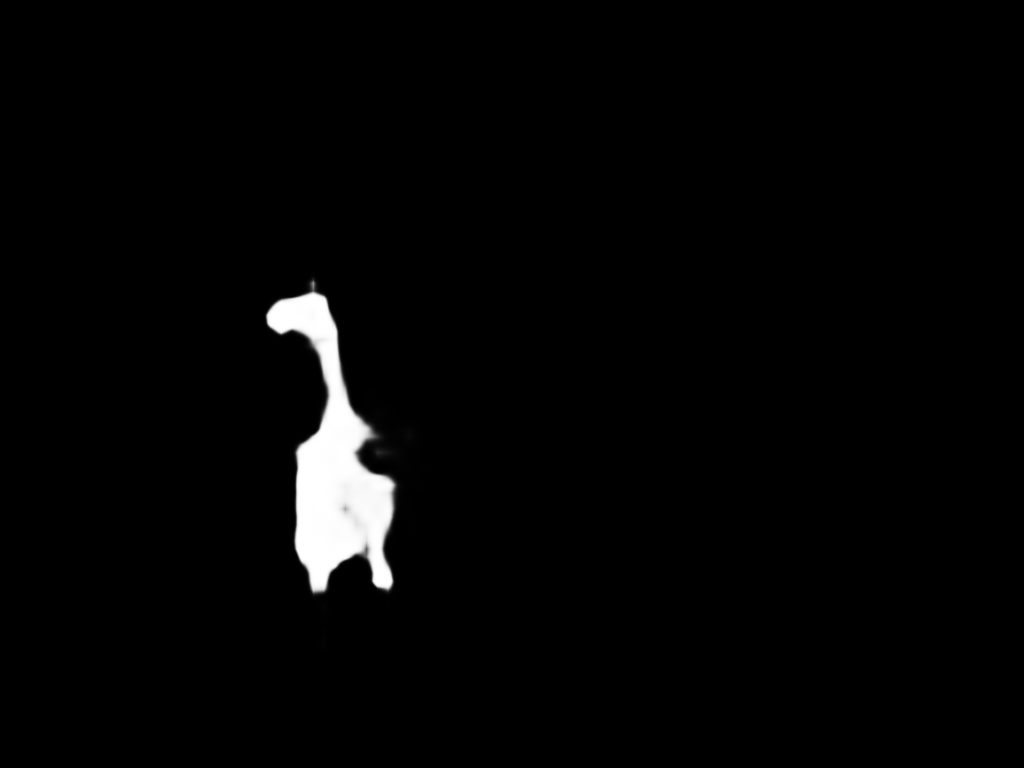}\\
			\vspace{0.01\linewidth}
            \includegraphics[width=1.15\linewidth,height=0.84\textwidth]{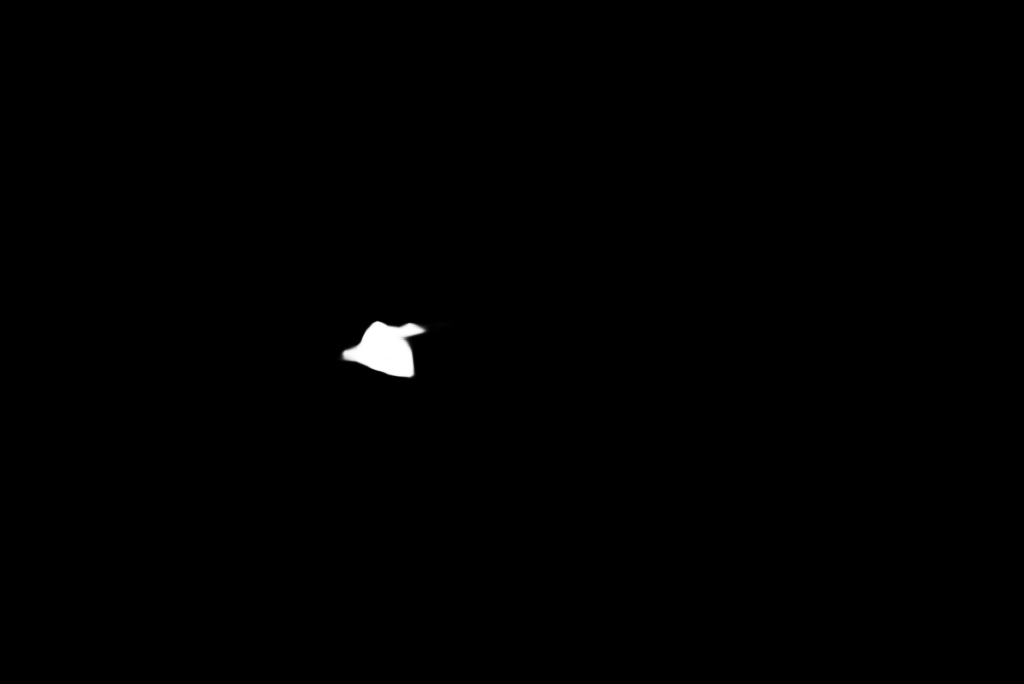}\\
            \vspace{0.01\linewidth}
			\includegraphics[width=1.15\linewidth,height=0.84\textwidth]{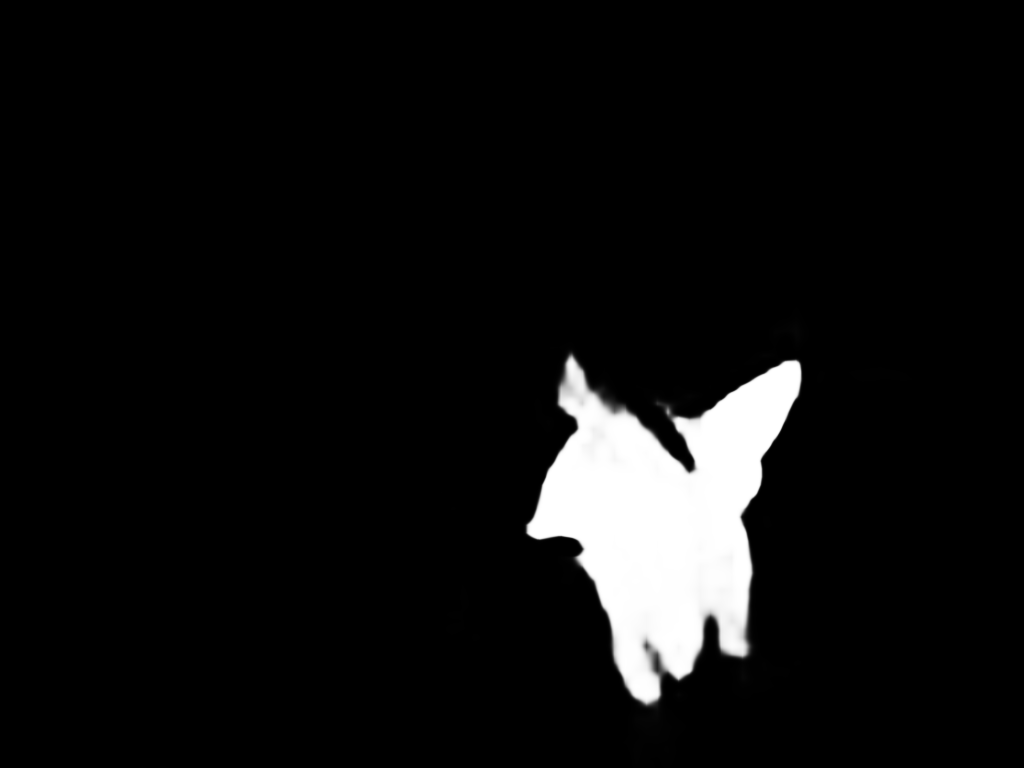}\\
			\vspace{0.01\linewidth}
			\includegraphics[width=1.15\linewidth,height=0.84\textwidth]{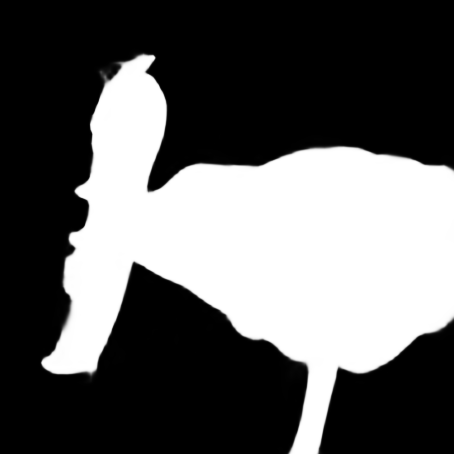}\\
			\vspace{0.01\linewidth}
			\includegraphics[width=1.15\linewidth,height=0.84\textwidth]{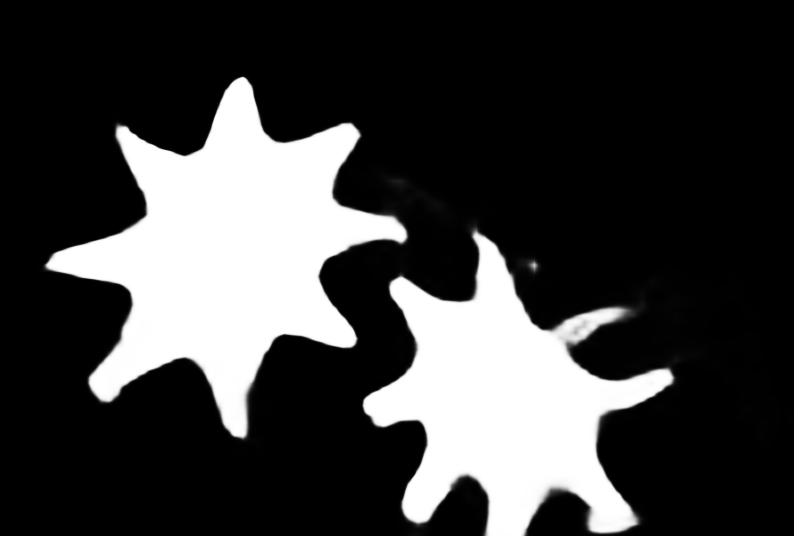}\\
			\vspace{0.01\linewidth}

            \includegraphics[width=1.15\linewidth,height=0.840\textwidth]{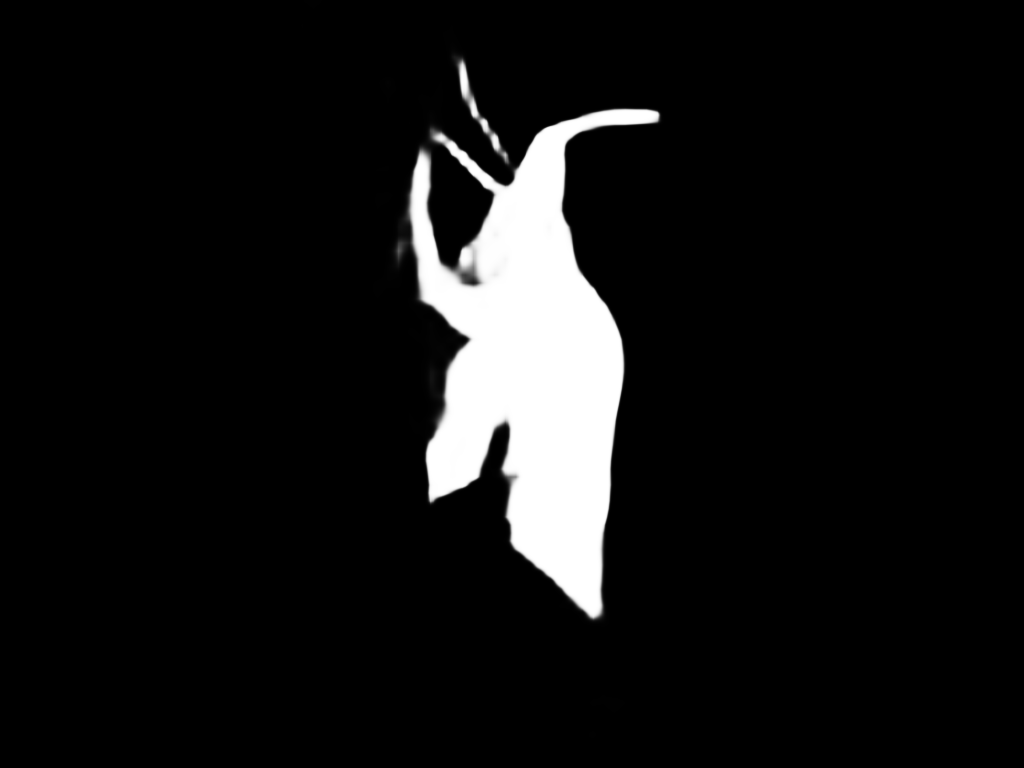}\\
			\vspace{0.08\linewidth}
		\end{minipage}%
	}\hspace{0.018\columnwidth}
    \subfigure[{\scriptsize FDCOD}]{
		\begin{minipage}[t]{0.1\textwidth}
			\centering
			\includegraphics[width=1.15\linewidth,height=0.84\textwidth]{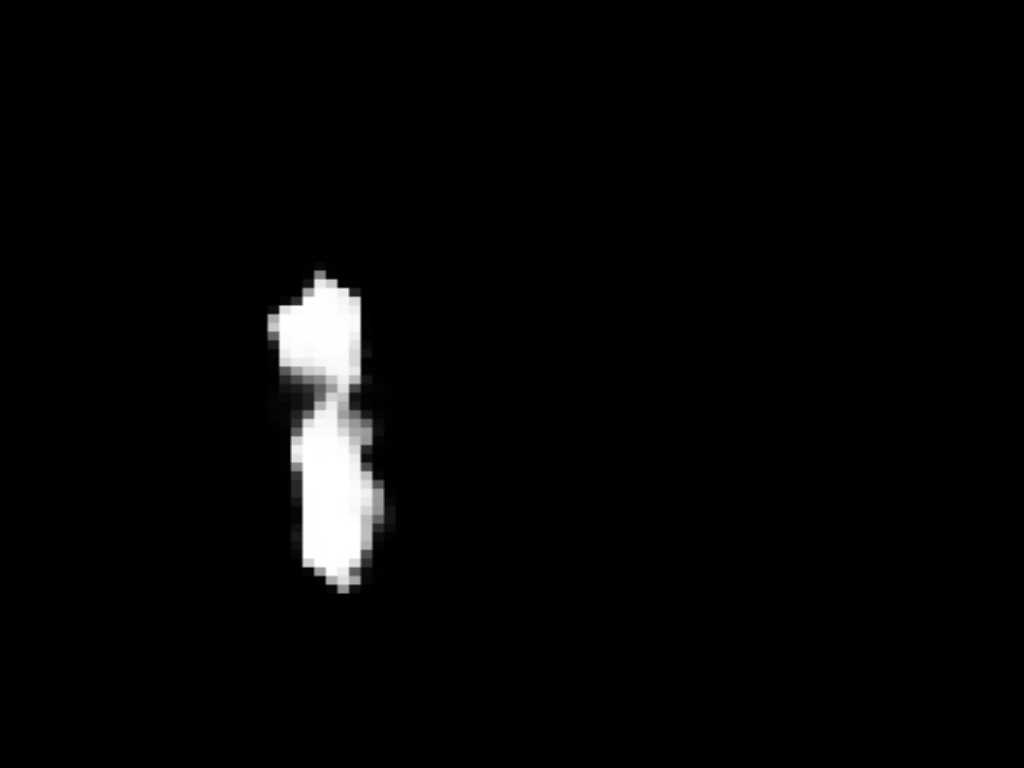}\\
			\vspace{0.01\linewidth}
            \includegraphics[width=1.15\linewidth,height=0.84\textwidth]{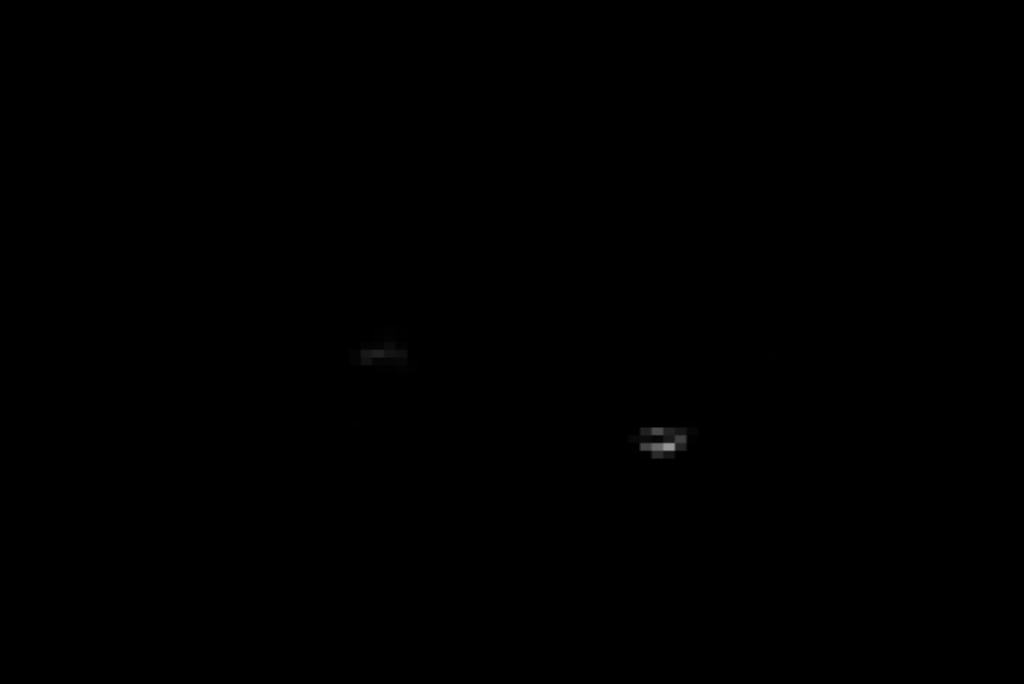}\\
            \vspace{0.01\linewidth}
			\includegraphics[width=1.15\linewidth,height=0.84\textwidth]{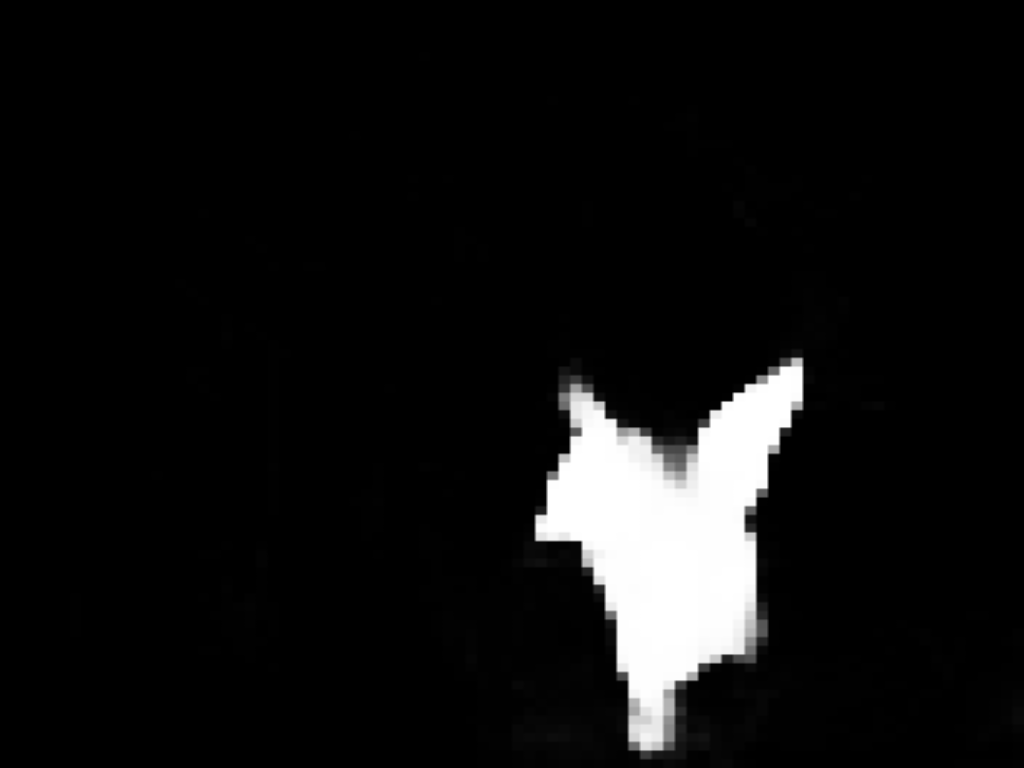}\\
			\vspace{0.01\linewidth}
			\includegraphics[width=1.15\linewidth,height=0.84\textwidth]{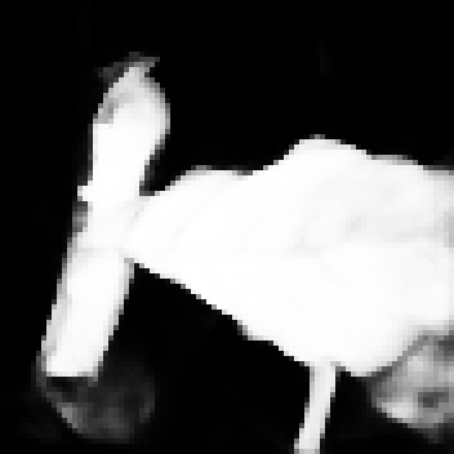}\\
			\vspace{0.01\linewidth}
			\includegraphics[width=1.15\linewidth,height=0.84\textwidth]{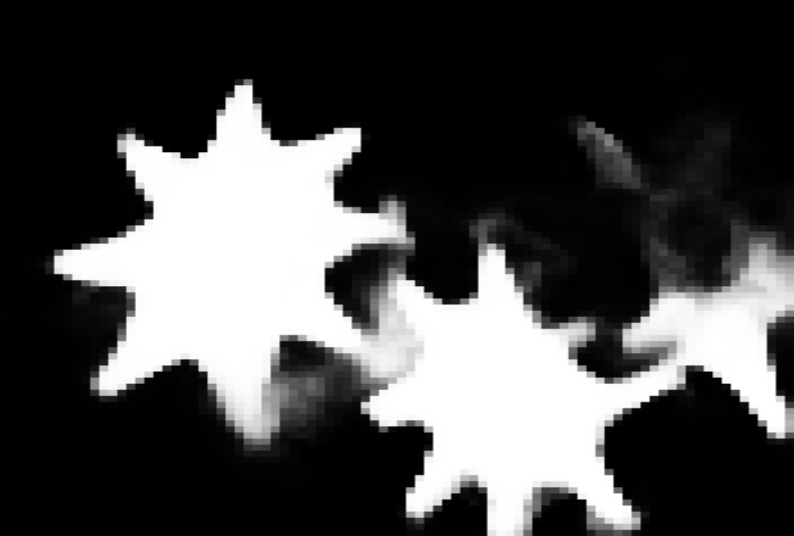}\\
			\vspace{0.01\linewidth}
            \includegraphics[width=1.15\linewidth,height=0.840\textwidth]{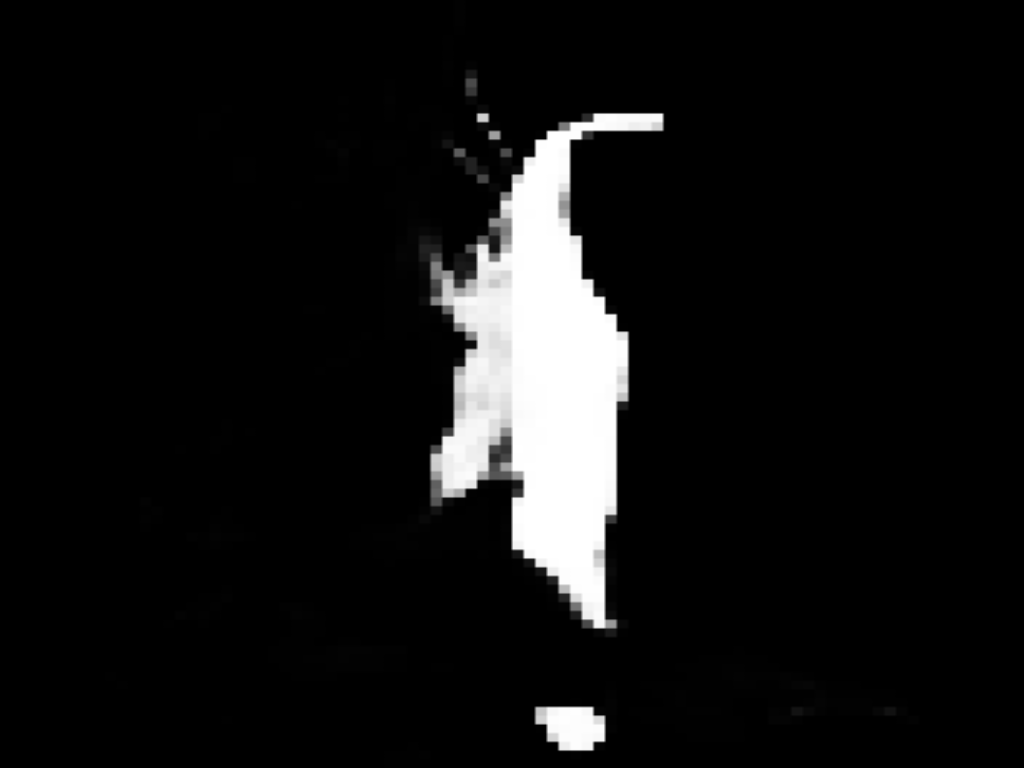}\\
			\vspace{0.08\linewidth}
		\end{minipage}%
	}\hspace{0.018\columnwidth}
    \subfigure[{\scriptsize DGNet}]{
		\begin{minipage}[t]{0.1\textwidth}
			\centering
			\includegraphics[width=1.15\linewidth,height=0.84\textwidth]{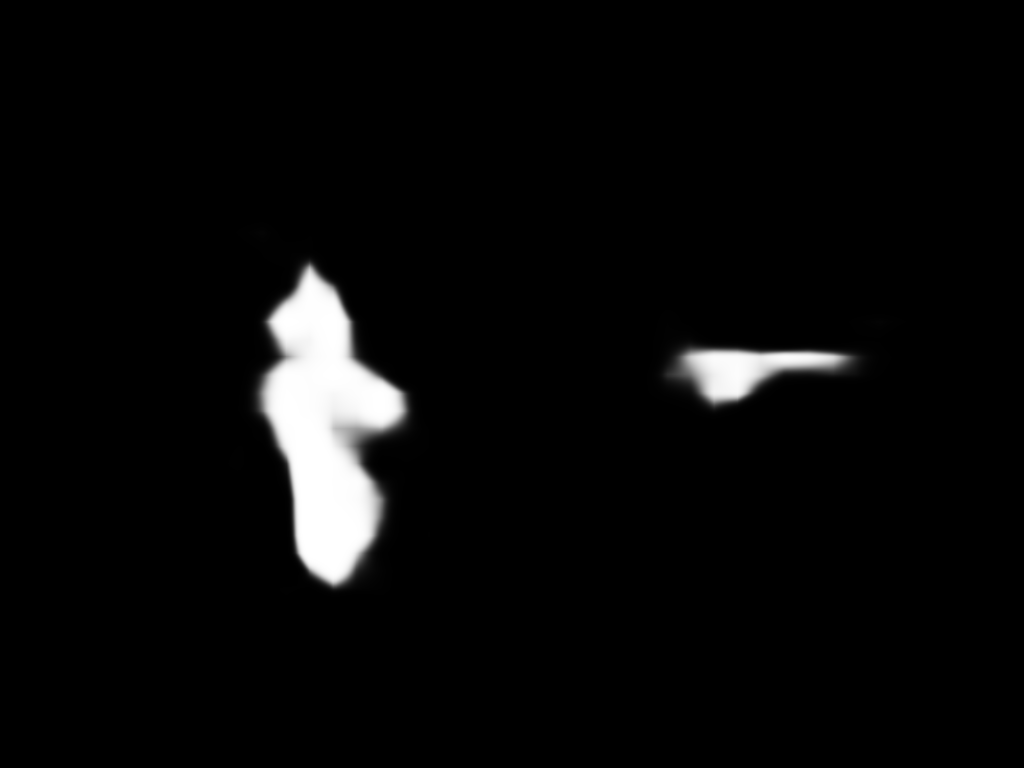}\\
			\vspace{0.01\linewidth}
            \includegraphics[width=1.15\linewidth,height=0.84\textwidth]{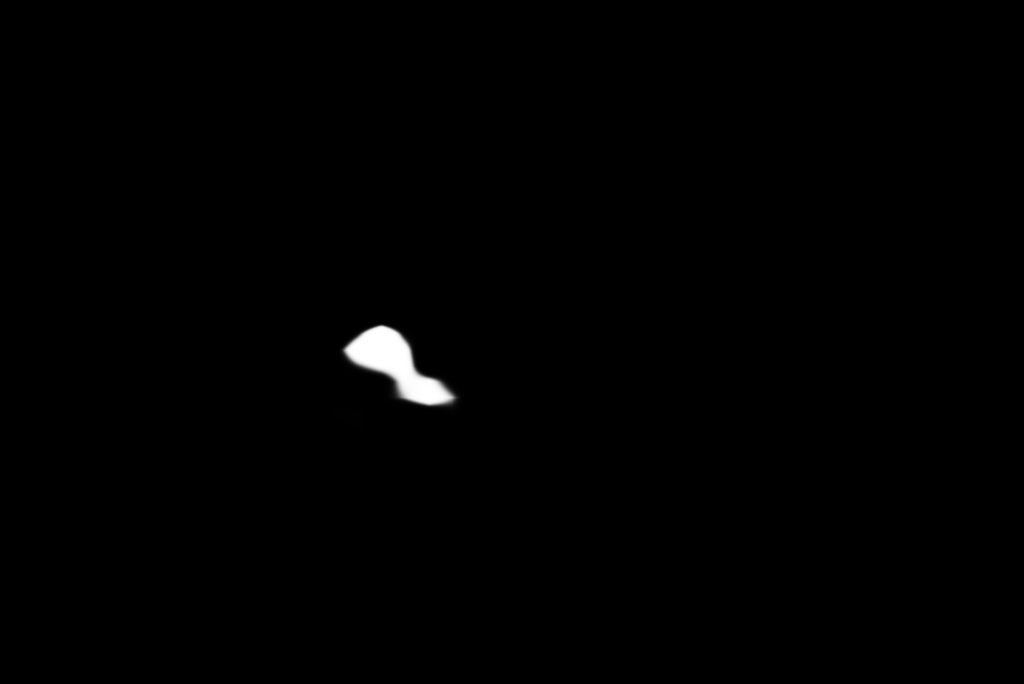}\\
            \vspace{0.01\linewidth}
			\includegraphics[width=1.15\linewidth,height=0.84\textwidth]{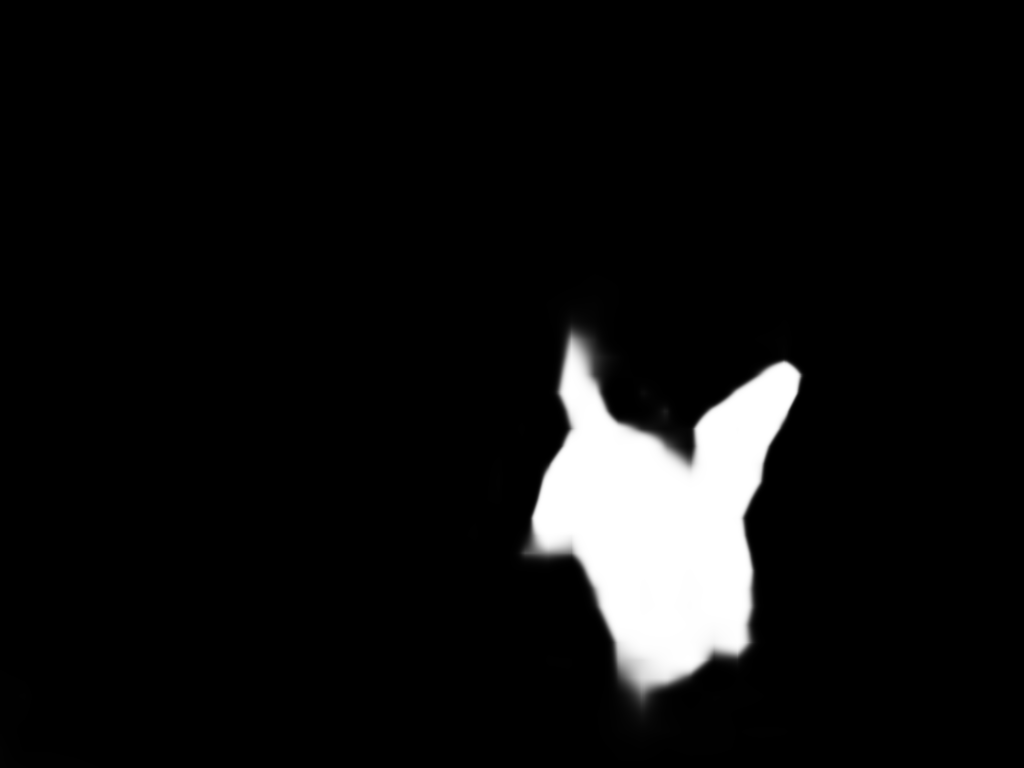}\\
			\vspace{0.01\linewidth}
			\includegraphics[width=1.15\linewidth,height=0.84\textwidth]{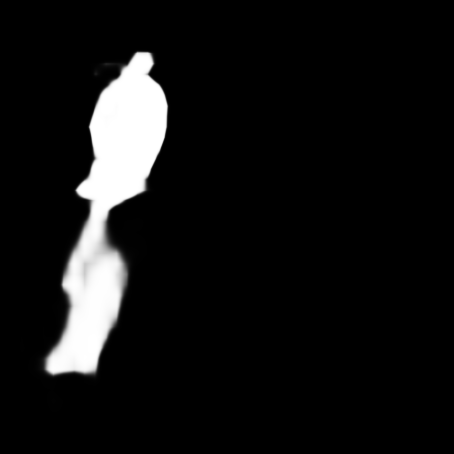}\\
			\vspace{0.01\linewidth}
			\includegraphics[width=1.15\linewidth,height=0.84\textwidth]{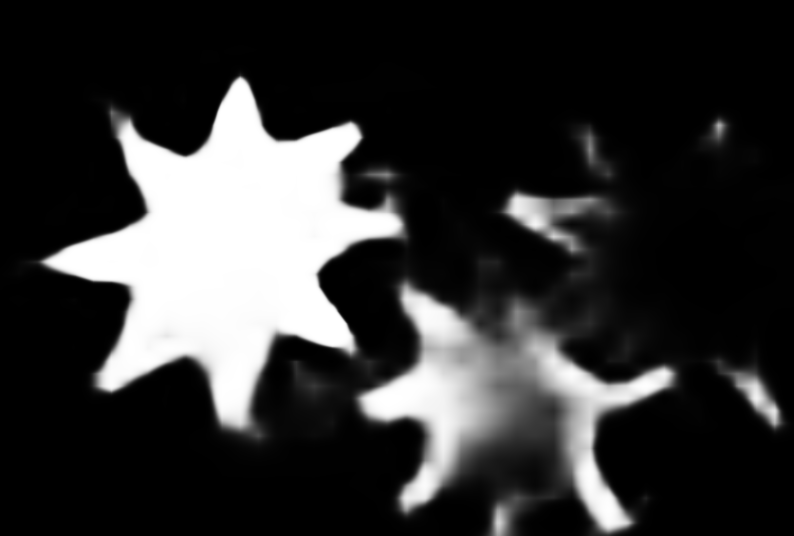}\\
			\vspace{0.01\linewidth}

            \includegraphics[width=1.15\linewidth,height=0.840\textwidth]{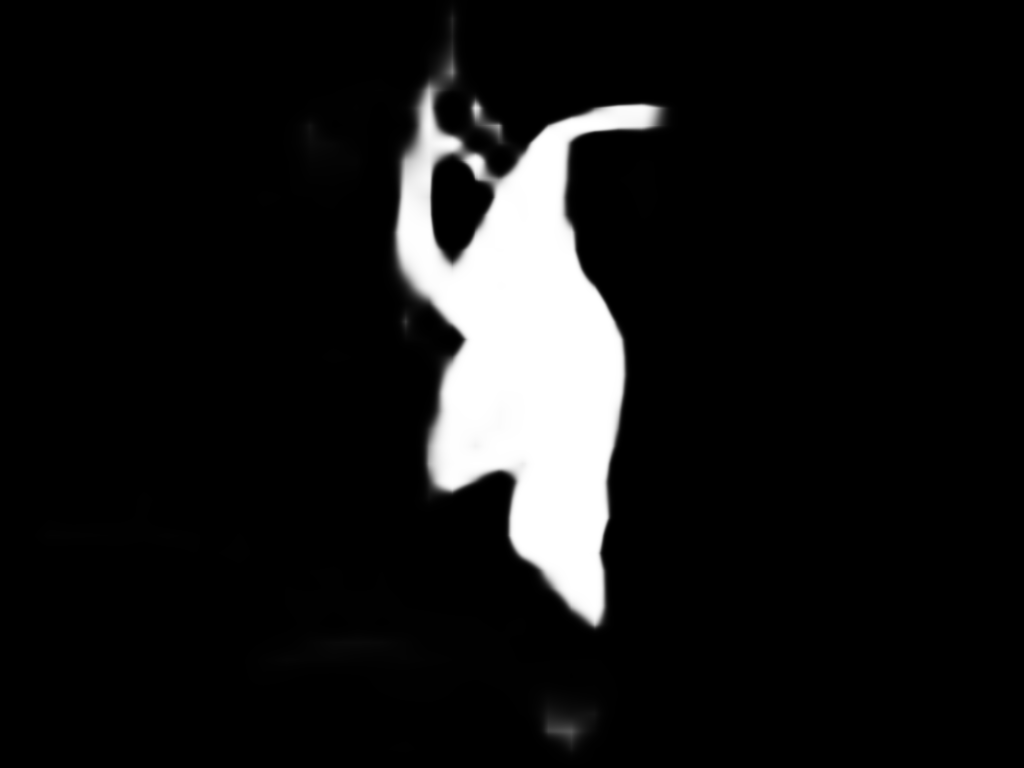}\\
			\vspace{0.08\linewidth}
		\end{minipage}%
	}\hspace{0.018\columnwidth}
	\subfigure[{\scriptsize SINet}]{
		\begin{minipage}[t]{0.1\textwidth}
			\centering
			\includegraphics[width=1.15\linewidth,height=0.84\textwidth]{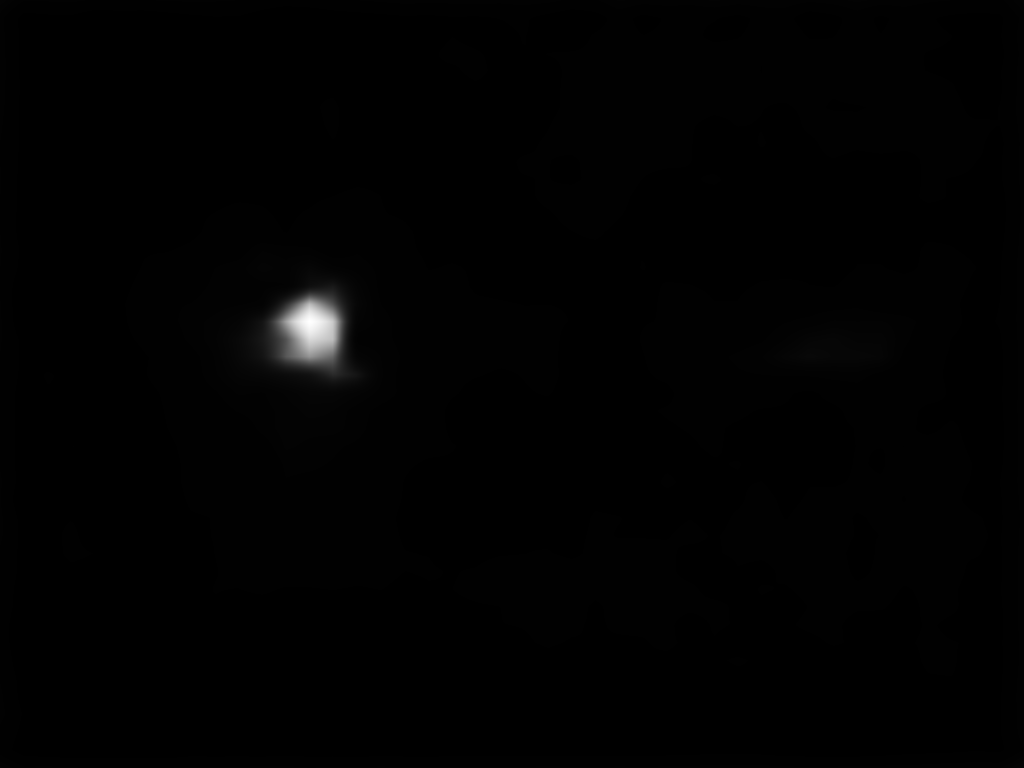}\\
			\vspace{0.01\linewidth}
            \includegraphics[width=1.15\linewidth,height=0.84\textwidth]{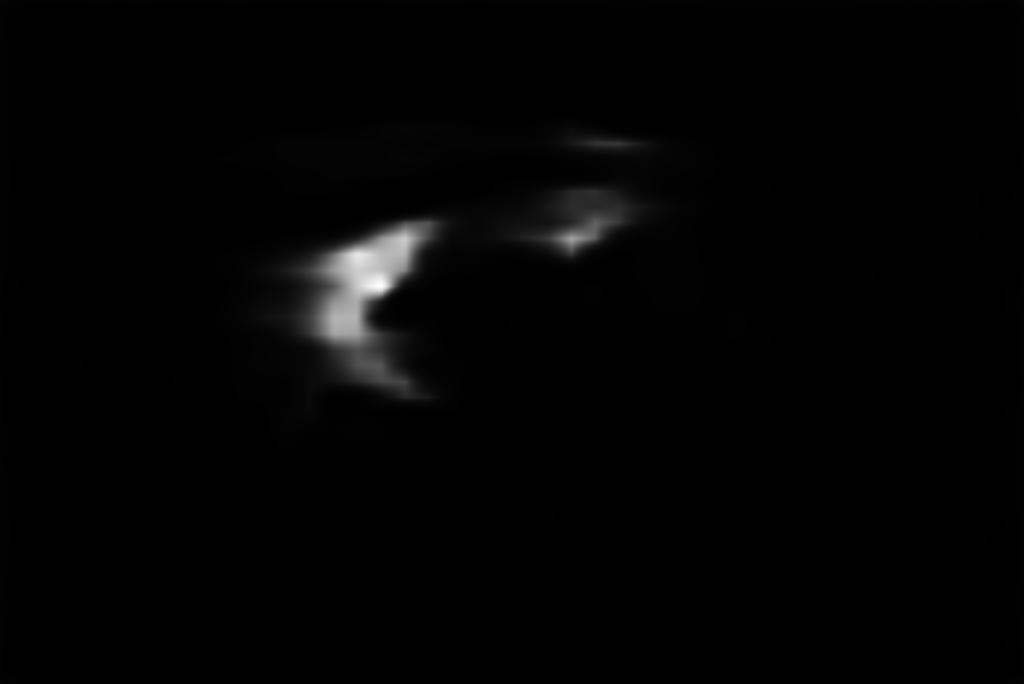}\\
            \vspace{0.01\linewidth}
			\includegraphics[width=1.15\linewidth,height=0.84\textwidth]{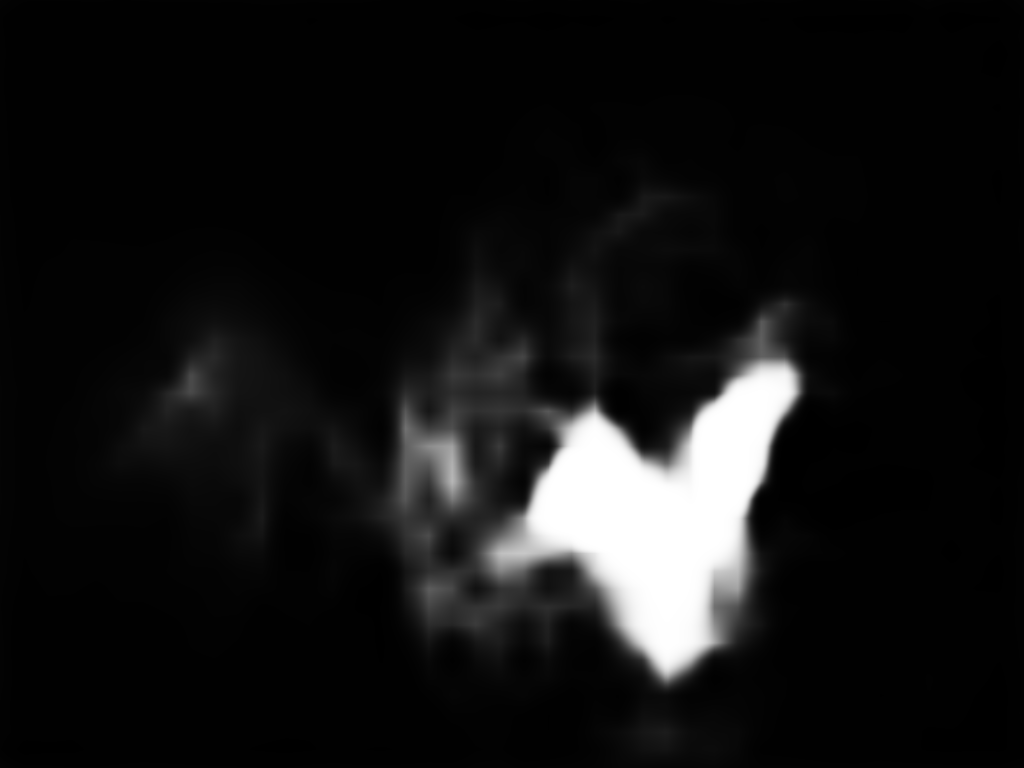}\\
			\vspace{0.01\linewidth}
			\includegraphics[width=1.15\linewidth,height=0.84\textwidth]{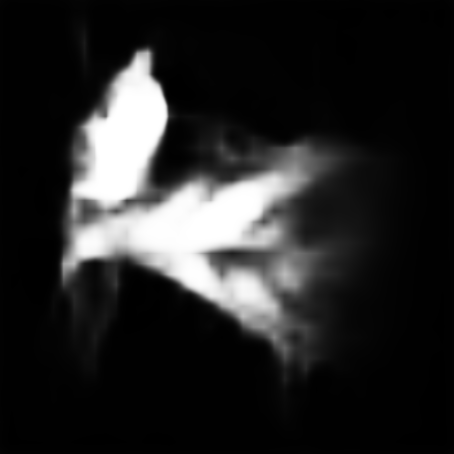}\\
			\vspace{0.01\linewidth}
			\includegraphics[width=1.15\linewidth,height=0.840\textwidth]{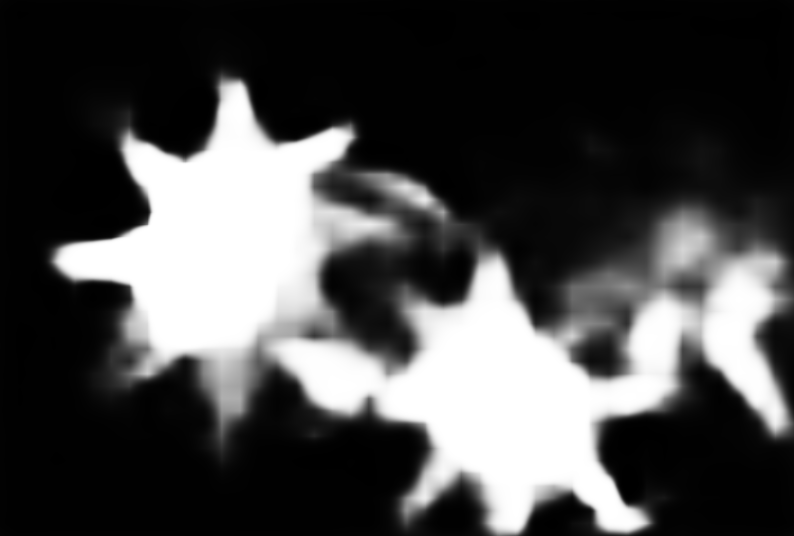}\\
			\vspace{0.01\linewidth}
            \includegraphics[width=1.15\linewidth,height=0.840\textwidth]{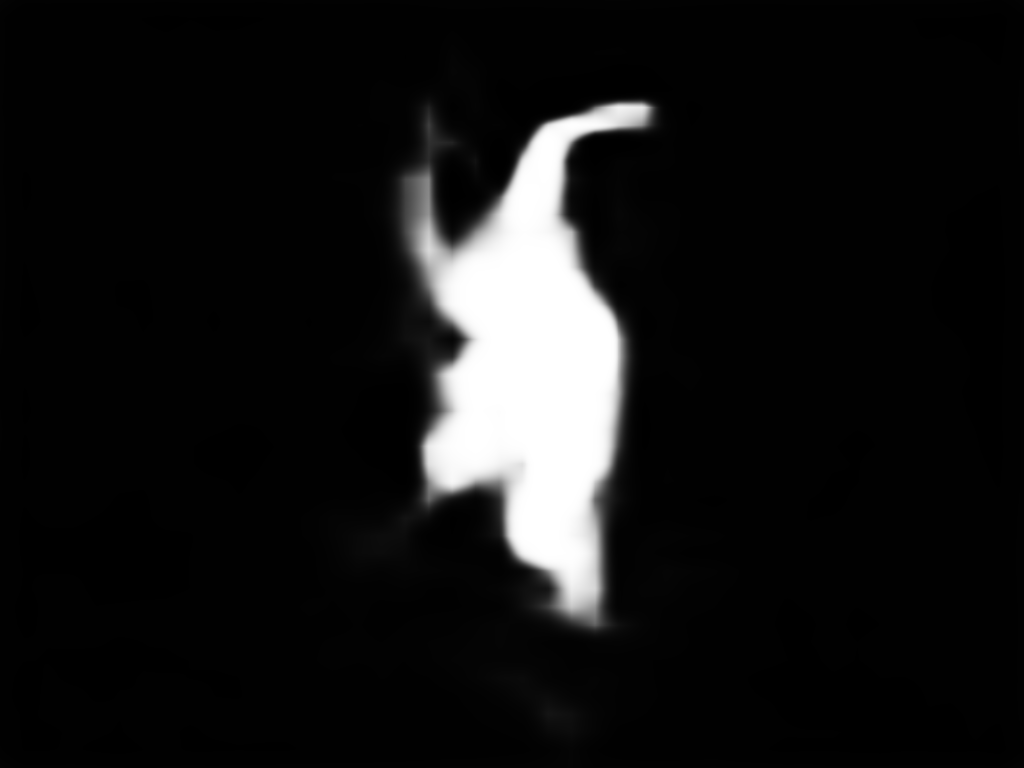}\\
			\vspace{0.08\linewidth}
		\end{minipage}%
	}\hspace{0.018\columnwidth}
	\centering
    
	\caption{\textbf{Qualitative comparison of our proposed method and other representative COD methods.} Our method provides better performance than all competitors for camouflaged object segmentation in various complex scenes. 
 }
    \label{fig:GSNet_Qualitative}
\end{figure*}

\section{Experiments}
\subsection{Experimental Setup}
\noindent\textbf{Datasets.} 
We conduct experiments on three widely used benchmark datasets of COD task, \textit{i.e.}, CAMO~\cite{le2019anabranch}, 
COD10K \cite{fan2020camouflaged} and NC4K~\cite{lv2021simultaneously}. 
In particular, CAMO, covering eight categories, contains 1,250 camouflaged images and 1,250 non-camouflaged images. 
COD10K consists of 5,066 camouflaged, 1,934 non-camouflaged, and 3,000 background images, and it is currently the largest dataset which covers 10 superclasses and 78 subclasses.
NC4K is a newly published dataset which has a total of 4,121 camouflaged images.
Following standard practice of COD tasks, 
we use 3,040  images from COD10K and 1,000 images from CAMO as the training set and the remaining data as the test set.

\vspace{10pt}
\noindent
\textbf{Evaluation Metrics.} 
According to the standard evaluation protocol of COD, we employ the five common metrics to evaluate our model, \textit{i.e.}, 
structure-measure ($S_{\alpha}$) \cite{fan2017structure}, weighted F-measure $F_{\beta}^{\omega}$ \cite{margolin2014evaluate}, mean F-measure ($F_{m}$), mean E-measure ($E_{m}$) \cite{fan2021cognitive} and mean absolute error ($MAE$).

\vspace{10pt}
\noindent
\textbf{Implementation Details.} 
All experiments are implemented with the PyTorch toolbox. 
The visual backbone we adopted is the EfficientNet-B4 \cite{tan2019efficientnet} pretrained on ImageNet unless otherwise stated. 
We use Adam~\cite{kingma2015adam} as our model optimizer with a learning rate of 1e-4, which is adjusted according to cosine annealing strategy \cite{loshchilov2016sgdr} with a period of 40 epochs and a minimum learning rate of 1e-5. 
We train our AGLNet for 100 epochs with a batch size of 8, which takes about 9 hours on an NVIDIA GeForce RTX 3090 GPU.
During the training and inference, the input images are resized to 704$\times$704 via bilinear interpolation and augmented by random flipping, cropping, and color jittering.

\vspace{10pt}
\noindent
\textbf{Competitors.} 
Our AGLNet is compared with 20 recent  state-of-the-art methods, including SINet \cite{fan2020camouflaged}, PFNet \cite{mei2021camouflaged}, LSR \cite{lv2021simultaneously}, C$^{2}$FNet \cite{sun2021context}, MGL \cite{zhai2021mutual}, UGTR \cite{yang2021uncertainty}, UJSC \cite{li2021uncertainty}, SINet-V2 \cite{fan2021concealed}, R-MGL\_v2 \cite{zhai2022mgl}, BSANet \cite{zhu2022can}, FAPNet \cite{zhou2022feature}, BGNet \cite{sun2022boundary}, SegMaR \cite{jia2022segment}, FDCOD \cite{zhong2022detecting}, ZoomNet \cite{pang2022zoom}, DGNet \cite{ji2022gradient}, FEDER \cite{he2023camouflaged}, PopNet \cite{wu2023source}, HitNet \cite{hu2023high}, FSPNet \cite{huang2023feature}. For a fair comparison, all results are either provided by the authors or reproduced by an open-source model re-trained on the same training set with the recommended setting.

\subsection{Comparisons with the State-of-the-arts}
\noindent\textbf{Quantitative Evaluation.} 
Table~\ref{tab:quant} shows the comparison results of AGLNet with 20 cutting-edge methods. 
We used the most common additional cues available currently, including boundary, texture, canny, and frequency. It can be seen that our proposed AGLNet achieves significant performance improvements, regardless of what additional information is used, and outperforms other comparison methods on all datasets. Note that unless otherwise specified, the following AGLNet refers to frequency as the additional clue. 
Compared with FDCOD~\cite{zhong2022detecting}, which also introduces the frequency domain cues for COD, our AGLNet shows a large performance improvement. 
Specifically, our method increases the performance on the three COD datasets by an average of 4.9\%, 8.6\%, 7.7\%, 2.8\%, and 26.4\% for $S_{\alpha}$, $F_{\beta}^{\omega}$, $F_{m}$, $E_{m}$ and $MAE$, respectively. 
The performance improvement can be attributed to the learnable additional information exploration and the deep multi-level integration within the same domain in the proposed AGLNet. This leads to a more effective incorporation of additional cues to guide object prediction. 
Compared with another well-performing method that does not use additional cues, \textit{e.g.}, ZoomNet, our method improves $S_{\alpha}$ by 5.1\%, $F_{\beta}^{\omega}$ by 8.3\%, $F_{m}$ by 6.9\%, $E_{m}$ by 3.5\% and $MAE$ by 23.6\% averagely. 
Compared with the second-best competitor - HitNet, which adopts a powerful transformer as the backbone, our method still achieves better detection performance, with 2.1\%, 1.3\%, 0.5\%, 1.4\% and 15.4\% increase on NC4K dataset, and 3.6\%, 3.0\%, 2.4\%, 1.8\% and 12.3\% increase on CAMO dataset.
As a result, our AGLNet shows effectiveness and superior performance in detecting camouflaged objects compared with the existing methods.

\begin{table*}[t]
\caption{Ablation studies of our AGLNet. Note that Combination and Decoupling are sub-components of HFC.}
\resizebox{\linewidth}{!}{
\renewcommand{\arraystretch}{1.3}
\begin{tabular}{c|ccccc|ccccc|ccccc|ccccc}
\toprule[1pt]
\multirow{2}{*}{No.} & \multicolumn{5}{c|}{\textbf{Component}}                                                                              & \multicolumn{5}{c|}{\textbf{COD10K}}                                                                     & \multicolumn{5}{c|}{\textbf{NC4K}}                                                                                                            & \multicolumn{5}{c}{\textbf{CAMO}}                                                                                              \\ \cline{2-21} 
                     & Baseline      & Combination & Decoupling & RD & AIG & $S_{\alpha}\uparrow$              & $F_{\beta}^{\omega}\uparrow$              & $F_{m}\uparrow$              & $E_{m}\uparrow$              & $MAE\downarrow$                                   & \multicolumn{1}{c}{$S_{\alpha}\uparrow$} & \multicolumn{1}{c}{$F_{\beta}^{\omega}\uparrow$} & \multicolumn{1}{c}{$F_{m}\uparrow$} & \multicolumn{1}{c}{$E_{m}\uparrow$} & \multicolumn{1}{c|}{$MAE\downarrow$}               & \multicolumn{1}{c}{$S_{\alpha}\uparrow$} & \multicolumn{1}{c}{$F_{\beta}^{\omega}\uparrow$} & \multicolumn{1}{c}{$F_{m}\uparrow$} & \multicolumn{1}{c}{$E_{m}\uparrow$} & \multicolumn{1}{c}{$MAE\downarrow$} \\ \midrule[1pt]
\#1                  & $\checkmark$ &                         &                         &                        &                          & 0.829          & 0.701          & 0.731          & 0.881          & \multicolumn{1}{c|}{0.033}          & 0.846                 & 0.732                 & 0.776                 & 0.871                 & \multicolumn{1}{c|}{0.050}          & 0.840                 & 0.736                 & 0.780                 & 0.868                 & 0.068                 \\
\#2                  & $\checkmark$ & $\checkmark$            &                         &                        &                          & 0.859          & 0.757          & 0.782          & 0.905          & \multicolumn{1}{c|}{0.026}          & 0.879                 & 0.813                 & 0.841                 & 0.912                 & \multicolumn{1}{c|}{0.034}          & 0.860                 & 0.808                 & 0.837                 & 0.905                 & 0.051                 \\  
\#3                  & $\checkmark$ & $\checkmark$            & $\checkmark$            &                        &                          & 0.862          & 0.773          & 0.794          & 0.918          & \multicolumn{1}{c|}{0.026}          & 0.880                 & 0.824                 & 0.850                 & 0.921                 & \multicolumn{1}{c|}{0.035}          & 0.863                 & 0.810                 & 0.838                 & 0.913                 & 0.052                 \\
\#4                  & $\checkmark$ & $\checkmark$            & $\checkmark$            & $\checkmark$           &                          & 0.865          & 0.779          & 0.799          & 0.921          & \multicolumn{1}{c|}{0.025}          & 0.882                 & 0.828                 & 0.852                 & 0.926                 & \multicolumn{1}{c|}{0.035}          & 0.866                 & 0.814                 & 0.841                 & 0.914                 & 0.052                 \\ \cline{1-21}
\rowcolor[HTML]{EFEFEF} 
\#OUR                & $\checkmark$ & $\checkmark$            & $\checkmark$            & $\checkmark$           & $\checkmark$             & \textbf{0.875} & \textbf{0.791} & \textbf{0.813} & \textbf{0.933} & \multicolumn{1}{c|}{\textbf{0.023}} & \textbf{0.889}        & \textbf{0.836}        & \textbf{0.858}        & \textbf{0.934}        & \multicolumn{1}{c|}{\textbf{0.033}} & \textbf{0.874}        & \textbf{0.825}        & \textbf{0.851}        & \textbf{0.918}        & \textbf{0.050}        \\ \bottomrule[1pt]
\end{tabular}
}
\label{tab:ablation}
\end{table*}

\begin{figure}[t]
	\centering
	\subfigure[Image]{
		\begin{minipage}[t]{0.23\columnwidth}
			\centering
			\includegraphics[width=1.01\linewidth,height=0.84\columnwidth]{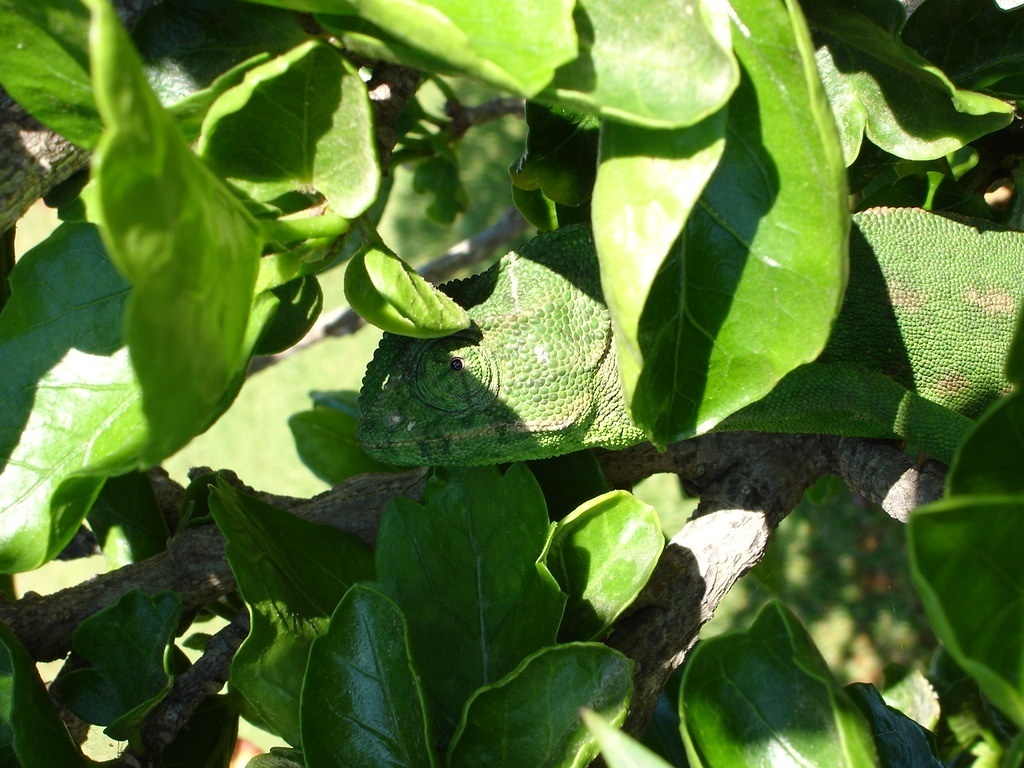}\\
			\vspace{0.01\linewidth}
            \includegraphics[width=1.01\linewidth,height=0.84\columnwidth]{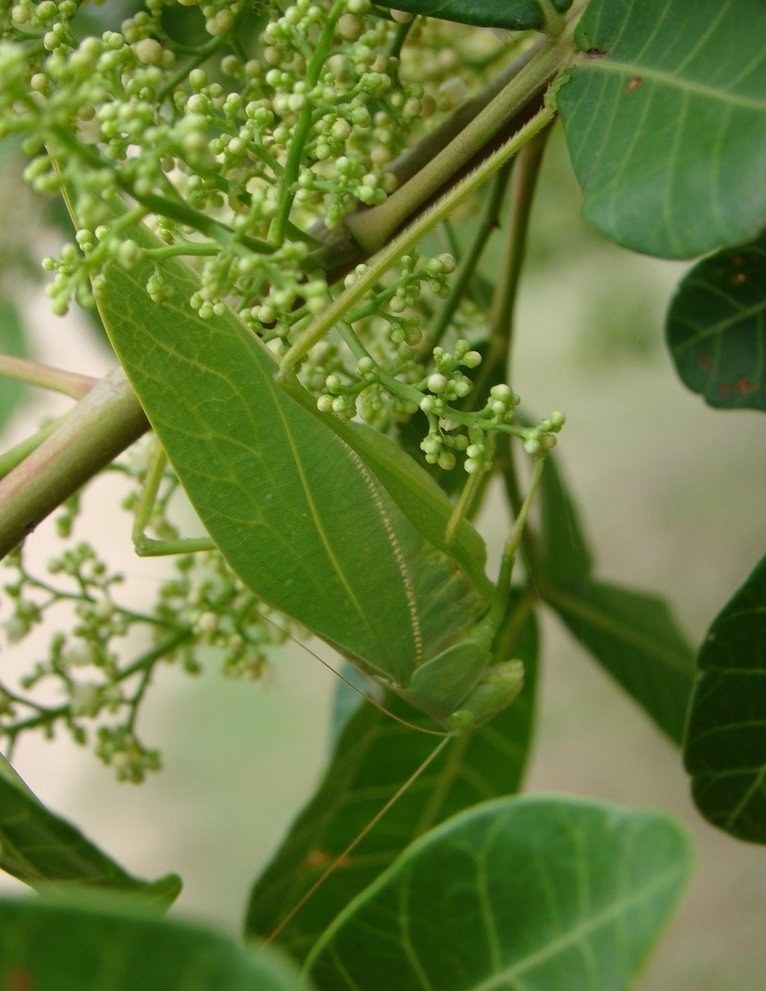}\\
            \vspace{0.01\linewidth}
			\includegraphics[width=1.01\linewidth,height=0.84\columnwidth]{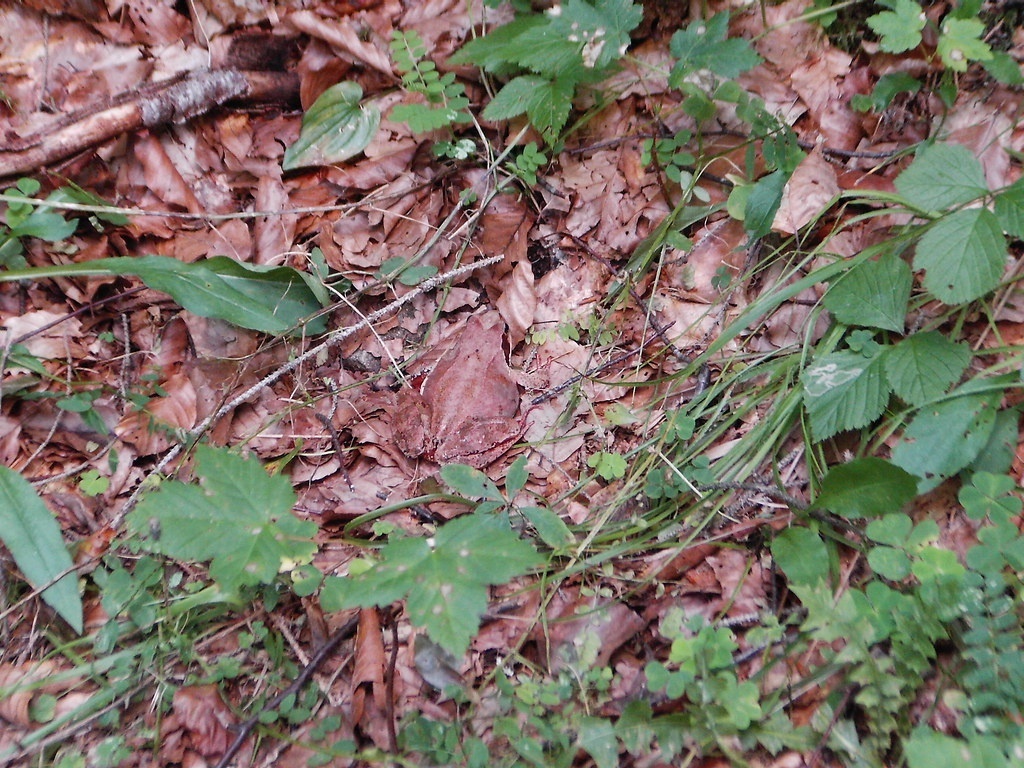}\\
			\vspace{0.11\linewidth}
		\end{minipage}%
	}\hspace{-0.012\columnwidth}
	\subfigure[GT]{
		\begin{minipage}[t]{0.23\columnwidth}
			\centering
			\includegraphics[width=1.01\linewidth,height=0.84\columnwidth]{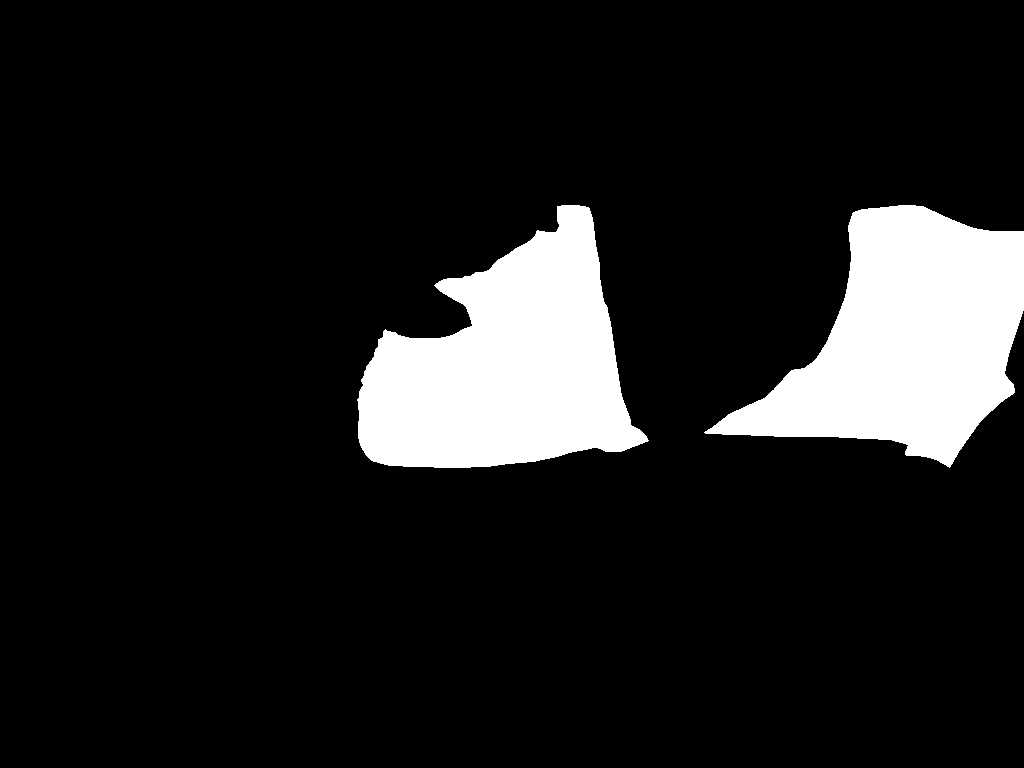}\\
			\vspace{0.01\linewidth}
            \includegraphics[width=1.01\linewidth,height=0.84\columnwidth]{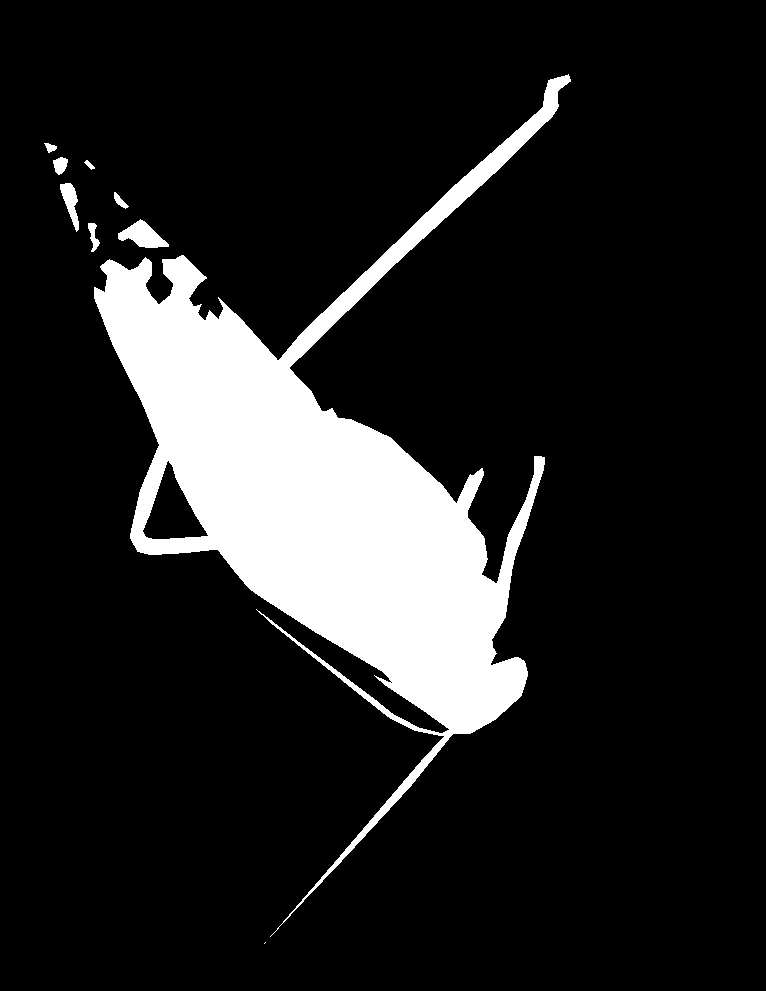}\\
            \vspace{0.01\linewidth}
			\includegraphics[width=1.01\linewidth,height=0.84\columnwidth]{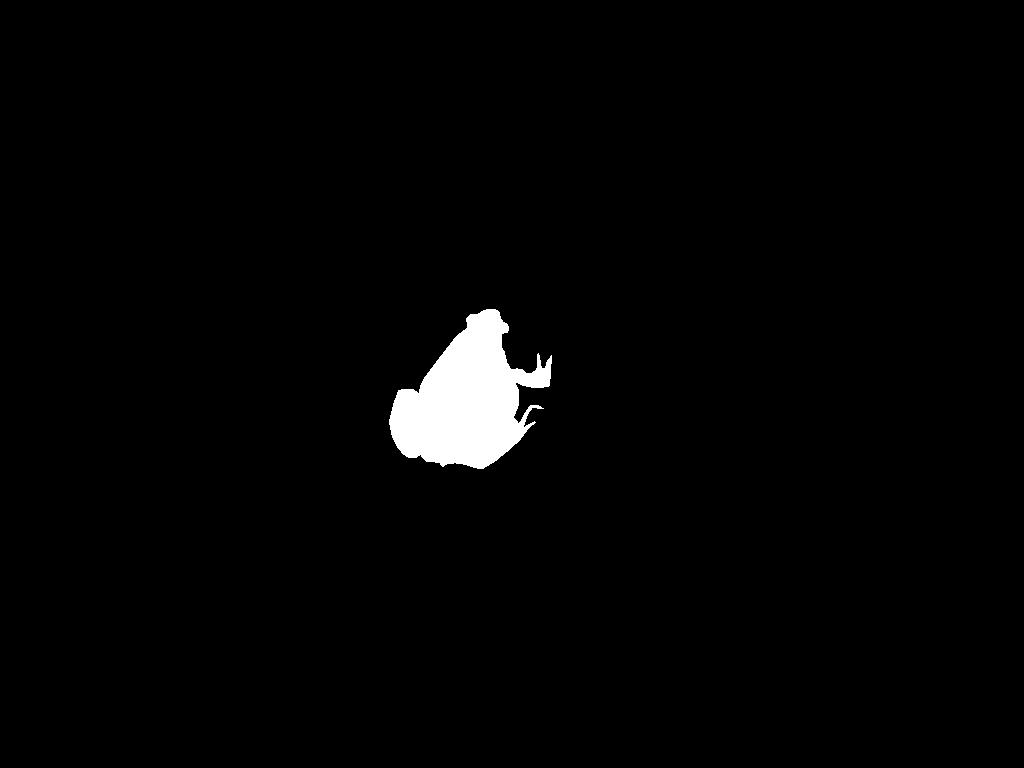}\\
			\vspace{0.11\linewidth}
		\end{minipage}%
	}\hspace{-0.012\columnwidth}
	\subfigure[Baseline]{
		\begin{minipage}[t]{0.23\columnwidth}
			\centering
			\includegraphics[width=1.01\linewidth,height=0.84\columnwidth]{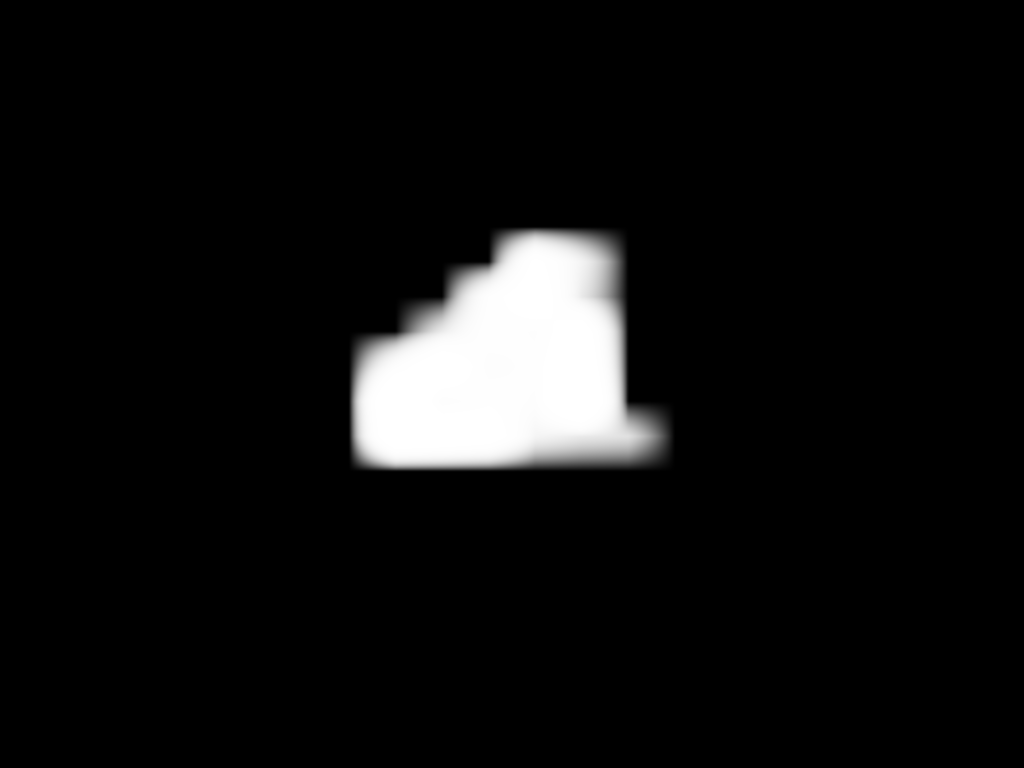}\\
			\vspace{0.01\linewidth}
            \includegraphics[width=1.01\linewidth,height=0.84\columnwidth]{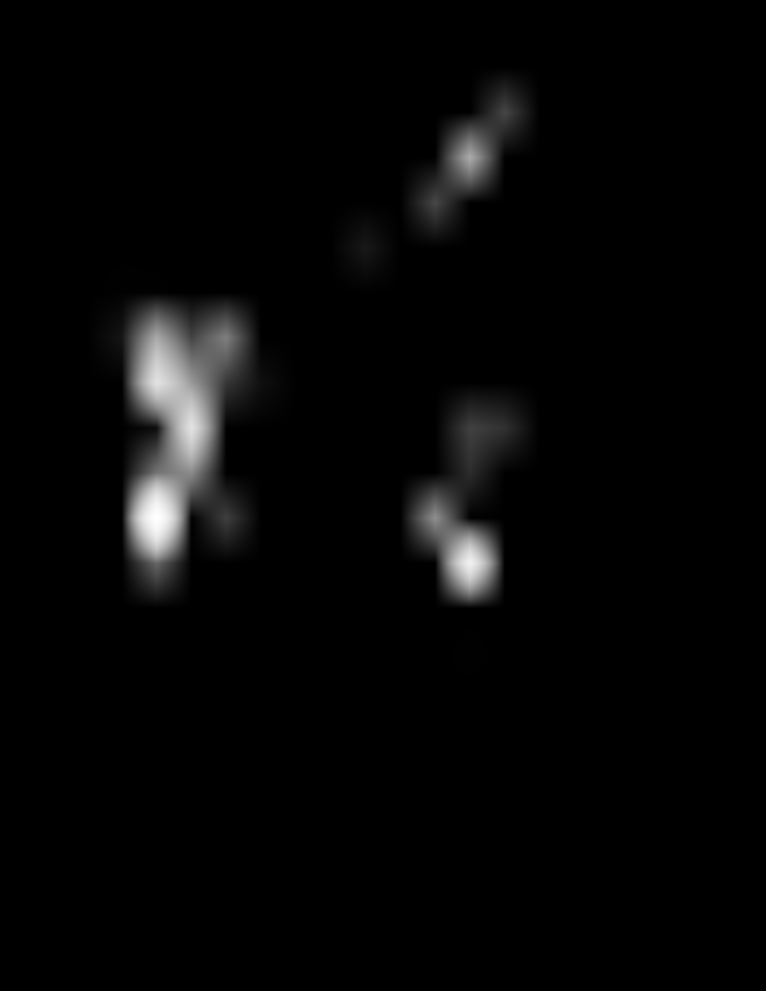}\\
            \vspace{0.01\linewidth}
			\includegraphics[width=1.01\linewidth,height=0.84\columnwidth]{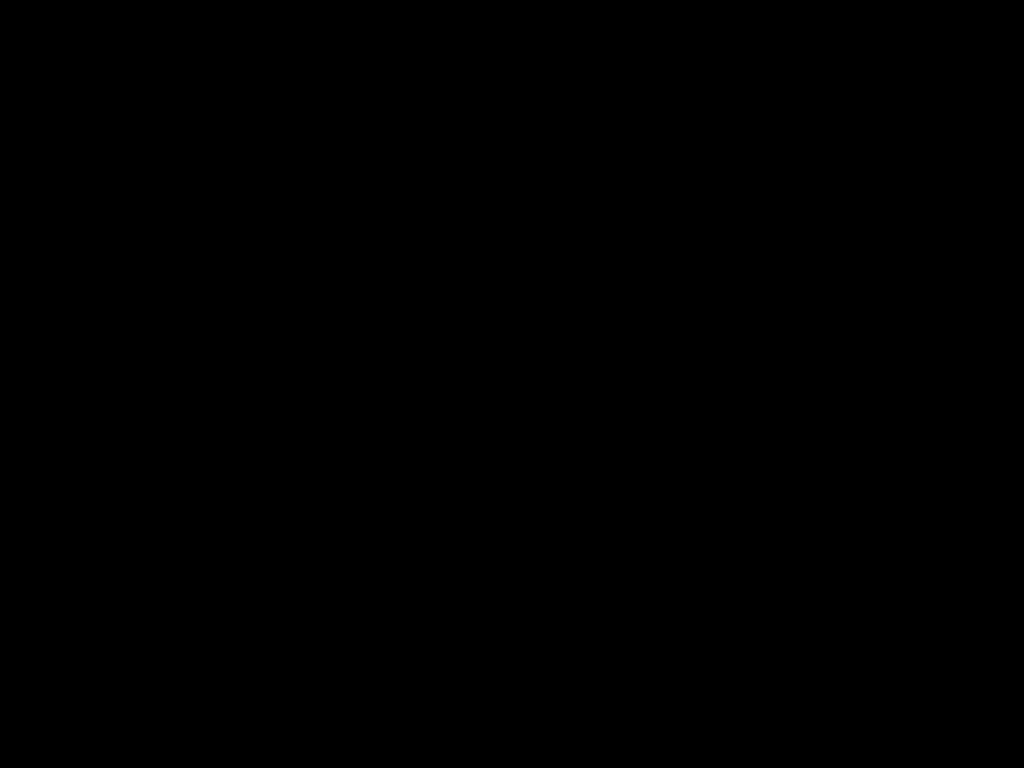}\\
			\vspace{0.11\linewidth}
		\end{minipage}%
	}\hspace{-0.012\columnwidth}
	\subfigure[\scriptsize +Combination]{
		\begin{minipage}[t]{0.23\columnwidth}
			\centering
			\includegraphics[width=1.01\linewidth,height=0.84\columnwidth]{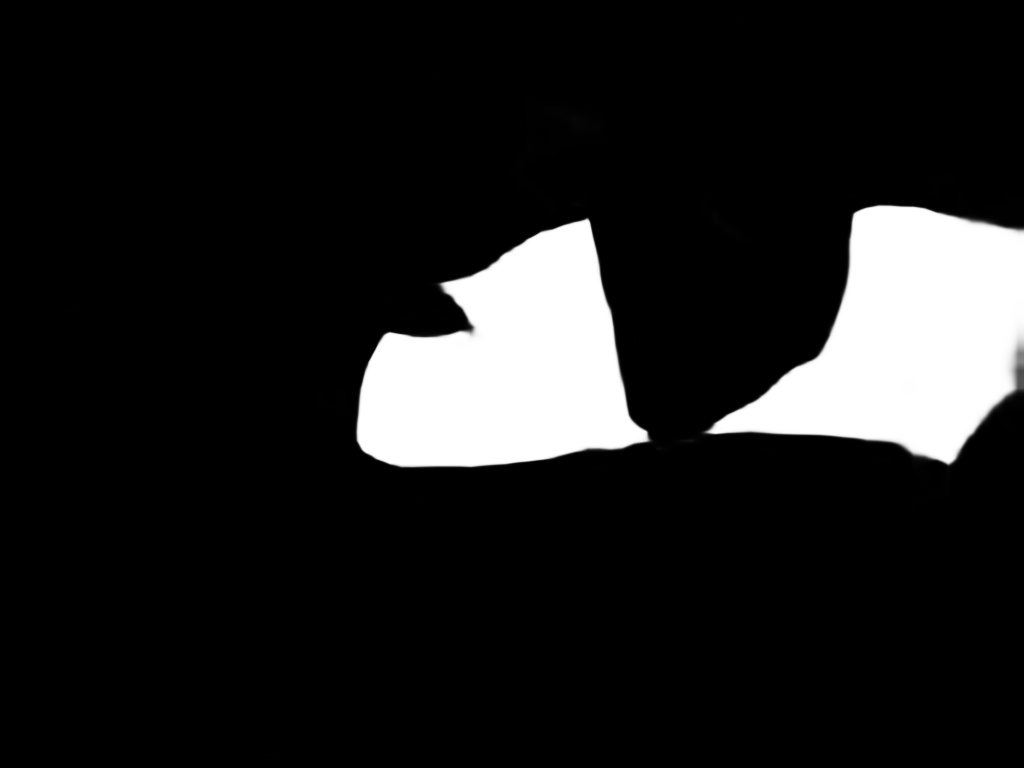}\\
			\vspace{0.01\linewidth}
            \includegraphics[width=1.01\linewidth,height=0.84\columnwidth]{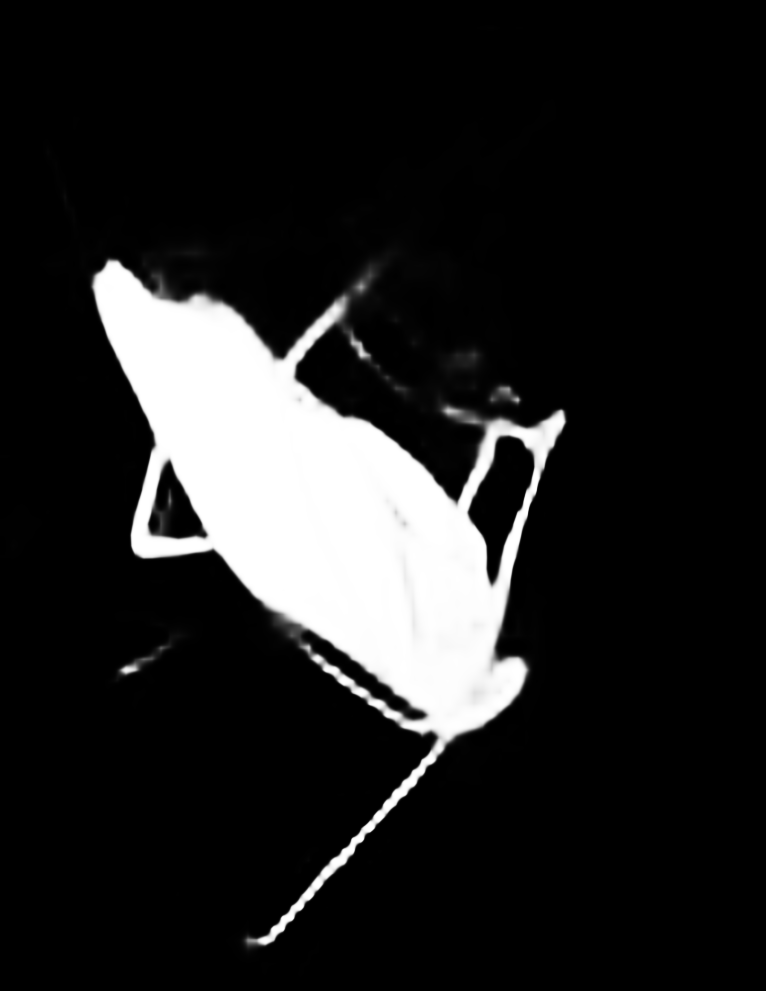}\\
            \vspace{0.01\linewidth}
			\includegraphics[width=1.01\linewidth,height=0.84\columnwidth]{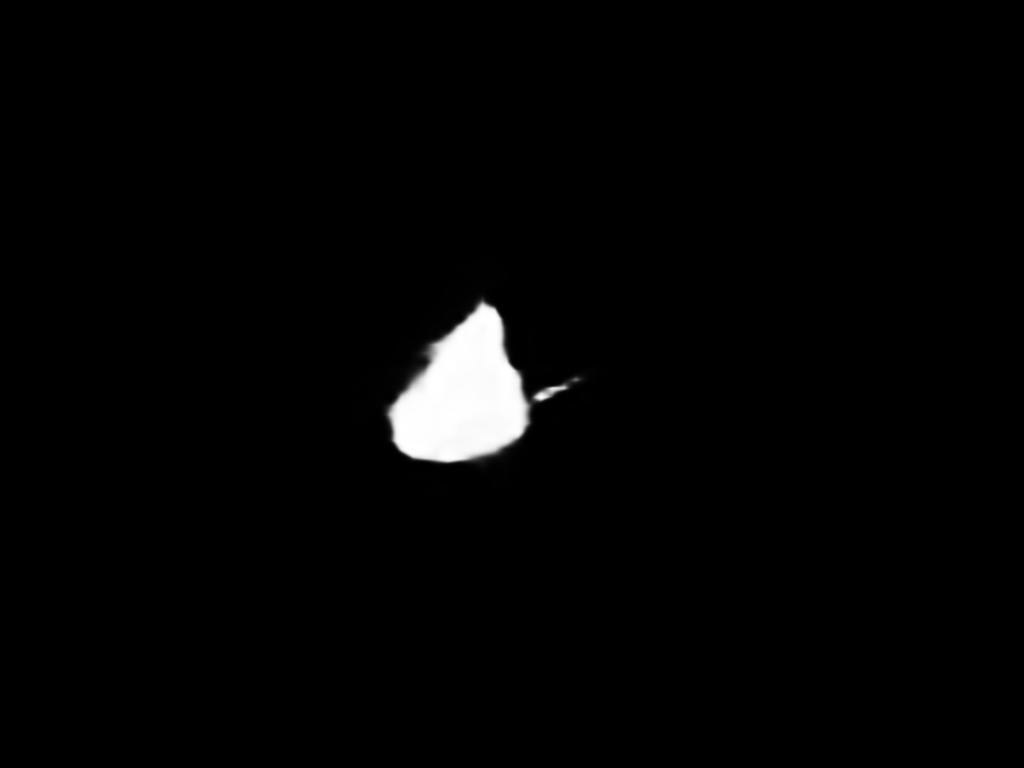}\\
			\vspace{0.11\linewidth}
		\end{minipage}%
	}\hspace{-0.012\columnwidth}
	\centering
	\caption{\textbf{Visual comparison of the proposed Combination part.} (a) input image, (b) ground-truth, (c) baseline, and (d) baseline + Combination.}
    \label{fig:mfc}
\end{figure}

\begin{figure}[t]
	\centering
	\subfigure[Image]{
		\begin{minipage}[t]{0.23\columnwidth}
			\centering
			\includegraphics[width=1.01\linewidth,height=0.84\columnwidth]{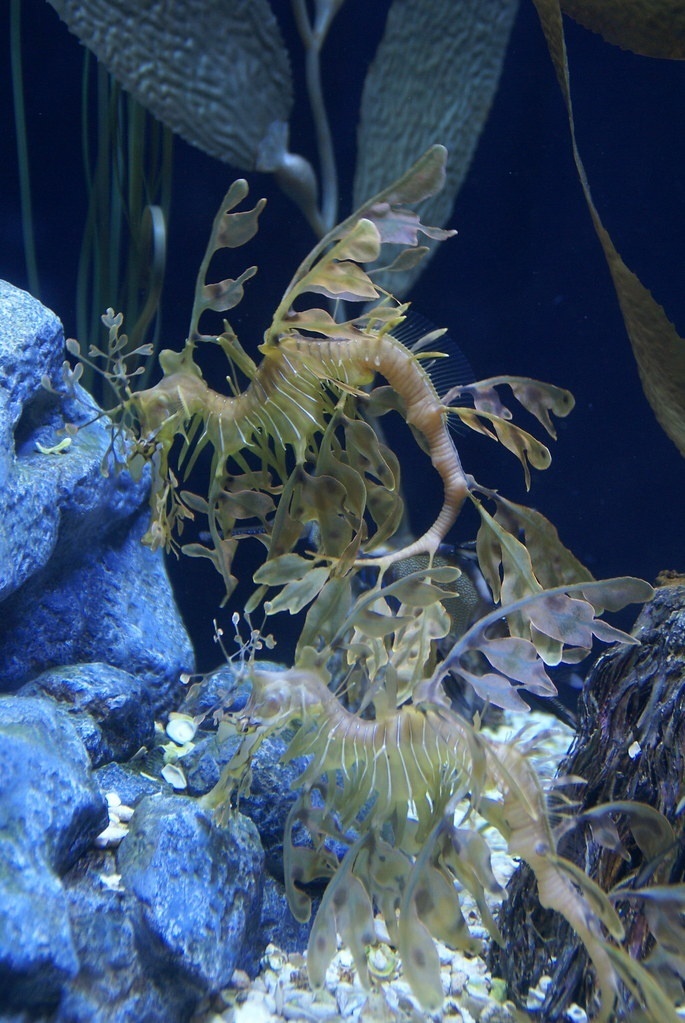}\\
			\vspace{0.01\linewidth}
            \includegraphics[width=1.01\linewidth,height=0.84\columnwidth]{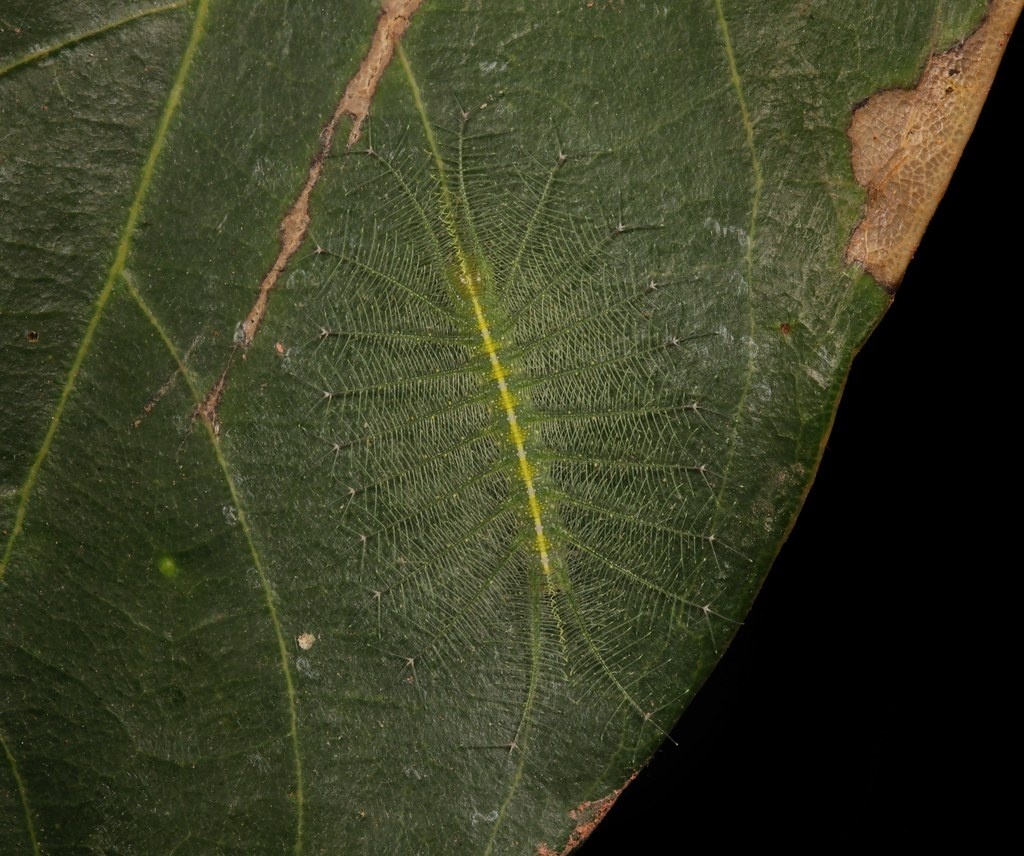}\\
            \vspace{0.01\linewidth}
			\includegraphics[width=1.01\linewidth,height=0.84\columnwidth]{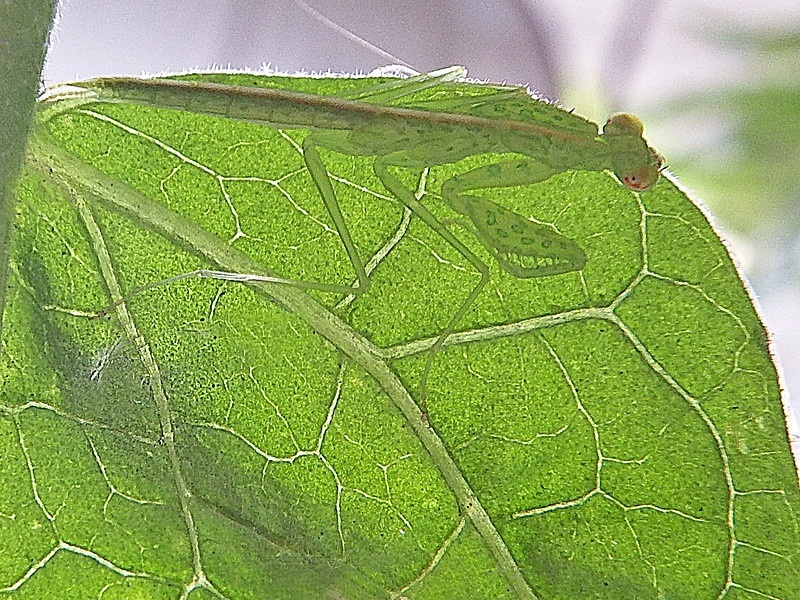}\\
			\vspace{0.11\linewidth}
		\end{minipage}%
	}\hspace{-0.012\columnwidth}
	\subfigure[GT]{
		\begin{minipage}[t]{0.23\columnwidth}
			\centering
			\includegraphics[width=1.01\linewidth,height=0.84\columnwidth]{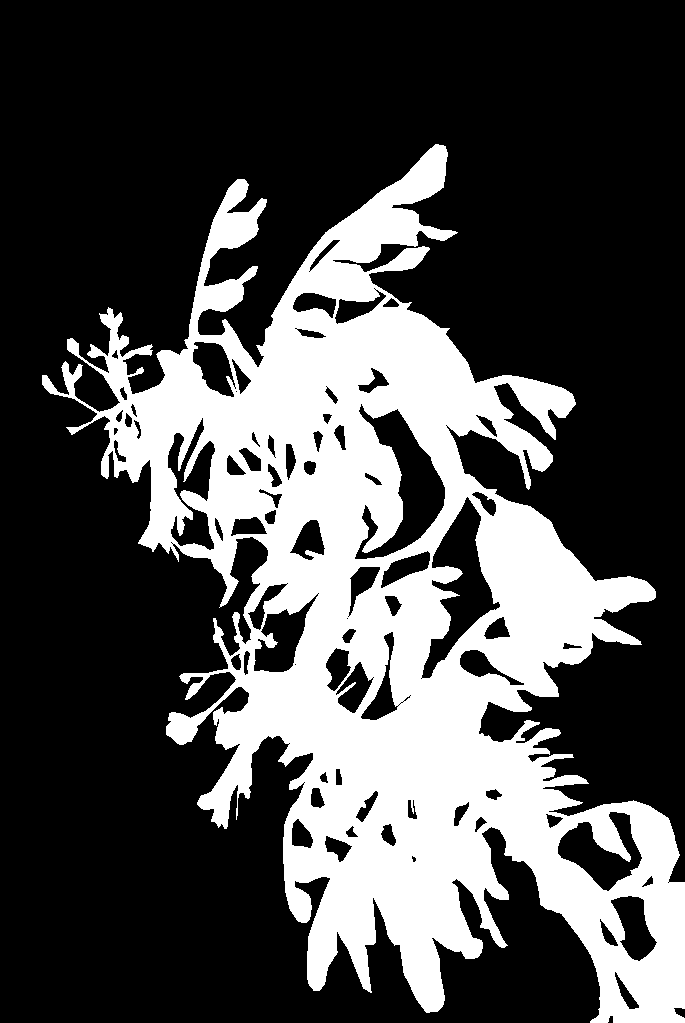}\\
			\vspace{0.01\linewidth}
            \includegraphics[width=1.01\linewidth,height=0.84\columnwidth]{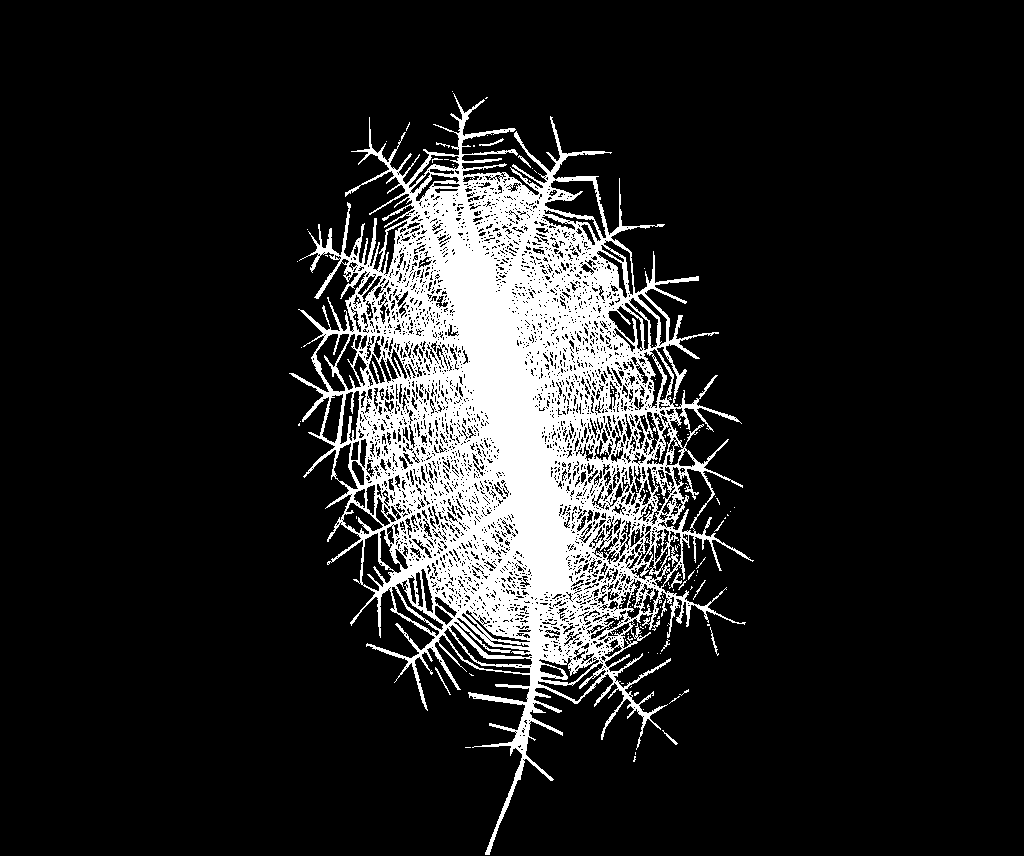}\\
            \vspace{0.01\linewidth}
			\includegraphics[width=1.01\linewidth,height=0.84\columnwidth]{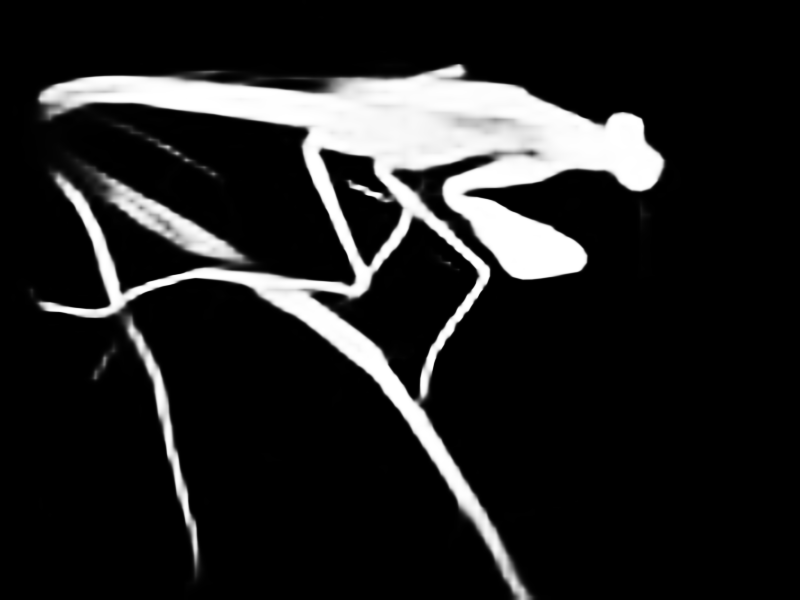}\\
			\vspace{0.11\linewidth}
		\end{minipage}%
	}\hspace{-0.012\columnwidth}
	\subfigure[\scriptsize +Combination]{
		\begin{minipage}[t]{0.23\columnwidth}
			\centering
			\includegraphics[width=1.01\linewidth,height=0.84\columnwidth]{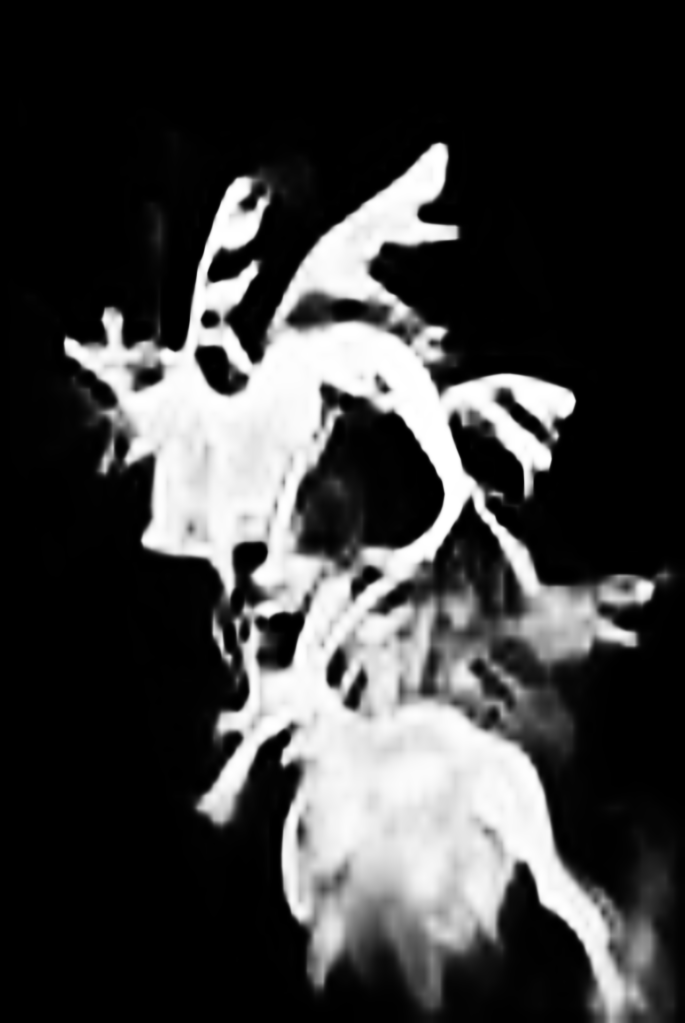}\\
			\vspace{0.01\linewidth}
            \includegraphics[width=1.01\linewidth,height=0.84\columnwidth]{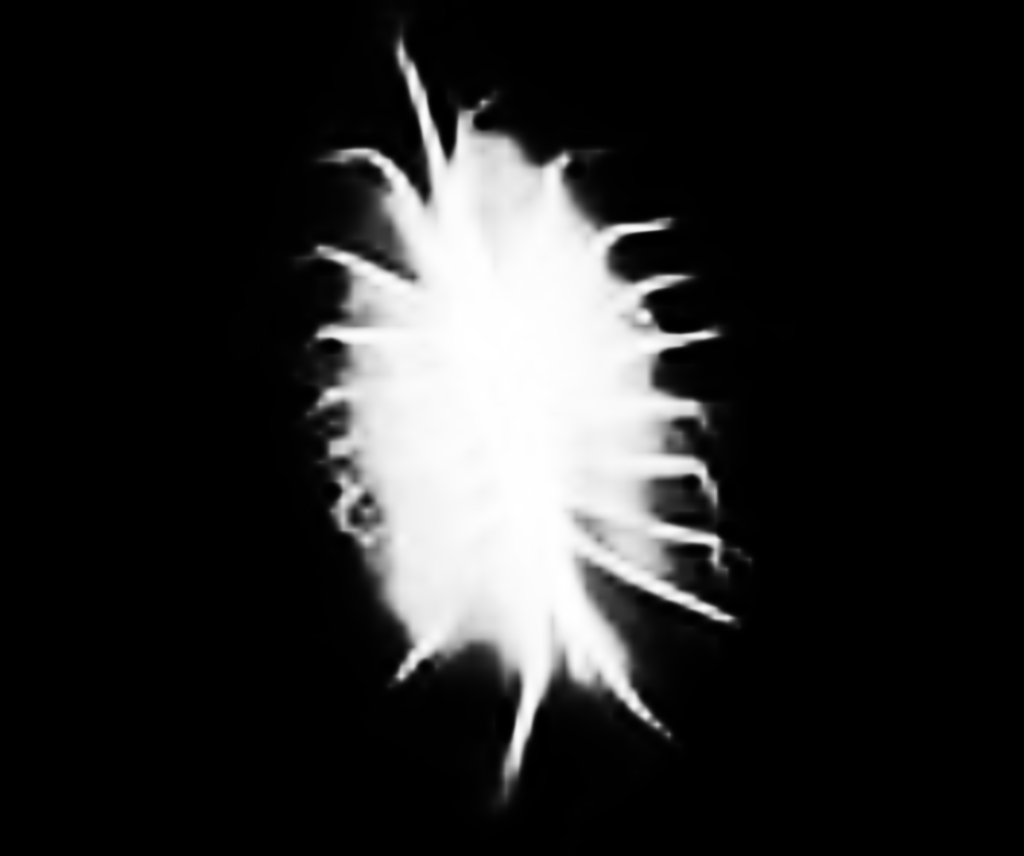}\\
            \vspace{0.01\linewidth}
			\includegraphics[width=1.01\linewidth,height=0.84\columnwidth]{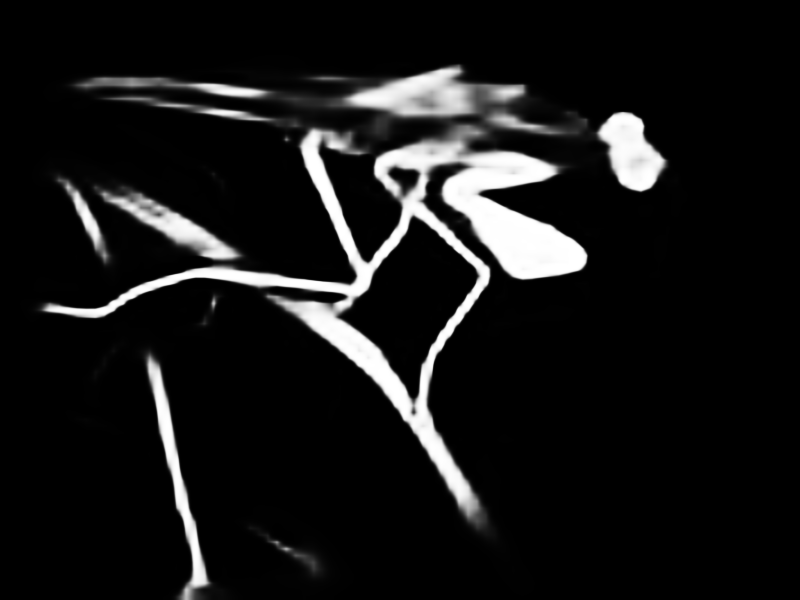}\\
			\vspace{0.11\linewidth}
		\end{minipage}%
	}\hspace{-0.012\columnwidth}
	\subfigure[\scriptsize +Decoupling]{
		\begin{minipage}[t]{0.23\columnwidth}
			\centering
			\includegraphics[width=1.01\linewidth,height=0.84\columnwidth]{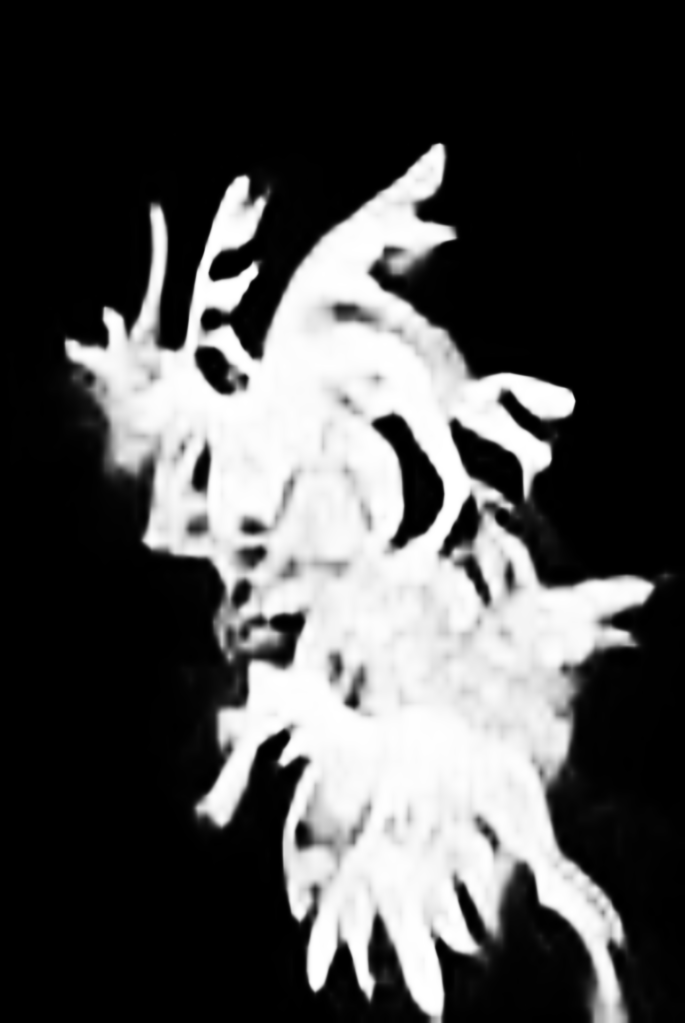}\\
			\vspace{0.01\linewidth}
            \includegraphics[width=1.01\linewidth,height=0.84\columnwidth]{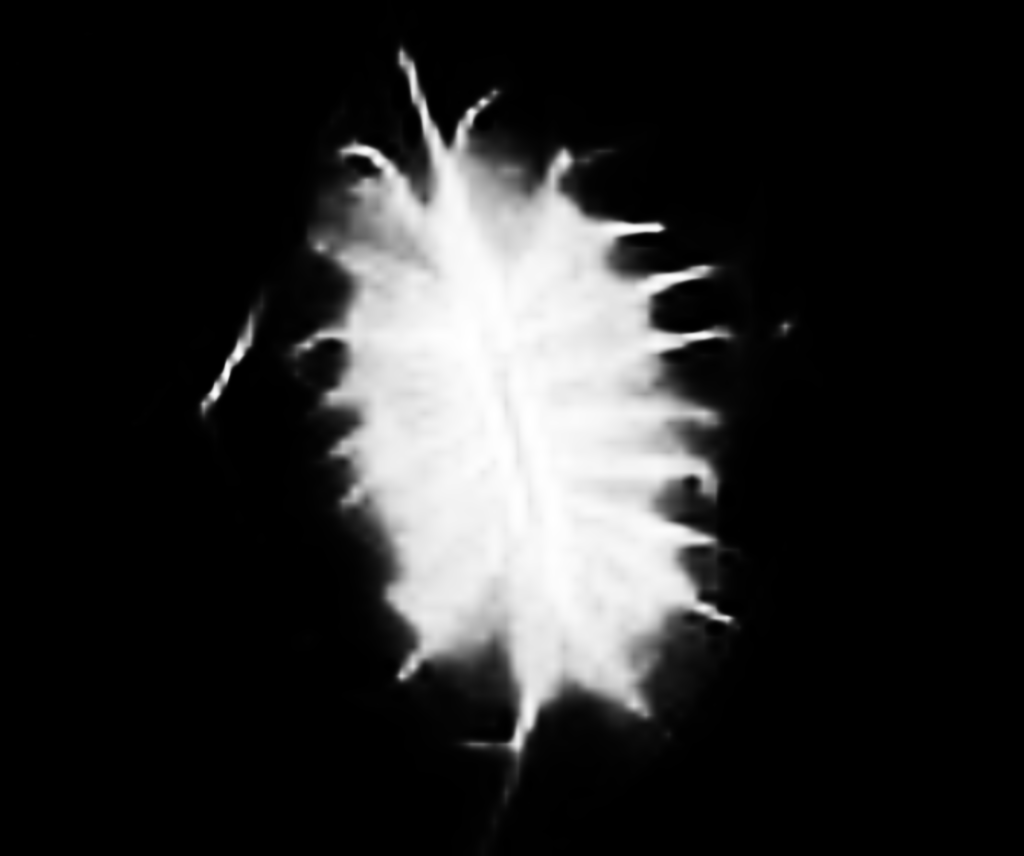}\\
            \vspace{0.01\linewidth}
			\includegraphics[width=1.01\linewidth,height=0.84\columnwidth]{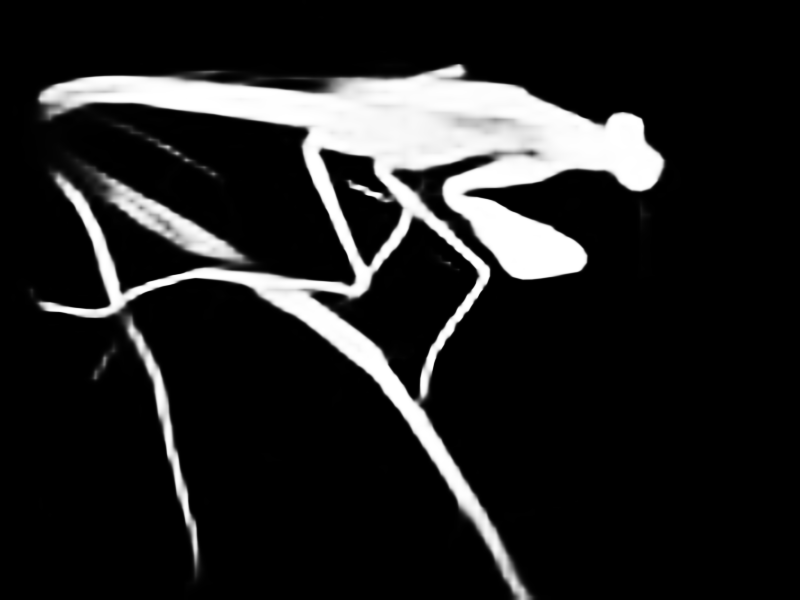}\\
			\vspace{0.11\linewidth}
		\end{minipage}%
	}\hspace{-0.012\columnwidth}
	\centering
	\caption{\textbf{Visual comparison of the Decoupling part.} (a) input image, (b) ground-truth, (c) baseline+Combination, (d) baseline+Combination+Decoupling. Red circles shows improvements.}
    \label{fig:fgd}
\end{figure}

\begin{figure}[th]
	\centering
	\subfigure[Image]{
		\begin{minipage}[t]{0.23\columnwidth}
			\centering
			\includegraphics[width=1.01\linewidth,height=0.84\columnwidth]{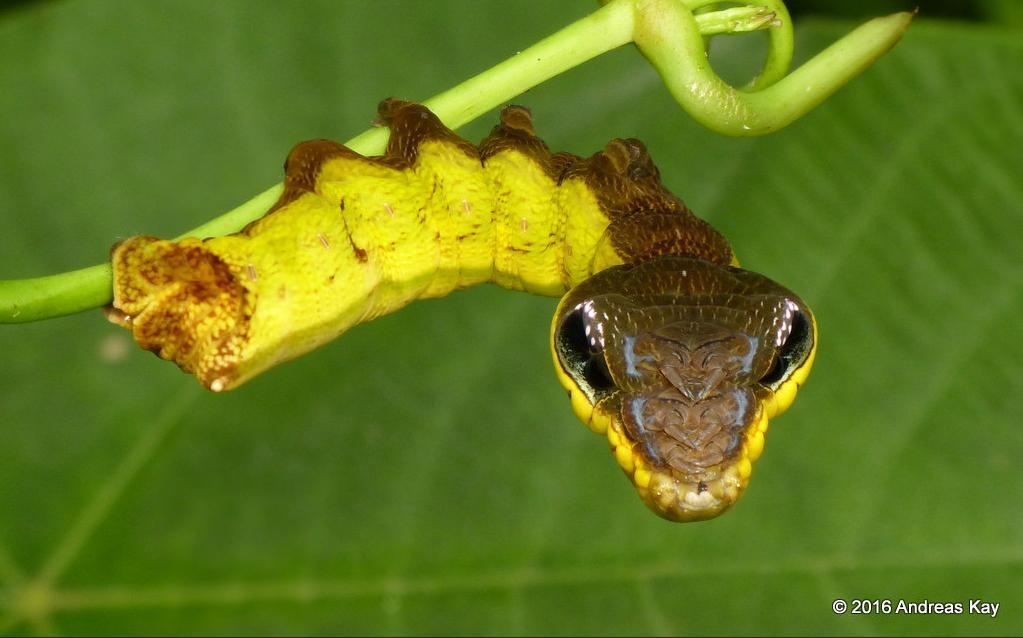}\\
			\vspace{0.01\linewidth}
            \includegraphics[width=1.01\linewidth,height=0.84\columnwidth]{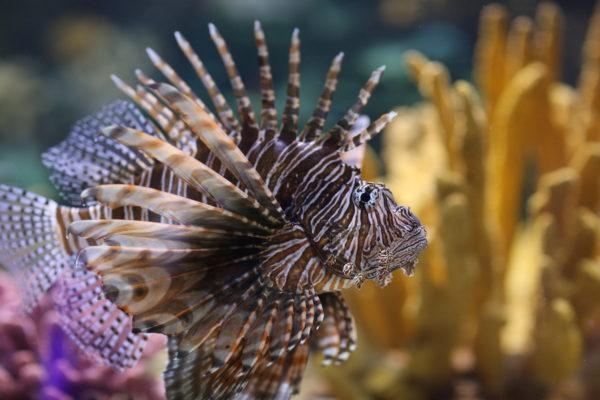}\\
            \vspace{0.01\linewidth}
			\includegraphics[width=1.01\linewidth,height=0.84\columnwidth]{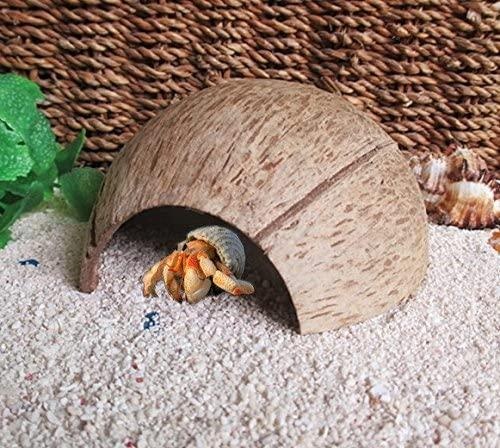}\\
			\vspace{0.11\linewidth}
		\end{minipage}%
	}\hspace{-0.012\columnwidth}
	\subfigure[GT]{
		\begin{minipage}[t]{0.23\columnwidth}
			\centering
			\includegraphics[width=1.01\linewidth,height=0.84\columnwidth]{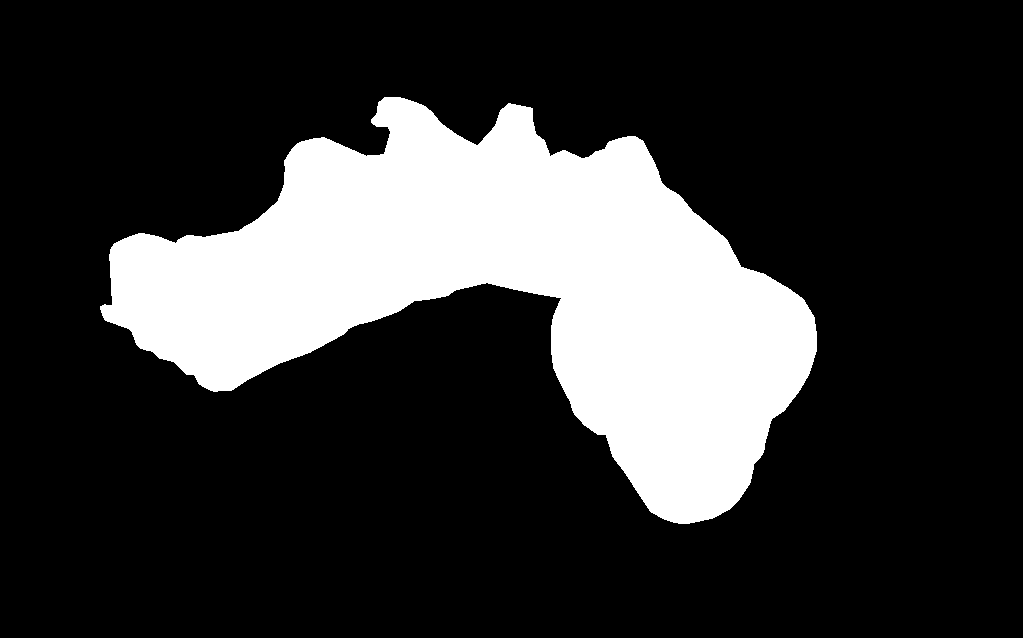}\\
			\vspace{0.01\linewidth}
            \includegraphics[width=1.01\linewidth,height=0.84\columnwidth]{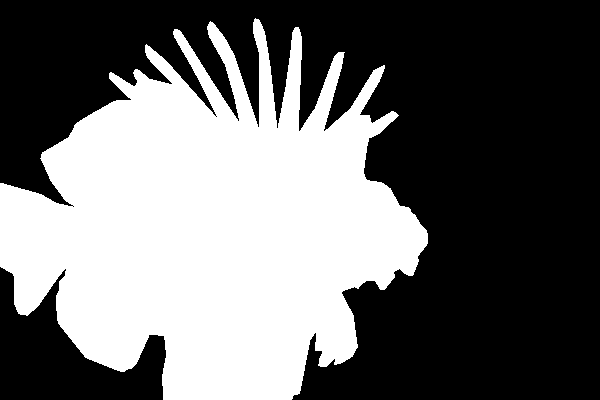}\\
            \vspace{0.01\linewidth}
			\includegraphics[width=1.01\linewidth,height=0.84\columnwidth]{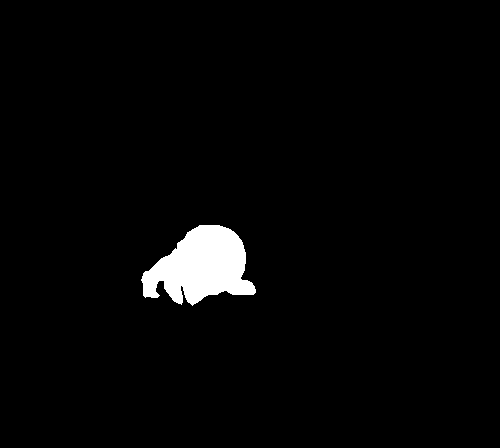}\\
			\vspace{0.11\linewidth}
		\end{minipage}%
	}\hspace{-0.012\columnwidth}
	\subfigure[+HFC]{
		\begin{minipage}[t]{0.23\columnwidth}
			\centering
			\includegraphics[width=1.01\linewidth,height=0.84\columnwidth]{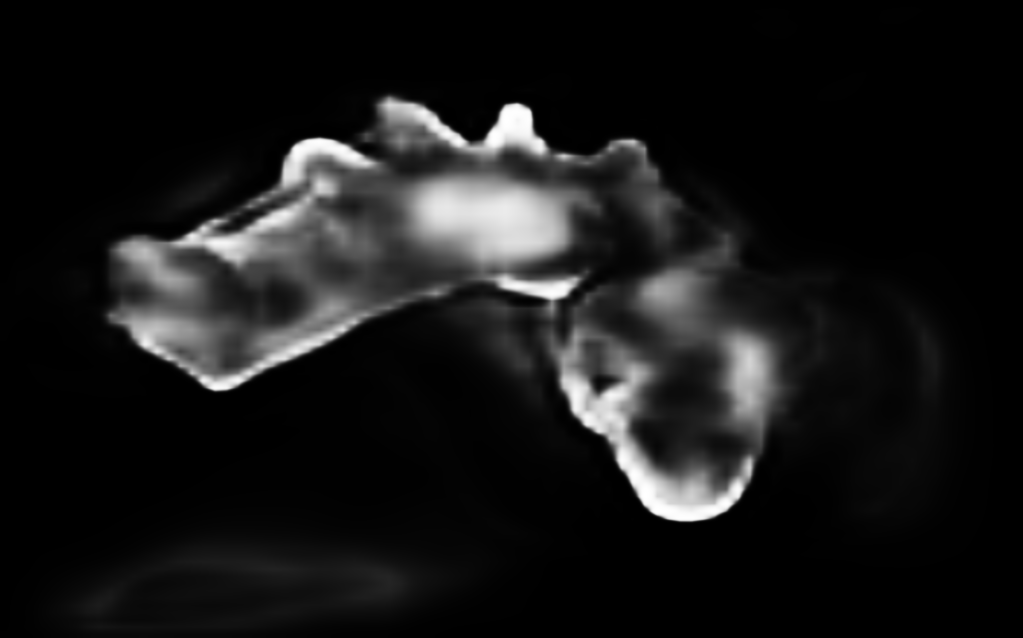}\\
			\vspace{0.01\linewidth}
            \includegraphics[width=1.01\linewidth,height=0.84\columnwidth]{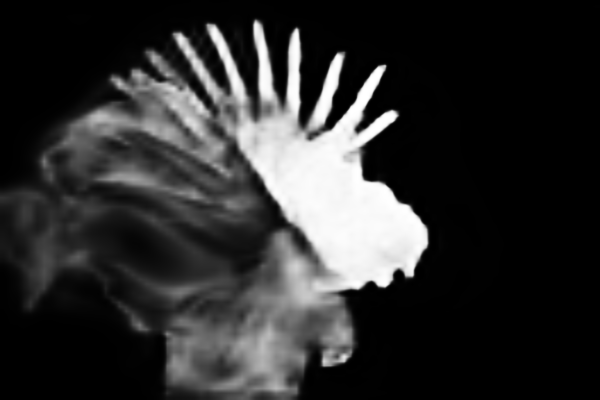}\\
            \vspace{0.01\linewidth}
			\includegraphics[width=1.01\linewidth,height=0.84\columnwidth]{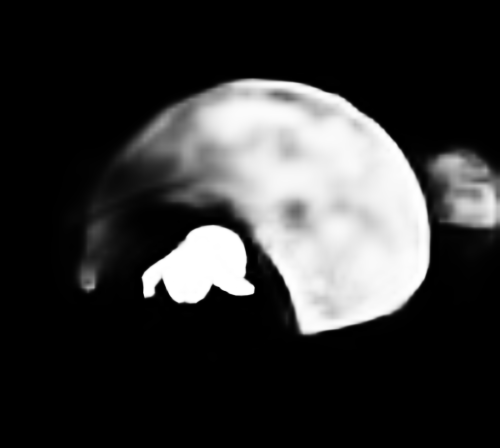}\\
			\vspace{0.11\linewidth}
		\end{minipage}%
	}\hspace{-0.012\columnwidth}
	\subfigure[+RD]{
		\begin{minipage}[t]{0.23\columnwidth}
			\centering
			\includegraphics[width=1.01\linewidth,height=0.84\columnwidth]{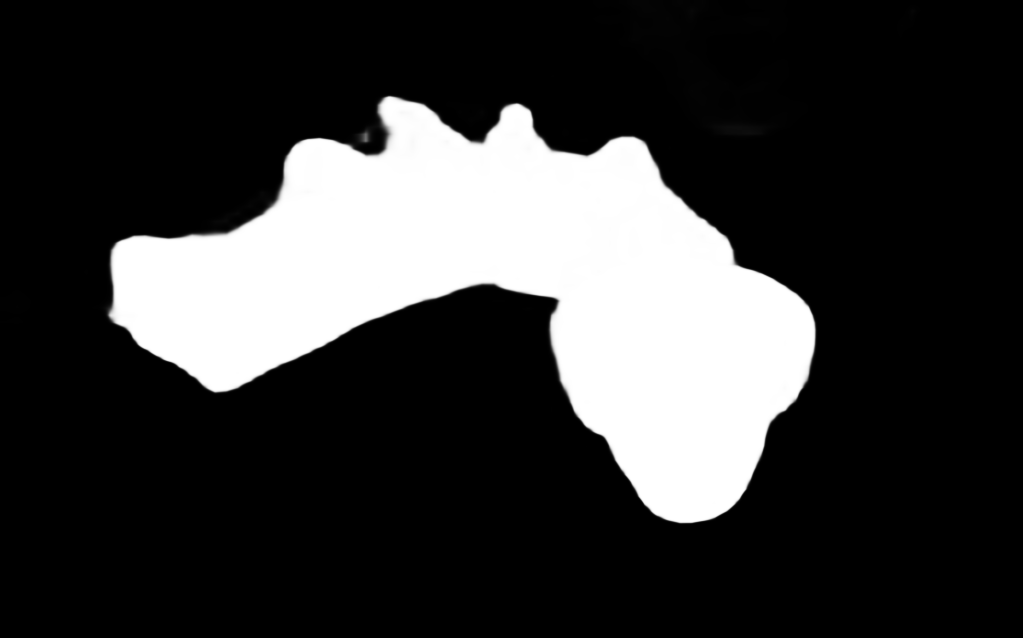}\\
			\vspace{0.01\linewidth}
            \includegraphics[width=1.01\linewidth,height=0.84\columnwidth]{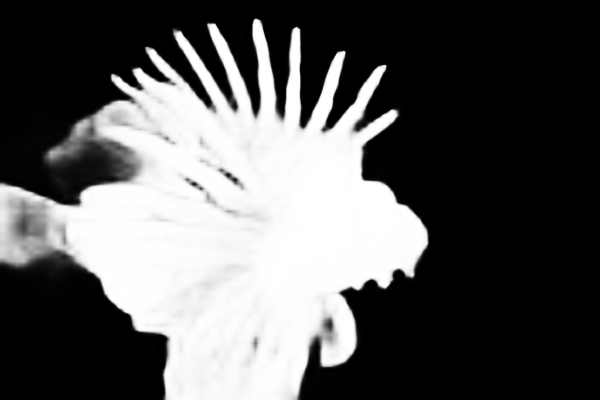}\\
            \vspace{0.01\linewidth}
			\includegraphics[width=1.01\linewidth,height=0.84\columnwidth]{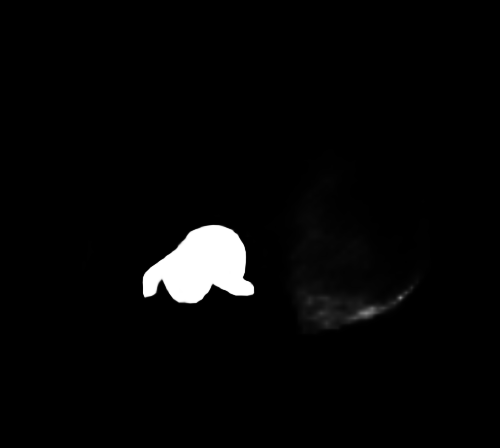}\\
			\vspace{0.11\linewidth}
		\end{minipage}%
	}\hspace{-0.012\columnwidth}
	\centering
	\caption{\textbf{Visual comparison of the proposed RD module.} (a) input image, (b) ground-truth, (c) baseline+HFC, and (d) baseline+HDC+RD.}
	\label{fig:RD}
\end{figure}

\begin{figure}[th]
	\centering
	\subfigure[Image]{
		\begin{minipage}[t]{0.23\columnwidth}
			\centering
			\includegraphics[width=1.01\linewidth,height=0.84\columnwidth]{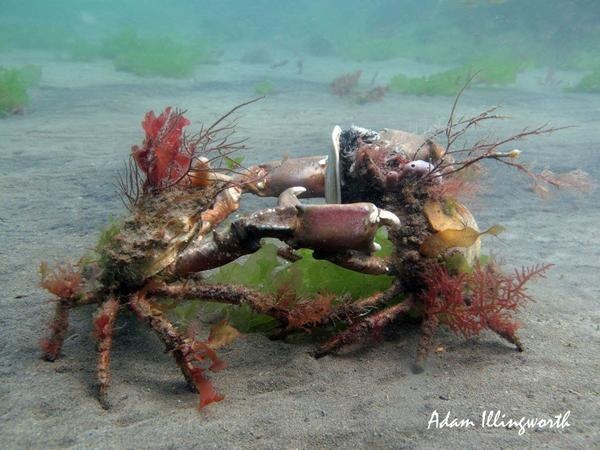}\\
			\vspace{0.01\linewidth}
            \includegraphics[width=1.01\linewidth,height=0.84\columnwidth]{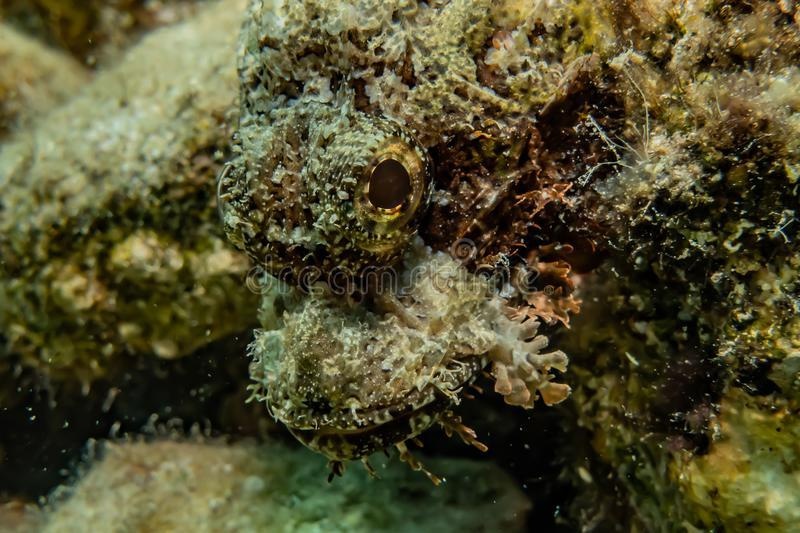}\\
            \vspace{0.01\linewidth}
			\includegraphics[width=1.01\linewidth,height=0.84\columnwidth]{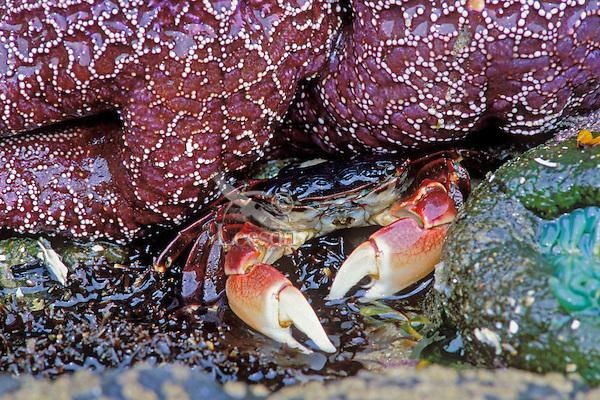}\\
			\vspace{0.11\linewidth}
		\end{minipage}%
	}\hspace{-0.012\columnwidth}
	\subfigure[GT]{
		\begin{minipage}[t]{0.23\columnwidth}
			\centering
			\includegraphics[width=1.01\linewidth,height=0.84\columnwidth]{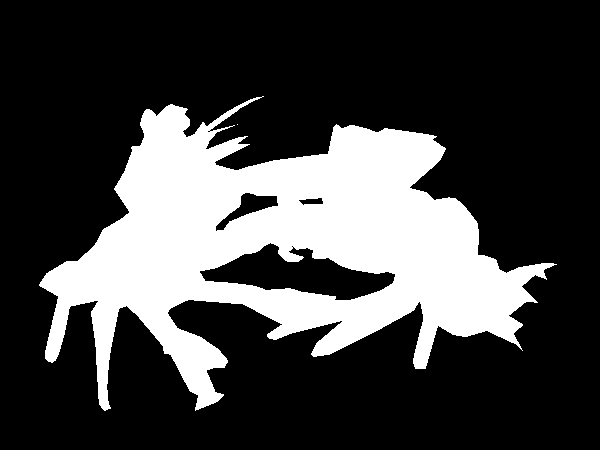}\\
			\vspace{0.01\linewidth}
            \includegraphics[width=1.01\linewidth,height=0.84\columnwidth]{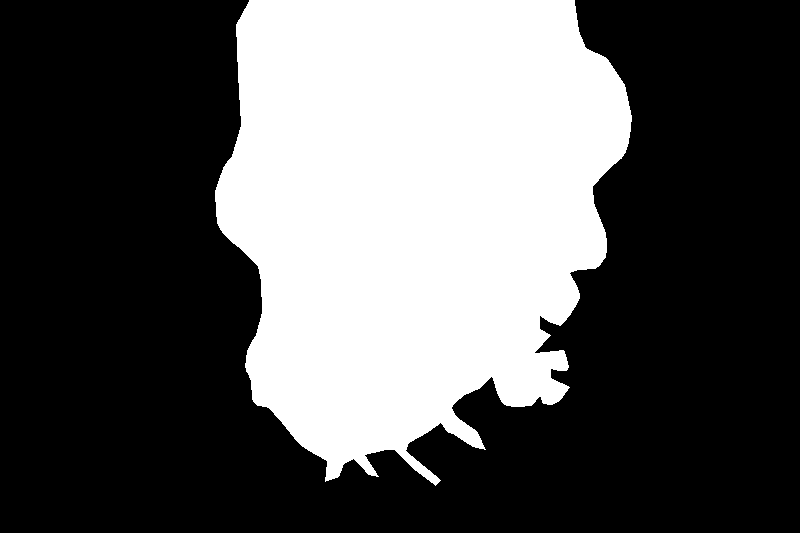}\\
            \vspace{0.01\linewidth}
			\includegraphics[width=1.01\linewidth,height=0.84\columnwidth]{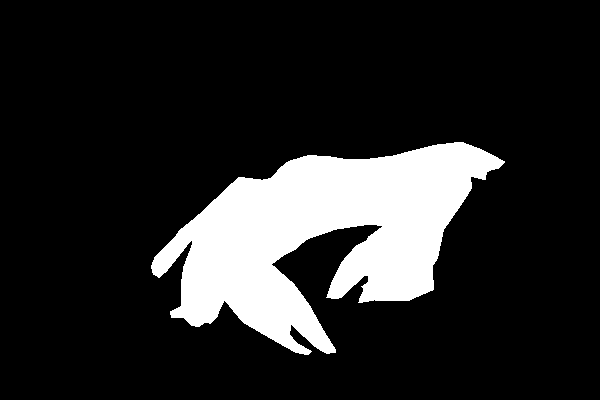}\\
			\vspace{0.11\linewidth}
		\end{minipage}%
	}\hspace{-0.012\columnwidth}
	\subfigure[w/o AIG]{
		\begin{minipage}[t]{0.23\columnwidth}
			\centering
			\includegraphics[width=1.01\linewidth,height=0.84\columnwidth]{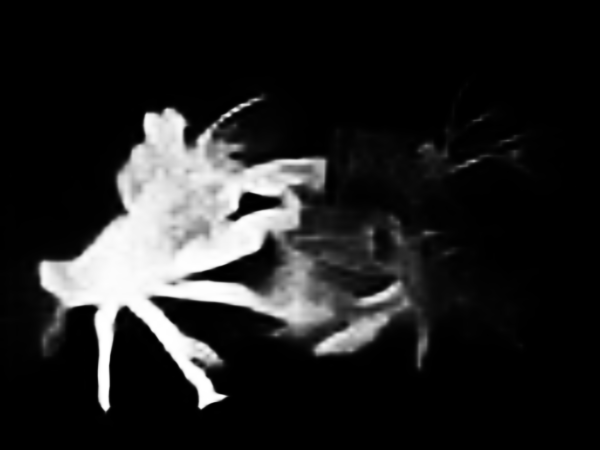}\\
			\vspace{0.01\linewidth}
            \includegraphics[width=1.01\linewidth,height=0.84\columnwidth]{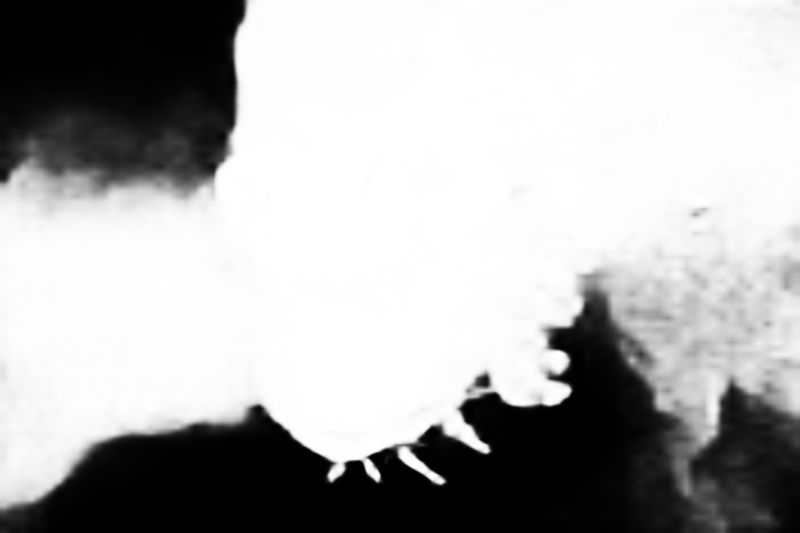}\\
            \vspace{0.01\linewidth}
			\includegraphics[width=1.01\linewidth,height=0.84\columnwidth]{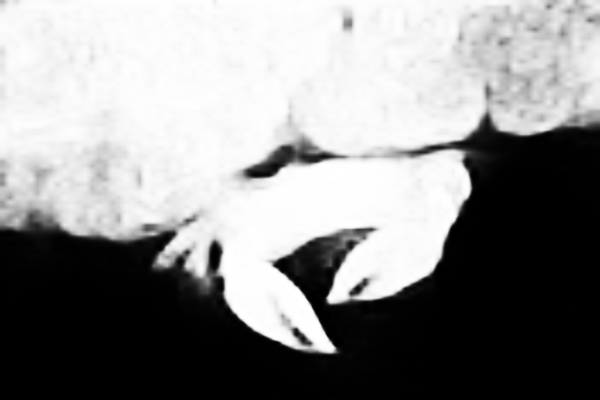}\\
			\vspace{0.11\linewidth}
		\end{minipage}%
	}\hspace{-0.012\columnwidth}
	\subfigure[AGLNet]{
		\begin{minipage}[t]{0.23\columnwidth}
			\centering
			\includegraphics[width=1.01\linewidth,height=0.84\columnwidth]{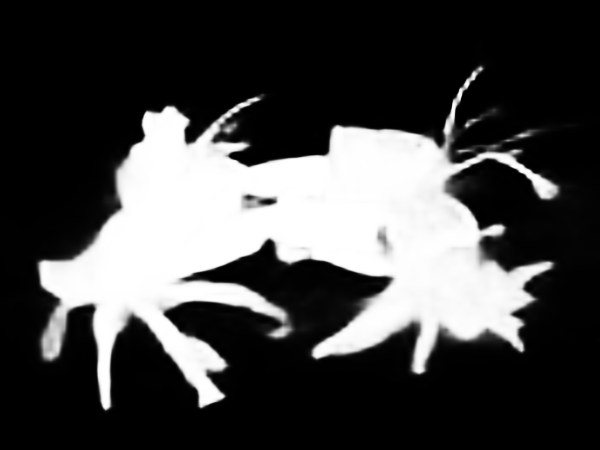}\\
			\vspace{0.01\linewidth}
            \includegraphics[width=1.01\linewidth,height=0.84\columnwidth]{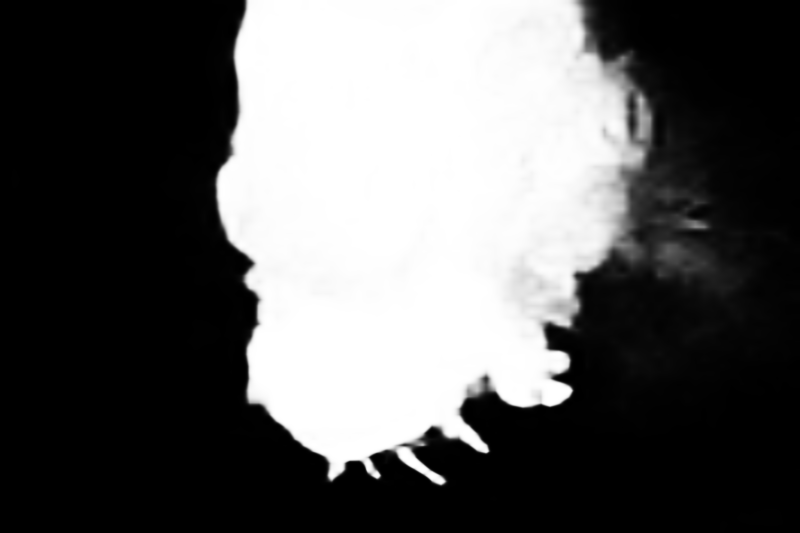}\\
            \vspace{0.01\linewidth}
			\includegraphics[width=1.01\linewidth,height=0.84\columnwidth]{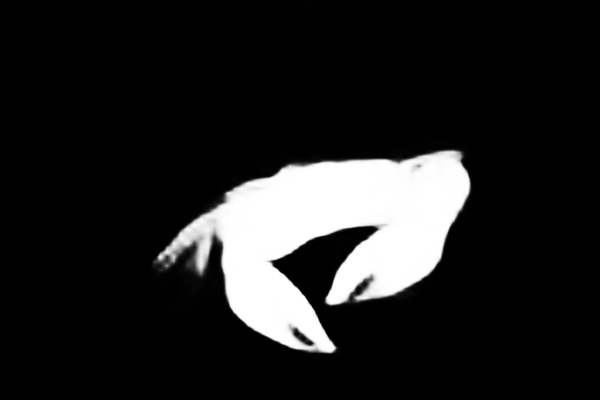}\\
			\vspace{0.11\linewidth}
		\end{minipage}%
	}\hspace{-0.012\columnwidth}
	\centering
	\caption{\textbf{Visual comparison of the proposed AIG.} (a) input image, (b) ground-truth, (c) AGLNet w/o AIG, (d) AGLNet.}
	\label{fig:fdg}
\end{figure}

\vspace{10pt}
\noindent
\textbf{Qualitative Evaluation.} 
Figure~\ref{fig:GSNet_Qualitative} shows the visual comparisons between our AGLNet and other representative COD methods in some challenging scenarios, including tiny objects (e.g., lines 1-2), occlusions (e.g., lines 3-4), and multiple objects (e.g., lines 5-6). These comparisons intuitively show a more competitive visual performance of our proposed AGLNet. With a good integration of the discriminative information provided by the additional cue, our AGLNet provides more accurate and complete camouflaged object localization and segmentation under various complex and highly similar backgrounds, even with the interference of noisy objects/regions (salient but non-camouflaged).

\begin{figure*}[t]
	\centering
	\subfigure[]{
		\begin{minipage}[b]{0.47\linewidth}
			\includegraphics[width=1\linewidth]{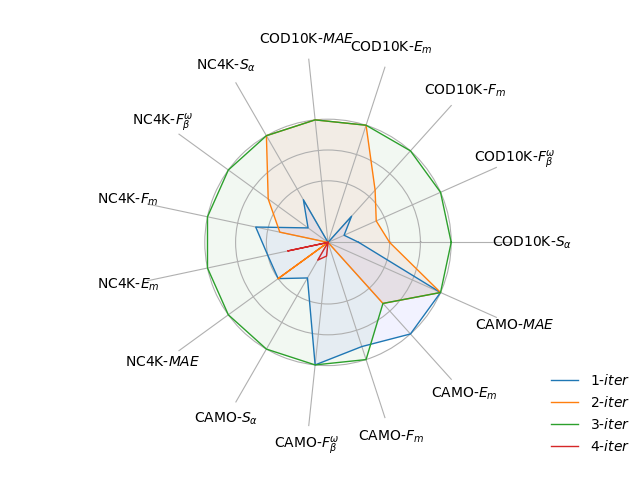}
		\end{minipage}
		\label{fig:mki_iteration}
	}
        \subfigure[]{
            \begin{minipage}[b]{0.47\linewidth}
                \includegraphics[width=1\linewidth]{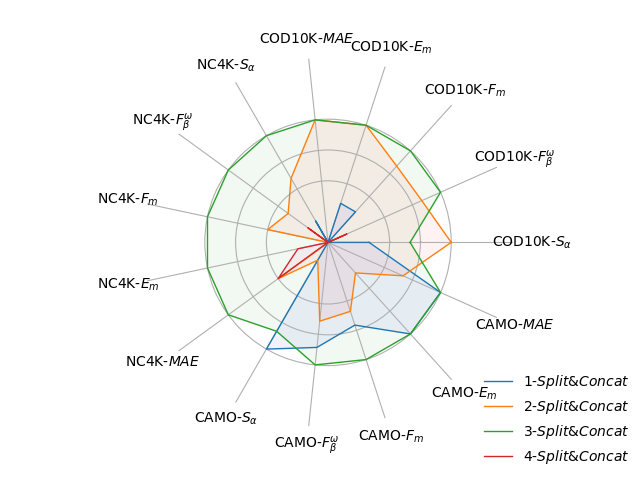}
            \end{minipage}
            \label{fig:mki_split_merge}
        }
        \\
        \subfigure[]{
            \begin{minipage}[b]{0.47\linewidth}
                \includegraphics[width=1\linewidth]{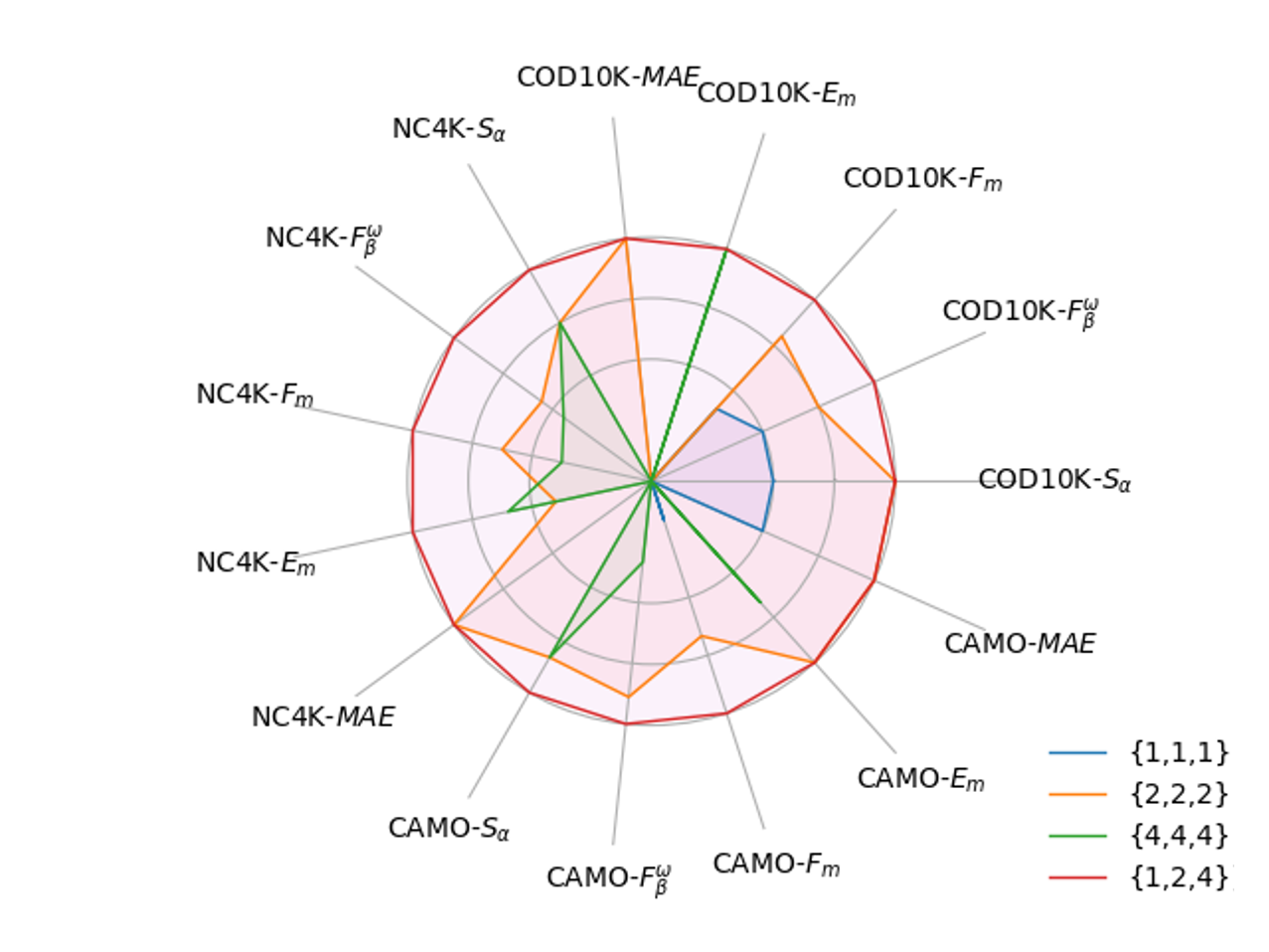}
            \end{minipage}
            \label{fig:q_in_FR}
        }
        \subfigure[]{
		\begin{minipage}[b]{0.47\linewidth}
			\includegraphics[width=1\linewidth]{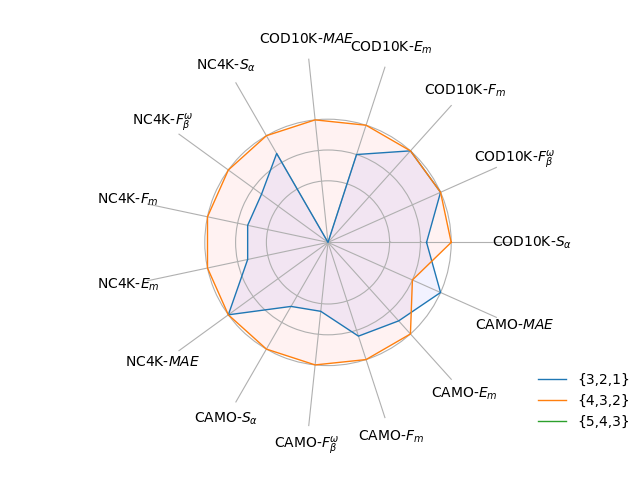}
		\end{minipage}
		\label{fig:n_in_FR}
	}
	\caption{Visualization of the ablation experiment results for each parameter of FR. Note that we have normalized the results for better display.}
	\label{fig:Ablation_FR.}
\end{figure*}

\subsection{Ablation Studies}
\noindent\textbf{Overview.} 
We perform ablation studies on key components to verify their effectiveness and analyze their impacts on performance, as shown in Table~\ref{tab:ablation}. 
Note that, for the baseline model, we remove all the additional modules, and then use convolution blocks to fuse the multi-level features in a top-down manner and generate predictions. 
Experimental results demonstrate that our designed HFC (including combination and decoupling), RD, and AIG can improve detection performance. When they are combined to build AGLNet, significant improvements in all evaluation metrics are observed. 

\vspace{10pt}
\noindent
\textbf{Effectiveness of Combination.} As can be seen from Table~\ref{tab:ablation} (\#2), compared with baseline, Combination achieves significant performance improvement, which provides a gain of 3.3\%, 9.6\%, 7.6\%, 3.9\%, and 26.1\% on $S_{\alpha}$, $F_{\beta}^{\omega}$, $F_{m}$, $E_{m}$ and $MAE$ on three datasets by an average, respectively. The Combination part fully interacts and accumulates critical cues by dense aggregation of multi-scale backbone features, thus greatly enhancing the feature representation for COD. Figure~\ref{fig:mfc} provides some visual results, showing the effectiveness of Combination for improving performance.


\begin{figure*}[t]
	\centering
	\subfigure[$S_\alpha$]{
		\begin{minipage}[b]{0.18\linewidth}
			\includegraphics[width=1\linewidth]{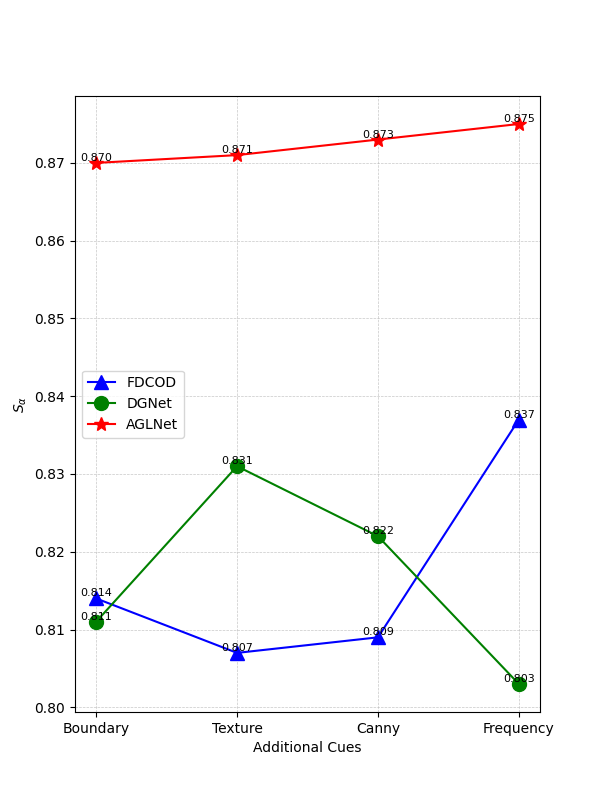}
		\end{minipage}
		\label{fig:COD10K_S}
	}
        \subfigure[$F_{\beta}^{\omega}$]{
            \begin{minipage}[b]{0.18\linewidth}
                \includegraphics[width=1\linewidth]{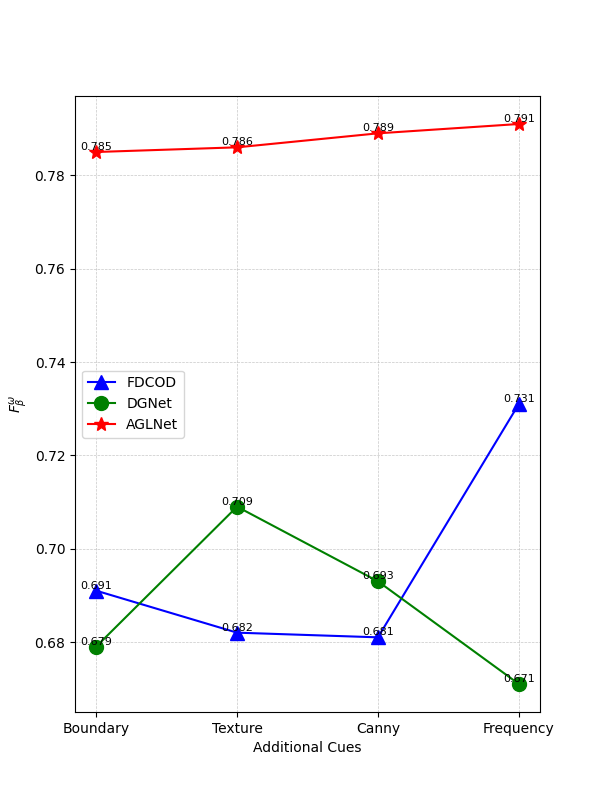}
            \end{minipage}
            \label{fig:COD10K_Fw}
        }
        \
        \subfigure[$F_{m}$]{
		\begin{minipage}[b]{0.18\linewidth}
			\includegraphics[width=1\linewidth]{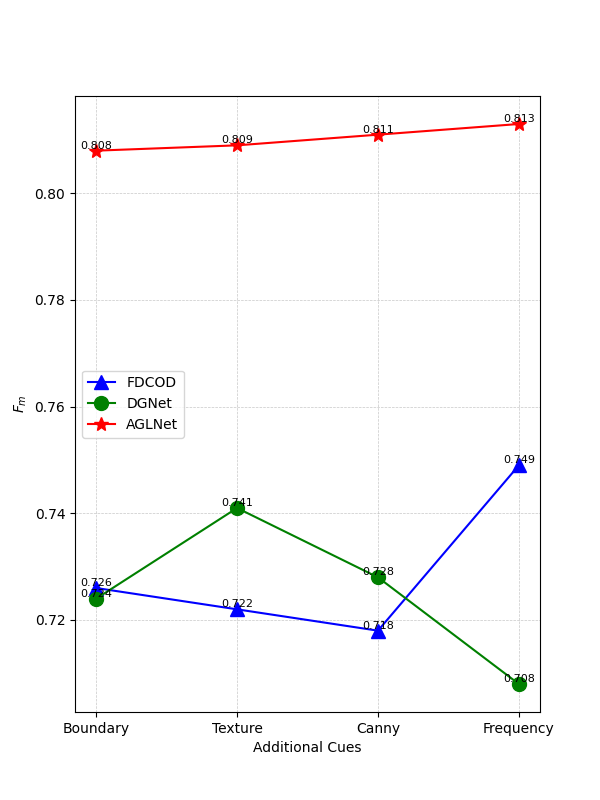}
		\end{minipage}
		\label{fig:COD10K_Fm}
	}
        \subfigure[$E_{m}$]{
            \begin{minipage}[b]{0.18\linewidth}
                \includegraphics[width=1\linewidth]{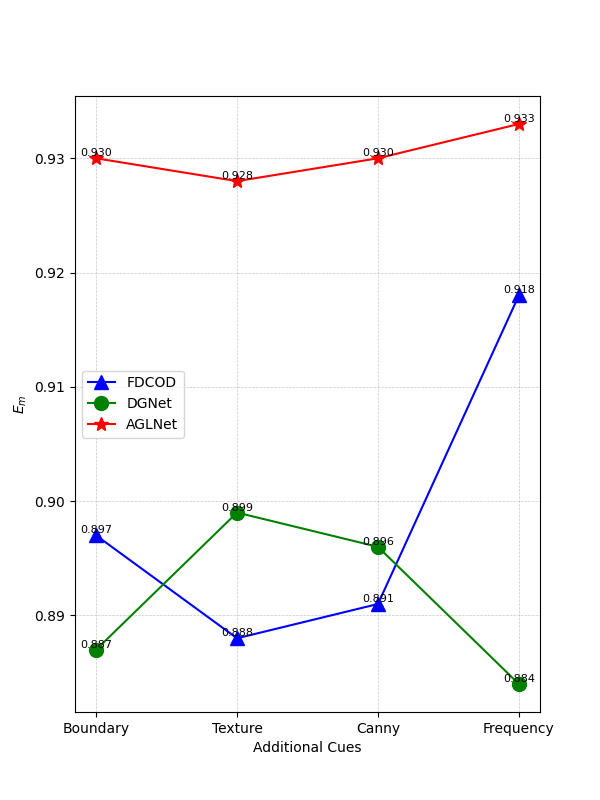}
            \end{minipage}
            \label{fig:COD10K_Em}
        }
        \subfigure[$MAE$]{
            \begin{minipage}[b]{0.18\linewidth}
                \includegraphics[width=1\linewidth]{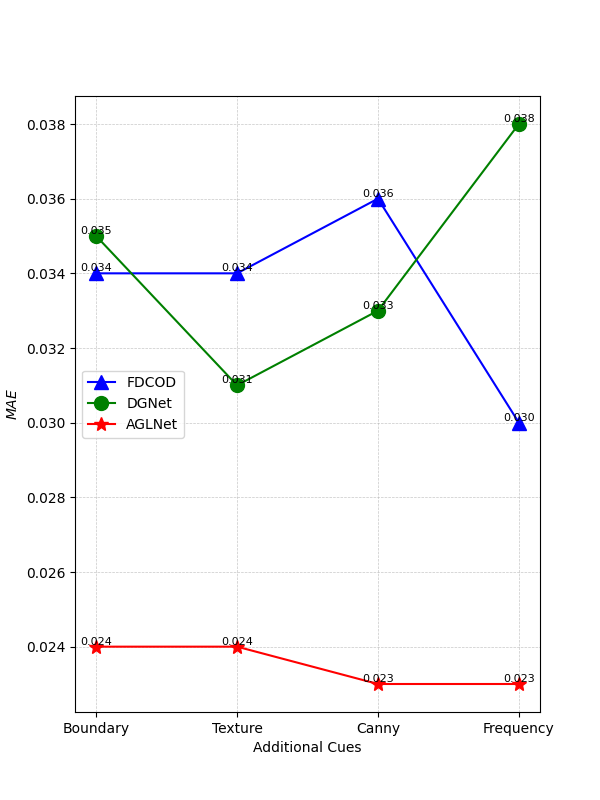}
            \end{minipage}
            \label{fig:COD10K_MAE}
        }
	\caption{Ablation studies of the model adaptability to different additional cues on the COD10K dataset. Please zoom in for more details.}
	\label{fig:ablation_add_cues}
\end{figure*}

\begin{figure*}[t]
	\centering
	\subfigure[Image]{
		\begin{minipage}[t]{0.12\textwidth}
			\centering
			\includegraphics[width=1.09\textwidth,height=0.840\textwidth]{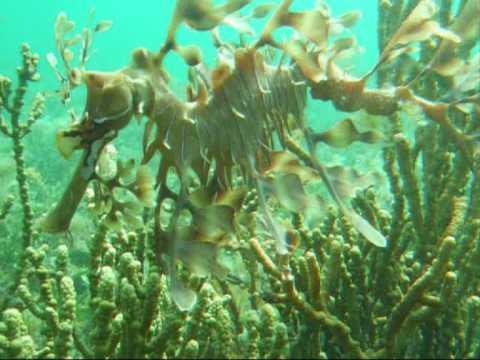}\\
			\vspace{0.01\linewidth}
            \includegraphics[width=1.09\textwidth,height=0.840\textwidth]{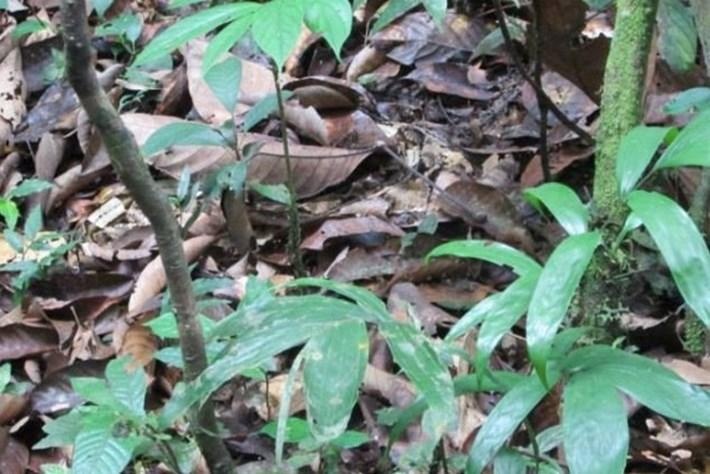}\\
            \vspace{0.01\linewidth}
			\includegraphics[width=1.09\linewidth,height=0.840\textwidth]{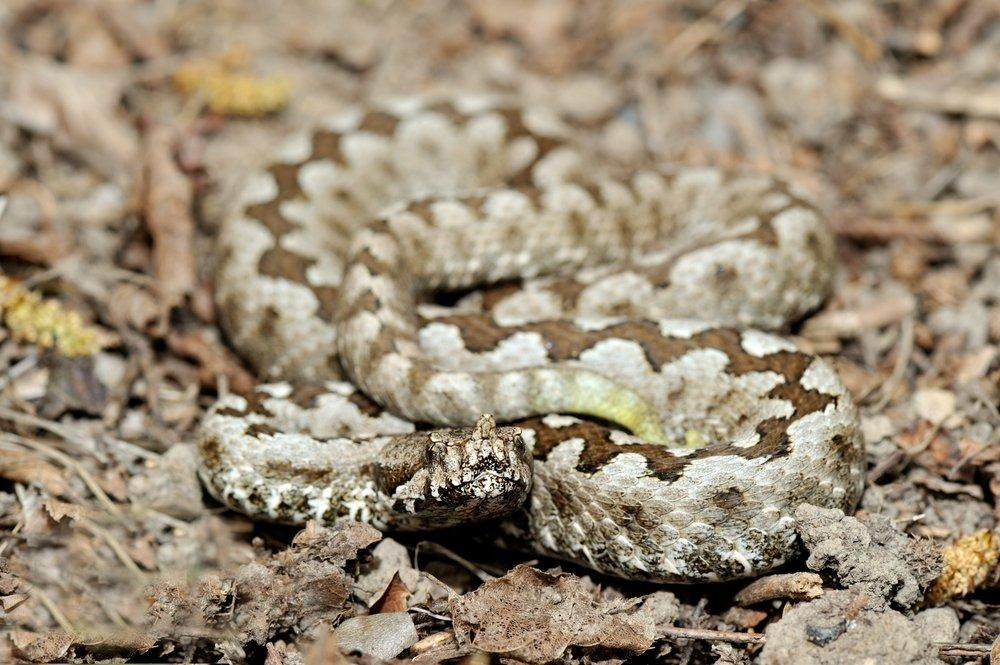}\\
			\vspace{0.01\linewidth}
		\end{minipage}%
	}\hspace{0.018\columnwidth}
	\subfigure[GT]{
		\begin{minipage}[t]{0.12\textwidth}
			\centering
			\includegraphics[width=1.09\textwidth,height=0.840\textwidth]{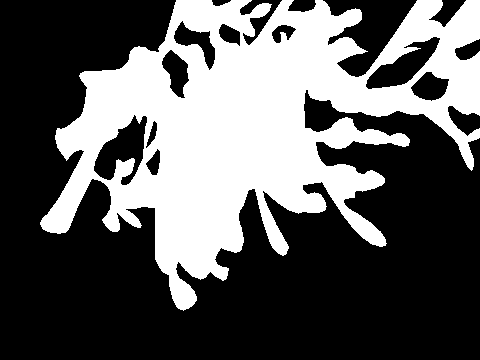}\\
			\vspace{0.01\linewidth}
            \includegraphics[width=1.09\textwidth,height=0.840\textwidth]{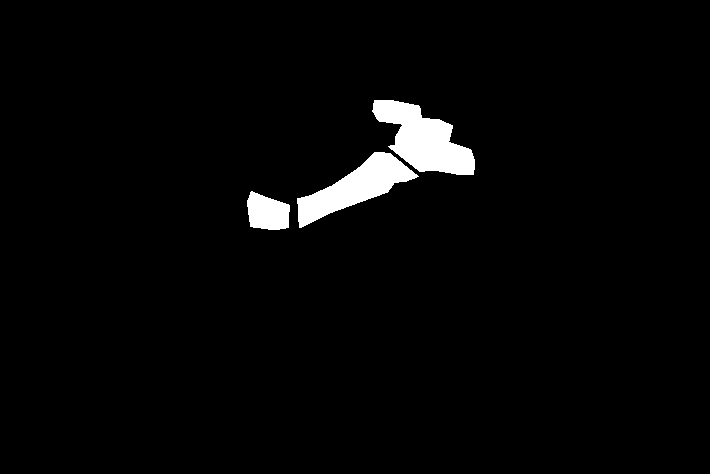}\\
            \vspace{0.01\linewidth}
			\includegraphics[width=1.09\linewidth,height=0.840\textwidth]{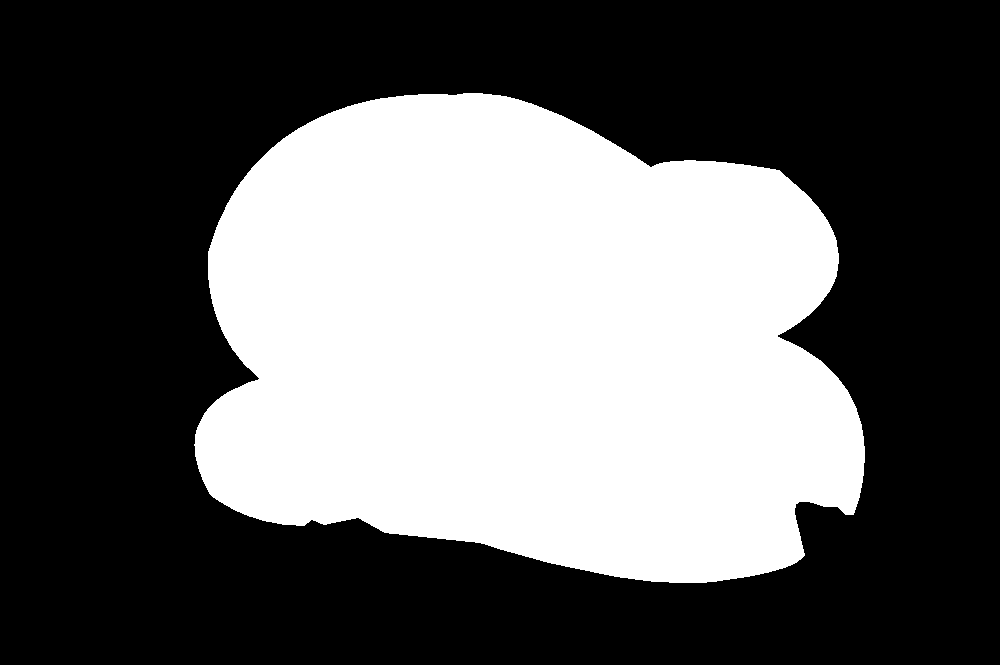}\\
			\vspace{0.01\linewidth}
		\end{minipage}%
	}\hspace{0.018\columnwidth}
	\subfigure[w/o AIG]{
		\begin{minipage}[t]{0.12\textwidth}
			\centering
			\includegraphics[width=1.09\textwidth,height=0.840\textwidth]{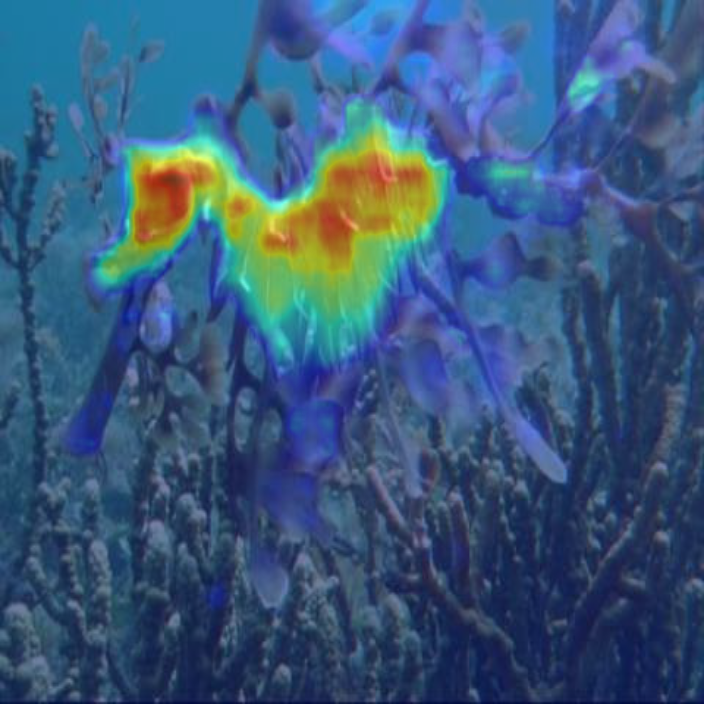}\\
			\vspace{0.01\linewidth}
            \includegraphics[width=1.09\textwidth,height=0.840\textwidth]{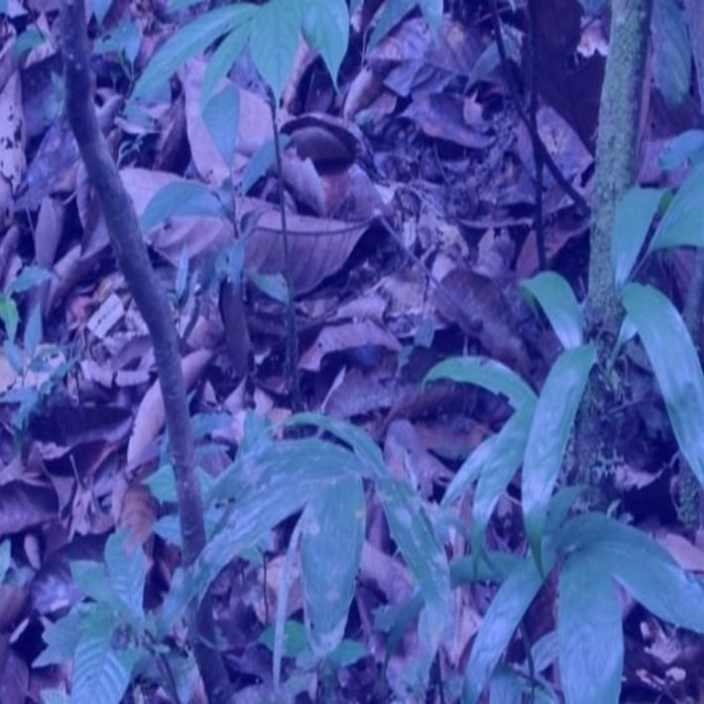}\\
            \vspace{0.01\linewidth}
			\includegraphics[width=1.09\linewidth,height=0.840\textwidth]{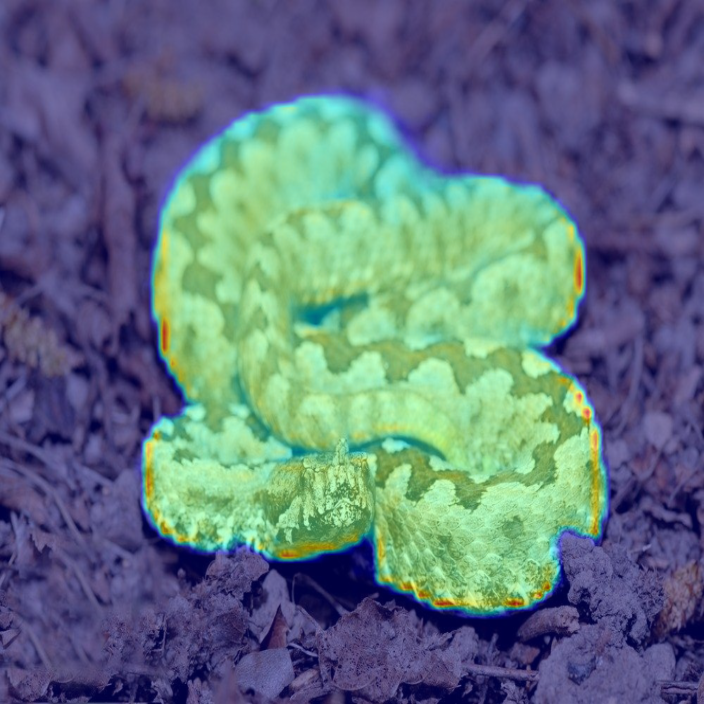}\\
			\vspace{0.01\linewidth}
		\end{minipage}%
	}\hspace{0.018\columnwidth}
	\subfigure[Boundary]{
		\begin{minipage}[t]{0.12\textwidth}
			\centering
			\includegraphics[width=1.09\textwidth,height=0.840\textwidth]{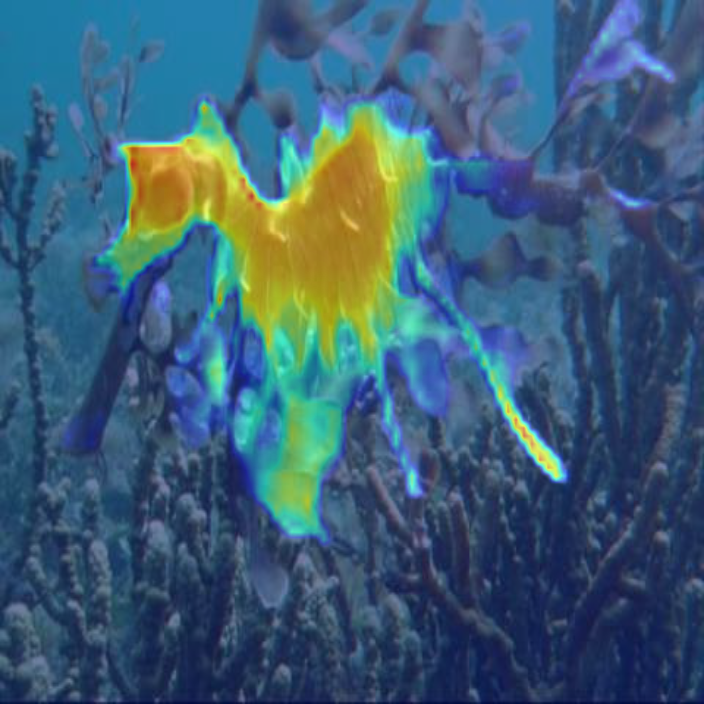}\\
			\vspace{0.01\linewidth}
            \includegraphics[width=1.09\textwidth,height=0.840\textwidth]{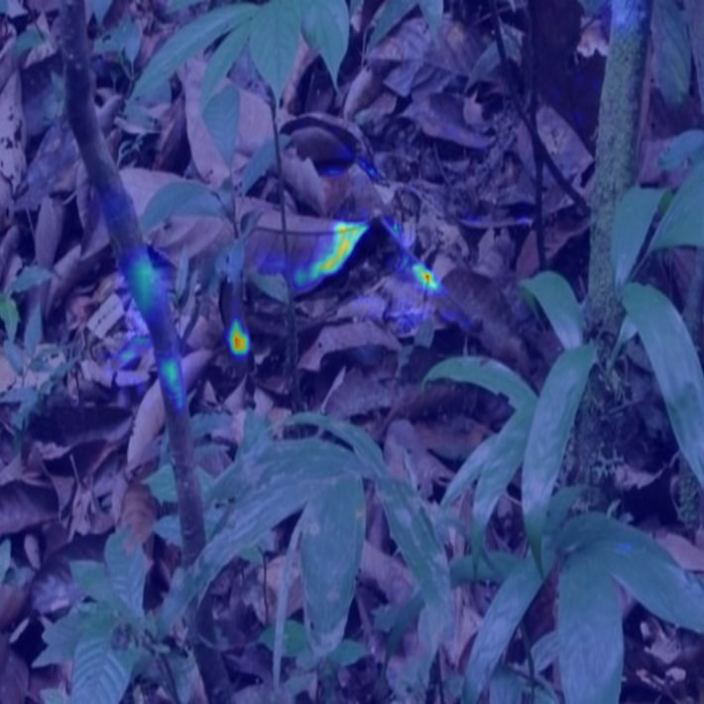}\\
            \vspace{0.01\linewidth}
			\includegraphics[width=1.09\linewidth,height=0.840\textwidth]{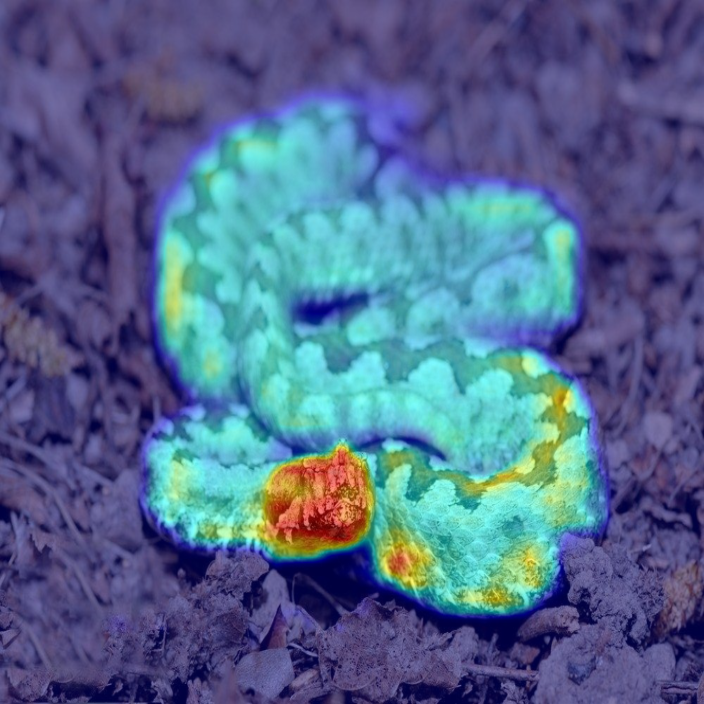}\\
			\vspace{0.01\linewidth}
		\end{minipage}%
	}\hspace{0.018\columnwidth}
	\subfigure[Texture]{
		\begin{minipage}[t]{0.12\textwidth}
			\centering
			\includegraphics[width=1.09\textwidth,height=0.840\textwidth]{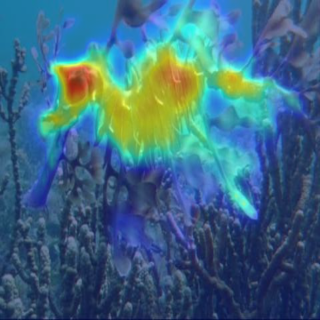}\\
			\vspace{0.01\linewidth}
            \includegraphics[width=1.09\textwidth,height=0.840\textwidth]{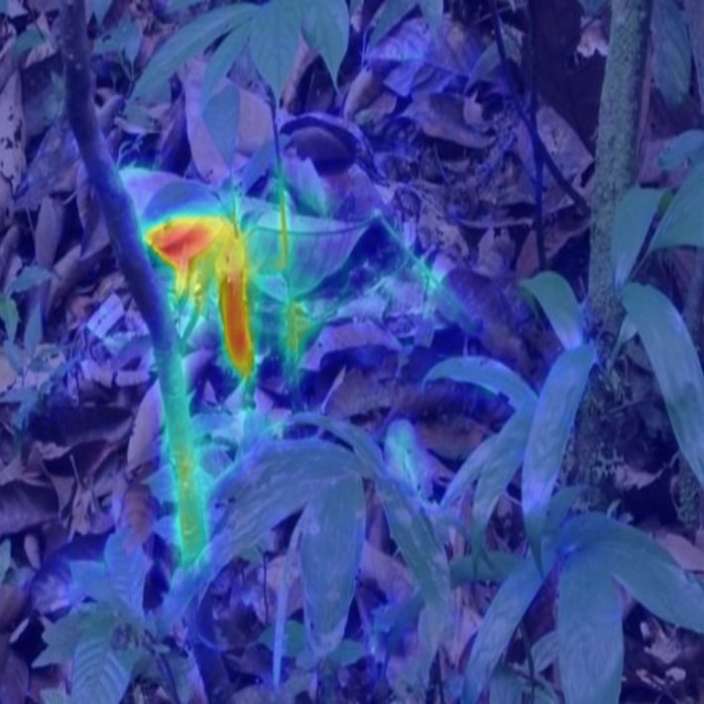}\\
            \vspace{0.01\linewidth}
			\includegraphics[width=1.09\linewidth,height=0.840\textwidth]{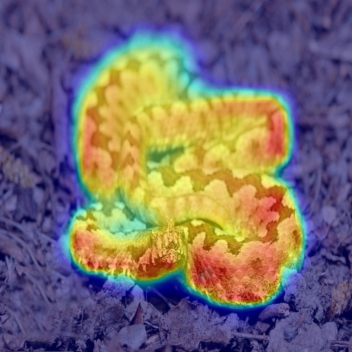}\\
			\vspace{0.01\linewidth}
		\end{minipage}%
	}\hspace{0.018\columnwidth}
	\subfigure[Canny]{
		\begin{minipage}[t]{0.12\textwidth}
			\centering
			\includegraphics[width=1.09\textwidth,height=0.840\textwidth]{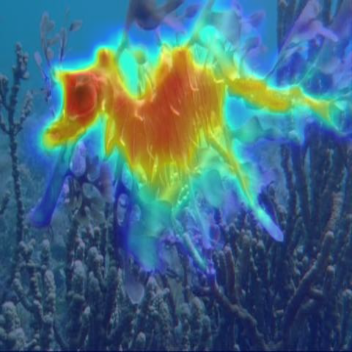}\\
			\vspace{0.01\linewidth}
            \includegraphics[width=1.09\textwidth,height=0.840\textwidth]{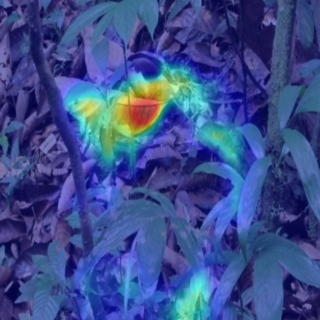}\\
            \vspace{0.01\linewidth}
			\includegraphics[width=1.09\linewidth,height=0.840\textwidth]{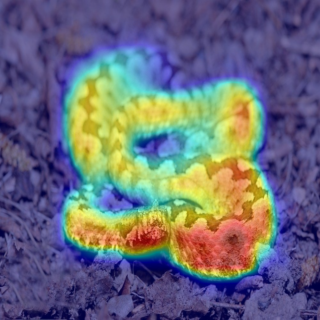}\\
			\vspace{0.01\linewidth}
		\end{minipage}%
	}\hspace{0.018\columnwidth}
	\subfigure[Frequency]{
		\begin{minipage}[t]{0.12\textwidth}
			\centering
			\includegraphics[width=1.09\textwidth,height=0.840\textwidth]{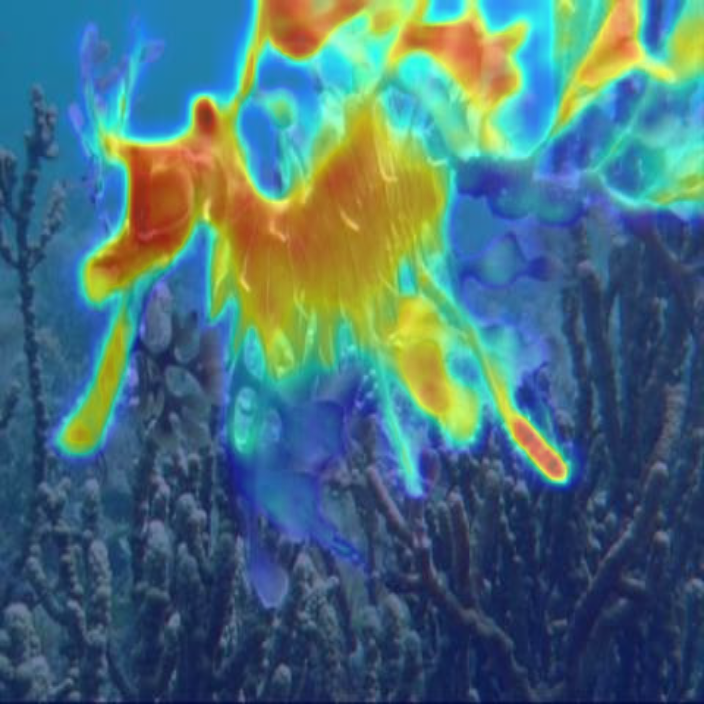}\\
			\vspace{0.01\linewidth}
            \includegraphics[width=1.09\textwidth,height=0.840\textwidth]{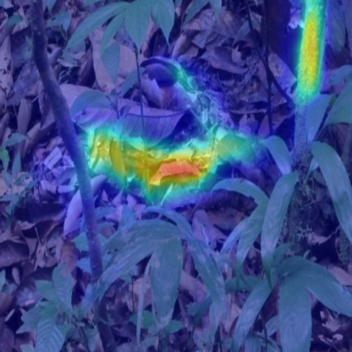}\\
            \vspace{0.01\linewidth}
			\includegraphics[width=1.09\linewidth,height=0.840\textwidth]{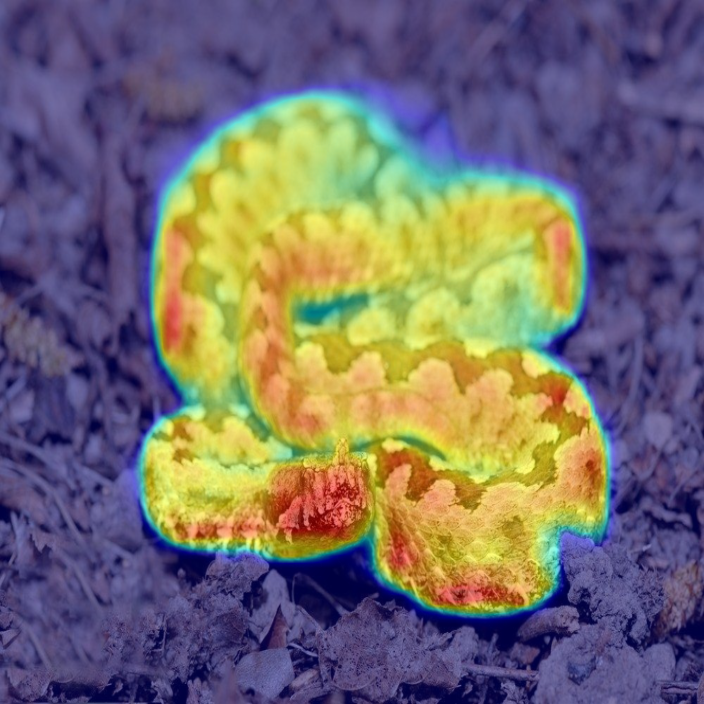}\\
			\vspace{0.01\linewidth}
		\end{minipage}%
	}\hspace{0.018\columnwidth}
	\centering
    
	\caption{\textbf{Visual comparison of different additional cues.} (a) input image, (b) ground-truth, (d)-(e) denote AGLNet using boundary, texture, canny and frequency as additional cues respectively.}
    \label{fig:ASG_source}
\end{figure*}

\begin{table*}[t]
\caption{Ablation studies of different visual backbones.}
\resizebox{\linewidth}{!}{
\renewcommand{\arraystretch}{1.3}
\begin{tabular}{c|ccccc|ccccc|ccccc|c}
\toprule[1pt]
\multirow{2}{*}{\textbf{Method (Backbone)}}  & \multicolumn{5}{c|}{\textbf{COD10K}}                                                                     & \multicolumn{5}{c|}{\textbf{NC4K}}                                                                                                                            & \multicolumn{5}{c|}{\textbf{CAMO}}                                                                                                                            & \multirow{2}{*}{\textbf{Params. (M)}} \\ \cline{2-16}
                                            & $S_{\alpha}\uparrow$   & $F_{\beta}^{\omega}\uparrow$     & $F_{m}\uparrow$   & $E_{m}\uparrow$    & $MAE\downarrow$                                   & \multicolumn{1}{c}{$S_{\alpha}\uparrow$}     & \multicolumn{1}{c}{$F_{\beta}^{\omega}\uparrow$}     & \multicolumn{1}{c}{$F_{m}\uparrow$}     & \multicolumn{1}{c}{$E_{m}\uparrow$}     & \multicolumn{1}{c|}{$MAE\downarrow$}               & \multicolumn{1}{c}{$S_{\alpha}\uparrow$}     & \multicolumn{1}{c}{$F_{\beta}^{\omega}\uparrow$}     & \multicolumn{1}{c}{$F_{m}\uparrow$}     & \multicolumn{1}{c}{$E_{m}\uparrow$}     & \multicolumn{1}{c|}{$MAE\downarrow$}                 &                                       \\ \midrule[1pt]
\multicolumn{1}{c|}{ZoomNet (ResNet-50)}        & 0.838          & 0.729          & 0.766          & 0.888          & \multicolumn{1}{c|}{0.029}          & 0.853                     & 0.784                     & 0.818                     & 0.896                     & \multicolumn{1}{c|}{0.043}          & 0.820                     & 0.752                     & 0.794                     & 0.878                     & \multicolumn{1}{c|}{0.066}          & 32.382                                \\
\rowcolor[HTML]{EFEFEF} 
\multicolumn{1}{c|}{AGLNet (ResNet-50)}       & 0.849          & 0.740          & 0.773          & 0.920          & \multicolumn{1}{c|}{0.028}          & 0.857                     & 0.789                     & 0.823                     & 0.902                     & \multicolumn{1}{c|}{0.042}          & 0.842                     & 0.768                     & 0.803                     & 0.888                     & \multicolumn{1}{c|}{0.064}          & 114.09                                \\ \hline
\multicolumn{1}{c|}{FDCOD (Res2Net-50)}      & 0.837 & 0.731  & 0.749          & 0.918          & \multicolumn{1}{c|}{0.030}          & 0.834          & 0.750          & 0.784          & 0.894          & \multicolumn{1}{c|}{0.052}          & 0.844          & 0.778          & 0.809          & 0.898          & \multicolumn{1}{c|}{0.062}          & 197.41                                \\
\rowcolor[HTML]{EFEFEF} 
\multicolumn{1}{c|}{AGLNet (Res2Net-50)}      & 0.856          & 0.753          & 0.784          & 0.926          & \multicolumn{1}{c|}{0.028}          & 0.863                     & 0.793                     & 0.826                     & 0.906                     & \multicolumn{1}{c|}{0.042}          & 0.851                     & 0.779                     & 0.819                     & 0.895                     & \multicolumn{1}{c|}{0.061}          & 114.69                                \\ \hline
\multicolumn{1}{c|}{DGNet (EfficientNet-B4)} & 0.822          & 0.693          & 0.728          & 0.896          & \multicolumn{1}{c|}{0.033}          & \multicolumn{1}{c}{0.857} & \multicolumn{1}{c}{0.784} & \multicolumn{1}{c}{0.814} & \multicolumn{1}{c}{0.911} & \multicolumn{1}{c|}{0.042}          & \multicolumn{1}{c}{0.839} & \multicolumn{1}{c}{0.769} & \multicolumn{1}{c}{0.806} & \multicolumn{1}{c}{0.901} & \multicolumn{1}{c|}{0.057}          & \textbf{21.02}                        \\
\rowcolor[HTML]{EFEFEF} 
\multicolumn{1}{c|}{AGLNet (EfficientNet-B4)} & \textbf{0.875} & \textbf{0.791} & \textbf{0.813} & \textbf{0.933} & \multicolumn{1}{c|}{\textbf{0.023}} & \textbf{0.889}            & \textbf{0.836}            & \textbf{0.858}            & \textbf{0.934}            & \multicolumn{1}{c|}{\textbf{0.033}} & \textbf{0.874}            & \textbf{0.825}            & \textbf{0.851}            & \textbf{0.918}            & \multicolumn{1}{c|}{\textbf{0.050}} & 93.65                                 \\ \bottomrule[1pt]
\end{tabular}
}
\label{tab:backbone}
\end{table*}

\begin{table*}[t]
\caption{Ablation studies of AGLNet at different input resolutions.}
\resizebox{\linewidth}{!}{
\renewcommand{\arraystretch}{1.3}
\begin{tabular}{c|ccccc|ccccc|ccccc}
\toprule[1pt]
\multirow{2}{*}{\textbf{Method}} & \multicolumn{5}{c|}{\textbf{COD10K}}                                                                     & \multicolumn{5}{c|}{\textbf{NC4K}}                                                                       & \multicolumn{5}{c}{\textbf{CAMO}}                                                  \\ \cline{2-16} 
                                 & $S_{\alpha}\uparrow$              & $F_{\beta}^{\omega}\uparrow$             & $F_{m}\uparrow$             & $E_{m}\uparrow$             & $MAE\downarrow$                                & $S_{\alpha}\uparrow$              & $F_{\beta}^{\omega}\uparrow$             & $F_{m}\uparrow$             & $E_{m}\uparrow$             & $MAE\downarrow$                                 & $S_{\alpha}\uparrow$              & $F_{\beta}^{\omega}\uparrow$             & $F_{m}\uparrow$             & $E_{m}\uparrow$             & $MAE\downarrow$            \\ \midrule[1pt]
SegMaR (704*704)                   & 0.830 & 0.708 & 0.745 & 0.894 & \multicolumn{1}{c|}{0.033} & 0.845 & 0.762 & 0.799 & 0.892 & \multicolumn{1}{c|}{0.050} & 0.792 & 0.701 & 0.748 & 0.843 & 0.085          \\
ZoomNet (704*704)                   & 0.842          & 0.738          & 0.778          & 0.891          & \multicolumn{1}{c|}{0.029}          & 0.854          & 0.786          & 0.822          & 0.896               & \multicolumn{1}{c|}{0.043}               & 0.797          & 0.721          & 0.768          & 0.845          & 0.080          \\ 
FDCOD (704*704)                   & 0.843          & 0.733          & 0.761          & 0.902          & \multicolumn{1}{c|}{0.030}          & 0.842          & 0.761          & 0.793          & 0.896               & \multicolumn{1}{c|}{0.049}               & 0.850          & 0.784          & 0.821          & 0.886          & 0.059          \\ 
HitNet (704*704)        & {0.868}         & \textbf{0.798}           & {0.806}          & {0.932}          &{0.024}         & 0.870          & {0.825}          & {0.853}          & {0.921}          & {0.039}          & 0.844          & {0.801}          & {0.831}          & {0.902}          & {0.057}
\\
\hline 
\rowcolor[HTML]{EFEFEF} 
AGLNet (704*704) & \textbf{0.875} & {0.791} & \textbf{0.813} & \textbf{0.933} & \multicolumn{1}{c|}{\textbf{0.023}} & \textbf{0.889}            & \textbf{0.836}            & \textbf{0.858}            & \textbf{0.934}            & \multicolumn{1}{c|}{\textbf{0.033}} & \textbf{0.874}            & \textbf{0.825}            & \textbf{0.851}            & \textbf{0.918}            & \multicolumn{1}{c}{\textbf{0.050}} \\ 
\bottomrule[1pt]
\end{tabular}
}
\label{tab:resolution}
\end{table*}


\vspace{10pt}
\noindent
\textbf{Effectiveness of Decoupling.} As shown in Table~\ref{tab:ablation} (\#3), the addition of Decoupling part significantly improves detection performance by 1.2\%, 0.9\% and 1.1\% on $F_{\beta}^{\omega}$, $F_{m}$ and $E_{m}$ on three datasets by an average, respectively. Decoupling adopts feature splitting and group-wise exploration to dig deep into different feature groups for fine discriminative features, which strengthen feature representation. The decoupling operation compensates for more details for fine feature exploration. Figure~\ref{fig:fgd} shows some visual results. We can see Decoupling part plays a crucial role in fine feature exploration, which compensates for more details, such as edges (\textit{e.g.} row 1), textures (\textit{e.g.} row 2), and torsos (\textit{e.g.} row 3) for camouflaged object detection.

\vspace{10pt}
\noindent
\textbf{Effectiveness of RD.} The RD module utilizes three FR components, which combine multi-scale backbone features to further refine feature representation for camouflaged object detection. As shown in Table~\ref{tab:ablation} (\#4), RD further increases the detection performance. Figure \ref{fig:RD} provides some visual comparison results, showing the effectiveness of the proposed RD module.

\vspace{10pt}
\noindent
\textbf{Parameter Analysis for FR.} 
Fig.~\ref{fig:mki_iteration} shows the performance comparison of $M^{k}_{i}$ module in FR under different iterations. We can see that the model performance is relatively robust for different iterations. In our experiments, we adopt three iterations of $M^{k}_{i}$ module, which achieves the slightly best performance for camouflaged object detection.
Besides, inside $M^{k}_{i}$ module, we perform multiple split and merge operations to deeply explore critical features for camouflaged object detection. Fig. \ref{fig:mki_split_merge} shows the performance comparison under the different numbers of split and merge operations. In our experiments, we employ three times of split-merge operations, which achieve the best detection performance.
Fig.~\ref{fig:q_in_FR} provides a quantitative comparison of parameter $q$ in FR under different settings. Fig. \ref{fig:n_in_FR} provides a quantitative comparison of parameter $n$ in FR under different settings. 
Experiments show that the performance of FR is relatively robust under different parameter settings. We chose $q=\{4,2,1\}$ and $n=\{4,3,2\}$, respectively, which achieve slightly better performance in camouflaged object detection.

\vspace{10pt}
\noindent
\textbf{Effectiveness of AIG.} 
The AIG module learns additional information features of camouflaged objects and then incorporates them into image features for camouflaged object segmentation. To make full use of the additional cues, multi-level fusion is adopted to incorporate additional information features into different stages of the model, including HFC and RD, for deep aggregation. As shown in Table~\ref{tab:ablation} (\#OUR), the performance gains are 1.5\%, 1.8\%, 1.3\% and 8.0\% in terms of $F_{\beta}^{\omega}$, $F_{m}$, $E_{m}$ and $MAE$ on COD10K, respectively, demonstrating the effectiveness of additional information features for camouflaged object detection. 


\vspace{10pt}
\noindent
\textbf{Model Adaptability to Different Additional Cues.} Fig.~\ref{fig:ablation_add_cues} shows a comparison of the model adaptability of some representative COD methods for different additional cues, including boundary, texture, canny, and frequency. We adopt FDCOD~\cite{zhong2022detecting} and DGNet~\cite{ji2022gradient} as comparison. The former introduces the frequency domain information and the latter integrates edge cues.
We can see, FDCOD achieves better results using frequency domain cues, but it significantly reduces the performance with other additional cues. Differently, DGNet shows poor performance when using the frequency domain cues. 
These methods are tailored to specific auxiliary cues and not applicable to other additional cues. By contrast, our proposed method shows outstanding adaptability for different additional cues. Our method achieves the best performance under different additional information with minor performance differences when compared to other competitors.  
Fig.~\ref{fig:ASG_source} shows the feature maps of different additional cues. We can see that: (a) the boundary provides relatively little object information (only the edges of the object silhouette), so it is more sensitive to the object silhouette, but explores more limited effective object features than the other three additional cues. (b) Boundary and texture only provide cues to object regions, while the canny and frequency also provide contextual information (\textit{i.e.}, background of the objects), which helps to improve the understanding of a scene and increase detection performance. (c) Frequency provides additional information beyond the human visual system and shows the best results.

\vspace{10pt}
\noindent
\textbf{Backbone Analysis.} We also test different backbones to verify the performance of the proposed method for COD. ZoomNet~\cite{pang2022zoom}, FDCOD~\cite{zhong2022detecting} and DGNet~\cite{ji2022gradient} are state-of-the-art methods, with common-used ResNet-50 \cite{he2016deep}, Res2Net-50 \cite{gao2019res2net} and EfficientNet-B4~\cite{tan2019efficientnet} as the backbones, respectively. Therefore, we take these three methods as competitors. As shown in Table~\ref{tab:backbone}, we test our proposed AGLNet with three kinds of backbones respectively, and find that the proposed method outperforms other competitors significantly. Specifically, compared with FDCOD, AGLNet (Res2Net-50) achieves 2.3\%, 3.0\% and 4.7\% improvement in $S_{\alpha}$, $F_{\beta}^{\omega}$ and $F_{m}$ on COD10K dataset, respectively. Compared with DGNet, AGLNet (EfficientNet-B4) achieves average improvements of 4.9\%, 9.4\%, 7.6\%, 2.8\% and 21.3\% in $S_{\alpha}$, $F_{\beta}^{\omega}$, $F_{m}$, $E_{m}$, and $MAE$ on three datasets by an average, respectively. 
Overall, our AGLNet achieves the best performance with EfficientNet-B4 as the backbone. Besides, in the AGLNet variant, AGLNet (EfficientNet-B4) has the smallest amount of parameters, but is still larger than that of DGNet and ZoomNet. The design of light-weight models is also the focus of our future work.

\vspace{10pt}
\noindent
\textbf{Input Resolution.} We also conduct a series of ablation experiments to analyze the impact of input image resolution on detection performance. 
As shown in Table~\ref{tab:resolution}, under the same resolution (704$\times$704), the proposed AGLNet significantly outperforms the comparison methods. 
This is because: a) high-resolution input provides more effective object details to improve the detection; b) high-resolution input also introduces noise interference, so a good network design can better explore critical cues. The proposed method designs the deep integration of additional features and image features and recalibration decoder, providing a very compelling performance. 
Actually, from our experiments, our method achieves the best results at all common input resolutions. 

\section{Conclusion}
This paper proposes an adaptive guidance learning framework that can handle any of the additional cues theoretically which are commonly used in COD tasks while achieving significant performance gains. 
To our knowledge, this is the first work to study a unified end-to-end model to adapt different additional information for COD tasks. 
The proposed method designs an additional information generation module to learn the additional cues, which are then deeply integrated with image features by the hierarchical feature combination module to guide the learning of camouflaged features. 
%
Extensive experiments show the superiority over 20 other state-of-the-art methods on three datasets.

\bibliographystyle{IEEEtran}
\bibliography{main}

\end{document}